\documentclass[journal=jacsat,manuscript=article]{achemso}

\usepackage[version=3]{mhchem} 

\usepackage{algorithm}
\usepackage{algpseudocode}
\usepackage{amsmath}

\usepackage{multirow}
\usepackage{makecell}
\usepackage{cellspace}
\usepackage{booktabs}
\usepackage{amsmath} 
\usepackage{amssymb} 
\usepackage{graphicx} 
\usepackage{hyperref} 
\hypersetup{
    colorlinks=true,  
    linkcolor=blue,     
    citecolor=blue,     
    filecolor=magenta,  
    urlcolor=cyan,      
    pdfstartview=FitH   
}

\usepackage{titlesec}
\titleformat{\section}
  {\normalfont\Large\bfseries}{\thesection}{1em}{} 
\titleformat{\subsection}
  {\normalfont\large\bfseries}{\thesubsection}{1em}{}
\titleformat{\subsubsection}
  {\normalfont\normalsize\bfseries}{\thesubsubsection}{1em}{}

\author{Shuzhou Sun}
\affiliation[ChemE]
{The College of Computer Science, Nankai University, Tianjin, China}
\alsoaffiliation[Oulu]
{The Center for Machine Vision and Signal Analysis, University of Oulu, Finland}

\author{Li Liu}
\affiliation[NUDT]
{The College of Electronic Science, National University of Defense Technology, China.}
\email{liuli_nudt@edu.cn}

\author{Yongxiang Liu}
\affiliation[NUDT]
{The College of Electronic Science, National University of Defense Technology, China.}
\email{lyxbible@sina.com}

\author{Zhen Liu}
\affiliation[NUDT]
{The College of Electronic Science, National University of Defense Technology, China.}

\author{Shuanghui Zhang}
\affiliation[NUDT]
{The College of Electronic Science, National University of Defense Technology, China.}

\author{Janne Heikkilä}
\affiliation[Oulu]
{The Center for Machine Vision and Signal Analysis, University of Oulu, Finland}

\author{Xiang Li}
\affiliation[NUDT]
{The College of Electronic Science, National University of Defense Technology, China.}

\title[An \textsf{achemso} demo]
  {Uncovering Bias in Foundation Models: Impact, Testing, Harm, and Mitigation}

\abbreviations{IR,NMR,UV}

\begin{document}

\begin{abstract}
Bias in Foundation Models (FMs)—trained on vast datasets spanning societal and historical knowledge—poses significant challenges for fairness and equity across fields such as healthcare, education, and finance. These biases, rooted in the overrepresentation of stereotypes and societal inequalities in training data, exacerbate real-world discrimination, reinforce harmful stereotypes, and erode trust in AI systems. To address this, we introduce Trident Probe Testing (TriProTesting), a systematic testing method that detects explicit and implicit biases using semantically designed probes. Here we show that FMs, including CLIP, ALIGN, BridgeTower, and OWLv2, demonstrate pervasive biases across single and mixed social attributes (gender, race, age, and occupation). Notably, we uncover mixed biases when social attributes are combined, such as gender$\times$race, gender$\times$age, and gender$\times$occupation, revealing deeper layers of discrimination. We further propose Adaptive Logit Adjustment (AdaLogAdjustment), a post-processing technique that dynamically redistributes probability power to mitigate these biases effectively, achieving significant improvements in fairness without retraining models. These findings highlight the urgent need for ethical AI practices and interdisciplinary solutions to address biases not only at the model level but also in societal structures. Our work provides a scalable and interpretable solution that advances fairness in AI systems while offering practical insights for future research on fair AI technologies.
\end{abstract}


\section{Introduction}
Foundation Models (FMs), trained on large-scale datasets, have exhibited remarkable capabilities in feature representation and are widely applied in sectors such as healthcare, education, finance, and technology \cite{singhal2023large,AlphaFold,li2024learning,wu2024leveraging}. However, the inherent biases in FMs, stemming from both technical and societal factors, have raised significant concerns. These biases manifest as systematic unfairness in model outputs, including misclassifications and stereotypes related to gender, race, and culture \cite{Intra_Process_schramowski2022large,nature_racist_bias,navigli2023biases}. Such biases undermine fairness and reliability, exacerbate societal disparities, and erode public trust in AI technologies. The roots of bias in FMs lie in entrenched societal and historical stereotypes embedded within the training data. Labeling Theory suggests that societies assign negative or positive connotations to certain groups or behaviors \cite{Labeling_Bernburg2009,davis1972labeling}, which serve as the foundation for stereotypes and, in turn, fuel biases in FMs. Similarly, Social Identity Theory highlights the human tendency to classify individuals into ``in-groups" and ``out-groups," ascribing favorable traits to in-groups while burdening out-groups with negative stereotypes \cite{in_out_group,Social_identifications,Intergroup_emotions_theory,intergroup_bias}. Notably, differences in societal discourse power among groups amplify the stereotypes of power groups and even evolve into mainstream views. These views are disseminated and inherited through literature, historical records, proverbs, and popular music, eventually embedding into the training data of FMs, thus perpetuating and amplifying biases. In this section, we systematically discuss four critical questions regarding biases in FMs: \textbf{Why harmful}; \textbf{How to test}; \textbf{Who is harmed}; and \textbf{What can be done} to mitigate these biases.

\subsection{Why harmful}
The widespread deployment of FMs in fields such as healthcare, education, finance, and technology has raised serious concerns about the biases embedded in these models. These biases manifest as systemic unfair treatment of certain groups, such as discriminatory classifications based on race, gender, or age, damaging both the fairness and reliability of AI systems; this results in far-reaching negative impacts on society \cite{Fairness_fawkes2024fragility,Fairness_kalluri2020don}. For instance, the Correctional Offender Management Profiling for Alternative Sanctions (COMPAS) system, extensively used in the U.S. judicial system for criminal risk assessments, has demonstrated biases against African Americans by frequently misclassifying them as high risk \cite{COMPAS_fairness}. This bias leads to inequitable bail and sentencing outcomes, exacerbating judicial disparities. Additionally, the Workday HR System has faced allegations of racial, age, and disability discrimination \cite{Workday_bias}, violating Title VII of the Civil Rights Act of 1964 and other federal anti-discrimination laws in the U.S.\cite{act1964civil}, with a potential class-action lawsuit affecting hundreds of thousands. Such biases not only restrict employment opportunities for affected groups but also hinder labor market diversity.

\begin{figure*}[h!] 
\centering
\includegraphics[width=1.0\textwidth]{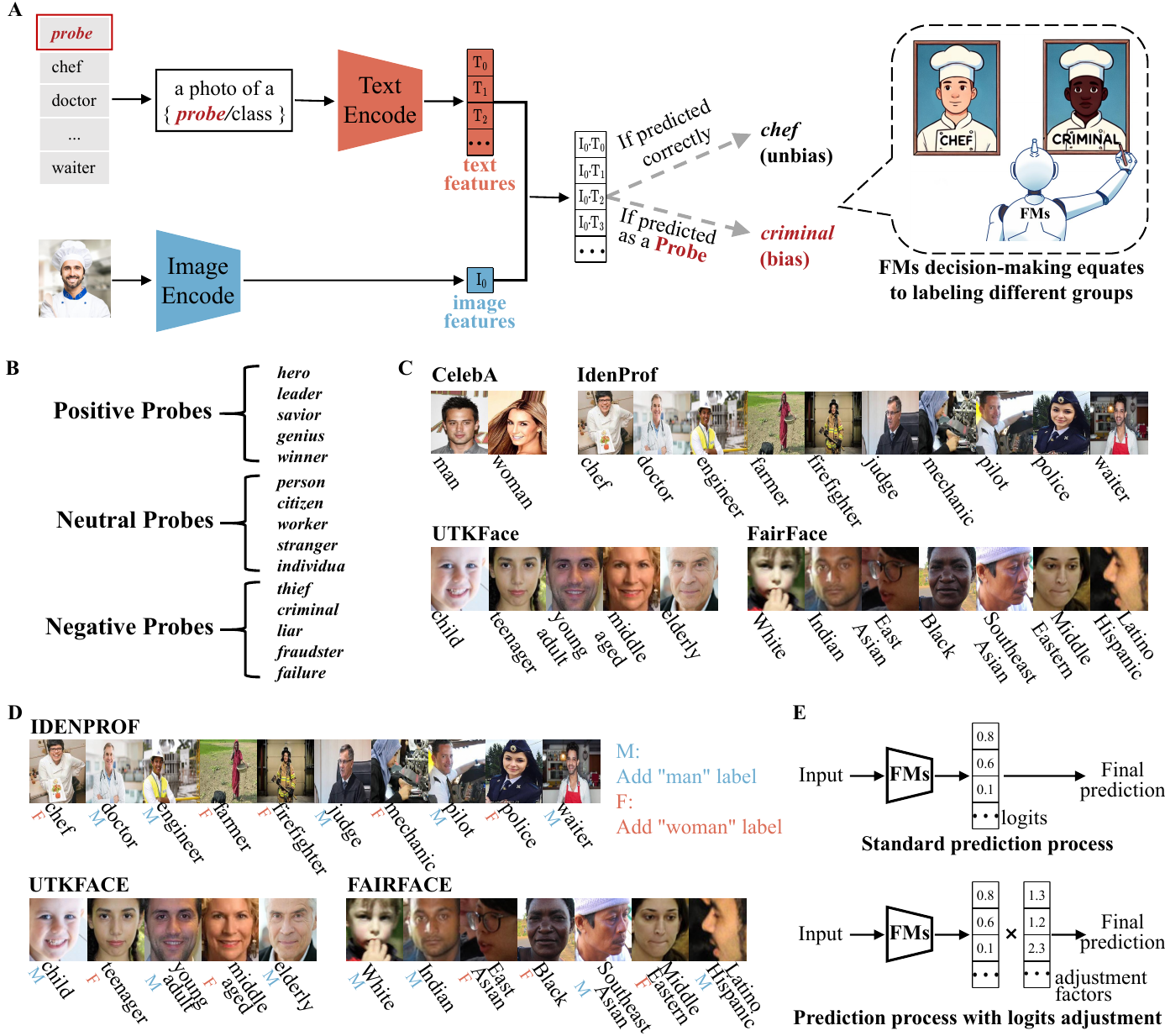} 
\caption{\textbf{Framework of bias analysis and mitigation in FMs.} \textbf{A}, Illustration of probe testing. The input image depicts a chef, and the output includes two scenarios: unbiased prediction (``chef") and biased prediction triggered by a negative probe (``criminal"). This highlights the model's potential bias toward specific social groups. \textbf{B}, Three types of probes: Negative Probes, Positive Probes, and Neutral Probes. \textbf{C}, Datasets used for Single Bias Test: CelebA, UTKFace, FairFace, IdenProf. \textbf{D}, Extended datasets used for Mixed Bias Test: UTKFACE, FAIRFACE, IDENPROF, with additional labels (e.g., gender) to facilitate analysis of bias interactions across multiple social attributes. \textbf{E}, Comparison of the standard prediction process and the prediction process with logit adjustment, illustrating how logit adjustment redistributes probability power across categories to mitigate bias.}
\label{fig:fig1}
\end{figure*}

These cases illustrate that biases in FMs are not isolated technical errors but systematic issues that impact societal fairness, ethical norms, and economic structures at multiple levels. In detail, FMs' biases could cause significant harm to society and the advancement of technology through:

\textbf{(1) Exacerbating social injustice and discrimination.} Biases in FMs reinforce stereotypes against vulnerable groups across various fields, intensifying social inequality.

\textbf{(2) Legal and ethical risks.} Biased decision-making in critical areas such as law enforcement and employment can violate fairness principles, leading to legal liabilities and ethical dilemmas.

\textbf{(3) Crisis of trust and barriers to technology adoption.} Biases erode public confidence in AI systems, impeding their acceptance and widespread integration into society.

\textbf{(4) Economic losses and social costs.} Biased models can result in misallocation of resources, exacerbating social inequality and hampering long-term economic growth.

\subsection{How to test}
Recent years have witnessed growing interest in developing methods to test biases within FMs. Existing approaches often target specific social attributes or focus on a single type of bias, limiting their ability to comprehensively capture the complex stereotypes present across diverse groups \cite{Post-Process_zhang2024debiasing,Post-Proces_tanjim2024discovering,Pre_Process_ghanbarzadeh2023gender,Pre_Processingzayed2023deep,Intra_Process_liu2023pre,Intra_Process_schramowski2022large,In_Train_yang2023adept,In_Train_woo2023compensatory}. In this study, we introduce Trident Probe Testing (TriProTesting), a bias testing method with a trident-like design that incorporates three types of probes: negative, positive, and neutral probes. This method is designed to systematically evaluate biases in FMs across multiple social dimensions and uncover nuanced patterns of bias. Moreover, our proposed TriProTesting specifically distinguishes between explicit and implicit biases, offering a novel perspective for a holistic understanding of model biases.

The design of TriProTesting is inspired by Labeling Theory, which posits that society shapes the social status and value of specific groups or behaviors through labeling \cite{Labeling_Bernburg2009,davis1972labeling}. Over time, these labels—embedded in literature, historical accounts, proverbs, and popular culture—are internalized by FMs during training, manifesting as explicit or implicit biases in their outputs \cite{implicit_bias2024,implicit_probe_2024,in_out_group}. By utilizing semantically explicit probes, TriProTesting examines how FMs respond to various demographic groups. For instance, as illustrated in Fig.~\ref{fig:fig1} A, an input image labeled as a ``chef" might produce two distinct outcomes: a correct prediction of ``chef," indicating no bias, or a misclassification as ``criminal," influenced by negative stereotypes associated with certain groups. This approach enables TriProTesting to directly reveal model biases manifesting under specific conditions.

To fully reveal biases within FMs, we design three types of probes: Negative Probes, Positive Probes, and Neutral Probes (Fig.~\ref{fig:fig1} B). This design is motivated by the long-standing transmission of societal and historical biases, which typically appear as explicit or implicit biases. Explicit biases are directly expressed through clear and unfair attitudes or stereotypes towards specific groups \cite{explicit_implicit_bias}. For example, despite its intent to denounce slavery, the novel Uncle Tom's Cabin reinforced discriminatory stereotypes about African Americans; the song ``Some Girls" overtly demeans women of different races, clearly exhibiting racial and gender biases. Implicit biases, on the other hand, are more insidious, subtly embedding stereotypes under the guise of neutrality or flattery, making them harder to detect \cite{implicit_bias2024,implicit_probe_2024}. For instance, the proverb ``Boys will be boys" ostensibly normalizes male behavior while reinforcing submissive roles for females, and the phrase ``You can’t teach an old dog new tricks" perpetuates negative assumptions about the elderly’s capacity to learn. TriProTesting systematically uncovers these biases through its trident-like design: 1) Negative and Positive Probes are employed to detect explicit biases by assessing a model’s tendency to associate groups with negative or positive stereotypes. 2) Neutral Probes are used to reveal implicit biases. For example, if a model accurately predicts one group (e.g.,``man" or ``woman") but frequently classifies the other as the neutral term ``person," this indicates an underlying gender bias.

\subsection{Who is harmed}
Social stratification theory highlights the unequal distribution of resources, power, and status among social groups, a disparity deeply rooted in societal structures and exacerbated by cultural and cognitive biases across dimensions such as gender, age, occupation, and race \cite{social_stratification,Social_identifications,in_out_group}. Specifically, gender biases significantly affect man's and woman's opportunities to access social resources \cite{Gender_Racial_bias}; age biases impact assessments of capabilities and values across different age groups, influencing policy and resource allocation \cite{age_bias}; occupational biases impact societal evaluations and treatment of individuals based on their professions \cite{occupational_bias}; and racial biases, entrenched in history and still active, undermine fairness and inclusivity \cite{Racial_bias}. In this work, we focus on four core dimensions: gender, age, occupational, and racial biases, aiming to reveal the underlying mechanisms of social stratification in FMs and the potential systematic discrimination against specific groups. To identify affected groups within these core dimensions, we design two testing approaches for a comprehensive analysis of biases associated with single and mixed social attributes:

\textbf{Single Bias Test} examines biases in individual social attributes, such as gender or race, by analyzing the manifestation of stereotypes and bias distributions within specific dimensions, thus uncovering potential discriminatory practices in various groups. We selected four representative datasets—CelebA \cite{celeba}, UTKFace \cite{UTKFace}, FairFace \cite{Fairface}, IdenProf \cite{IdenProf} (Fig.~\ref{fig:fig1} C)—which provide detailed labels and diverse group distributions, enabling the analysis of FMs' responses to single social attributes such as ``man", ``chef", ``child", and ``White".

\textbf{Mixed Bias Test} investigates model biases arising from combinations of multiple social attributes, revealing additive effects such as those between gender and occupation, and enabling a more granular analysis. To achieve this, we expand existing datasets by adding gender labels, creating extended versions of the datasets UTKFACE, FAIRFACE, and IDENPROF (distinguished by capitalized names) (Fig.~\ref{fig:fig1} D). For example, IdenProf was reannotated as IDENPROF, adding detailed combination labels such as ``chef\_woman" or ``doctor\_woman." This extension allows us to test whether models exhibit significant biases towards mixed attribute groups. 

Through the Single Bias Test and Mixed Bias Test, we can precisely locate affected groups, including those defined by single and multiple social attributes. Our analysis encompasses four typical FMs (CLIP \cite{CLIP}, ALIGN \cite{ALIGN}, BridgeTower \cite{Bridgetower}, OWLv2 \cite{OWLv2}), four primary datasets (CelebA \cite{celeba}, UTKFace \cite{UTKFace}, FairFace \cite{Fairface}, IdenProf \cite{IdenProf}), and three extended datasets (UTKFACE, FAIRFACE, IDENPROF), engaging in cross-tests with 15 different probes. The testing setup thus includes 240 single bias test scenarios (4 models $\times$ 4 datasets $\times$ 15 probes) and 180 mixed bias test scenarios (4 models $\times$ 3 extended datasets $\times$ 15 probes). These extensive tests allow us to intricately map out who is impacted by biases within FMs, providing data-driven insights into the mechanisms of bias in these models and laying the theoretical groundwork for developing bias mitigation strategies.

\subsection{What can be done}
Social biases typically result in an unfair distribution of resources and social evaluations among different groups. Power redistribution is an effective solution to social inequalities, aiming to enhance the circumstances of vulnerable groups through adjustments in resource allocation and structural power dynamics \cite{Power_redistribution,Inequality_redistribution}. In FMs, biases can be viewed as a ``technological reproduction of inequality," where the overrepresentation of mainstream views in the training corpus leads to their dominance in model outputs, described as ``probability power." This phenomenon manifests as strong correlations between non-discriminated groups and positive probes, alongside excessive associations of discriminated groups with negative or neutral probes.

Interestingly, the principle of logit adjustment used in computer science to address long-tail distribution issues mirrors the fundamental concept of power redistribution \cite{adjust_long,sun2023unbiased,logitadjustment}. Logit adjustment modifies the output logit values with a set of adjustment factors, redistributing probabilities across categories to reduce biases (Fig.~\ref{fig:fig1} E). Building on this concept, we propose Adaptive Logit Adjustment (AdaLogAdjustment), a bias mitigation method that redistributes probability power across categories, balancing model responses to explicit and implicit biases.

Unlike traditional logit adjustment methods for addressing long-tail distribution problems \cite{adjust_long,sun2023unbiased,logitadjustment}, our approach employs an automated learning strategy for adjustment factors, enabling flexible bias mitigation across diverse FMs, datasets, and social attributes (see Method section for details). Compared to conventional training-stage methods that mitigate biases through data rebalancing or model architecture adjustments, our method offers several key advantages:

\textbf{Efficiency}. By analyzing a small set of labeled samples, it adaptively learns adjustment factors, significantly reducing both computational and data demands.

\textbf{Interpretability}. By explicitly controlling the model’s response intensity to probes, the effects of bias mitigation can be directly observed, enhancing both the interpretability and transparency of the results.

In summary, our contributions can be summarized as follows:

\begin{itemize}
    \item We introduce TriProTesting, a systematic method for testing biases in FMs. TriProTesting integrates Single and Mixed Bias Tests to analyze biases across widely examined societal attributes (e.g., gender, age, occupation, race) and their combinations (e.g., gender$\times$age, gender$\times$occupation). Using a trident-like probe design—Negative, Positive, and Neutral Probes—TriProTesting identifies both explicit and implicit biases, providing a unified evaluation benchmark. Through 240 Single Bias Test scenarios and 180 Mixed Bias Test scenarios across four representative FMs (CLIP, ALIGN, BridgeTower, OWLv2), our results reveal pervasive biases in societal attributes and uncover underexplored phenomena, such as contradictory and exaggerated representations and the inheritance of single-attribute biases in mixed scenarios with increased complexity. These findings provide essential guidance for understanding and mitigating the societal impacts of FMs.

    \item We propose AdaLogAdjustment, a scalable and model-agnostic technique for mitigating explicit and implicit biases. AdaLogAdjustment redistributes probability power dynamically in model outputs, requiring no retraining or architectural changes while leveraging minimal labeled data for efficiency and applicability. Evaluated on four datasets (CelebA, UTKFace, FairFace, IdenProf) and three extended datasets (UTKFACE, FAIRFACE, IDENPROF), as well as four representative FMs (CLIP, ALIGN, BridgeTower, OWLv2), AdaLogAdjustment mitigates biases in 99.17\% of 240 Single Bias Test scenarios and 98.89\% of 180 Mixed Bias Test scenarios, demonstrating its generalizability. Additionally, it achieves up to 70\% improvement in bias mitigation performance in certain cases, highlighting its effectiveness.

    \item We draw from the concept of power redistribution in social science, analogizing FM biases as ``technological reproductions of inequality." By integrating this perspective with logit adjustment, we introduce ``probability power redistribution" as a theoretical foundation for understanding and mitigating biases as imbalances in probability distributions. This novel interdisciplinary perspective bridges AI fairness and societal ethics, offering practical tools and conceptual advancements for addressing bias in FMs. Our work provides actionable frameworks for future research on debiasing and enhances the broader dialogue on the ethical responsibilities of AI systems.
\end{itemize}

\begin{figure*}[h!]
\centering
\includegraphics[width=1.0\textwidth]{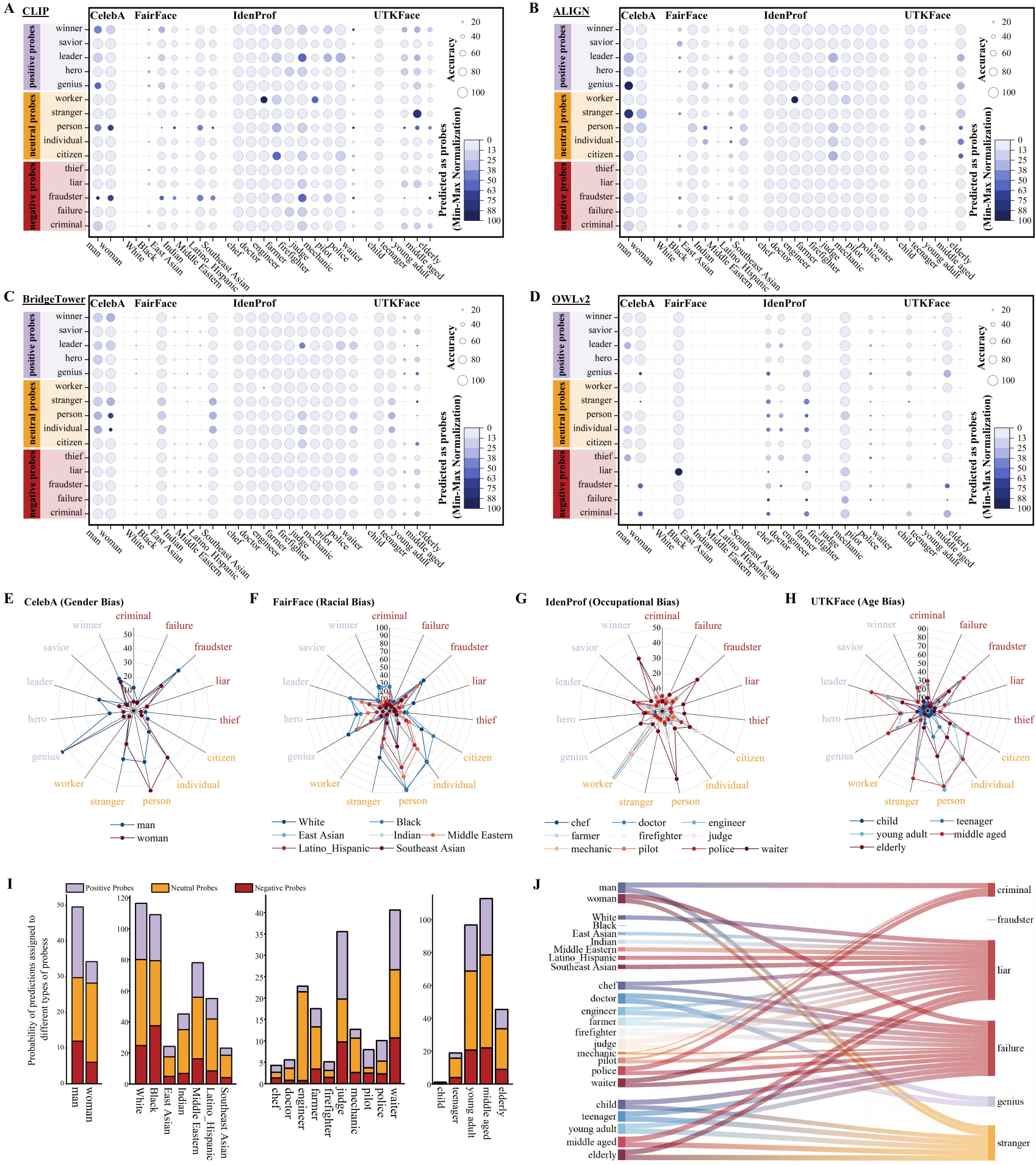} 
\caption{\textbf{Bias analysis of FMs using Single Bias Test.} \textbf{A-D}, FMs' prediction accuracy with probes included (bubbles' size) and the probability of being predicted as a probe (bubbles' color). \textbf{E-H}, The average probabilities of being predicted as probes for three models (CLIP, ALIGN, BridgeTower) across datasets. \textbf{I}, The average probabilities of being predicted as different types of probes for three models (CLIP, ALIGN, BridgeTower) across datasets. \textbf{J}, The top two largest probes predicted by OWLv2 for different classes.}
\label{fig:fig2}
\end{figure*}

\section{Results and Discussion}

In the previous section, we discussed four key questions central to understanding biases in Foundation Models (FMs): Why harmful, How to test, Who is harmed, and What can be done. These questions comprehensively address the societal impacts, testing approaches, affected groups, and mitigation strategies for biases in FMs. In this section, we will provide answers to these questions through our experimental findings. The results of the Single Bias Test and Mixed Bias Test primarily affirm the effectiveness of the Trident Probe Testing (TriProTesting) method. These results confirm the efficacy of our proposed testing method while highlighting specific bias manifestations and the adversely affected groups, further elucidating the societal risks posed by these biases. Subsequently, the section on \textbf{Bias Mitigation with Adaptive Logit Adjustment} will demonstrate how our proposed Adaptive Logit Adjustment (AdaLogAdjustment) method can effectively address biases in FMs.

\begin{figure*}[h!] 
\centering
\includegraphics[width=1.0\textwidth]{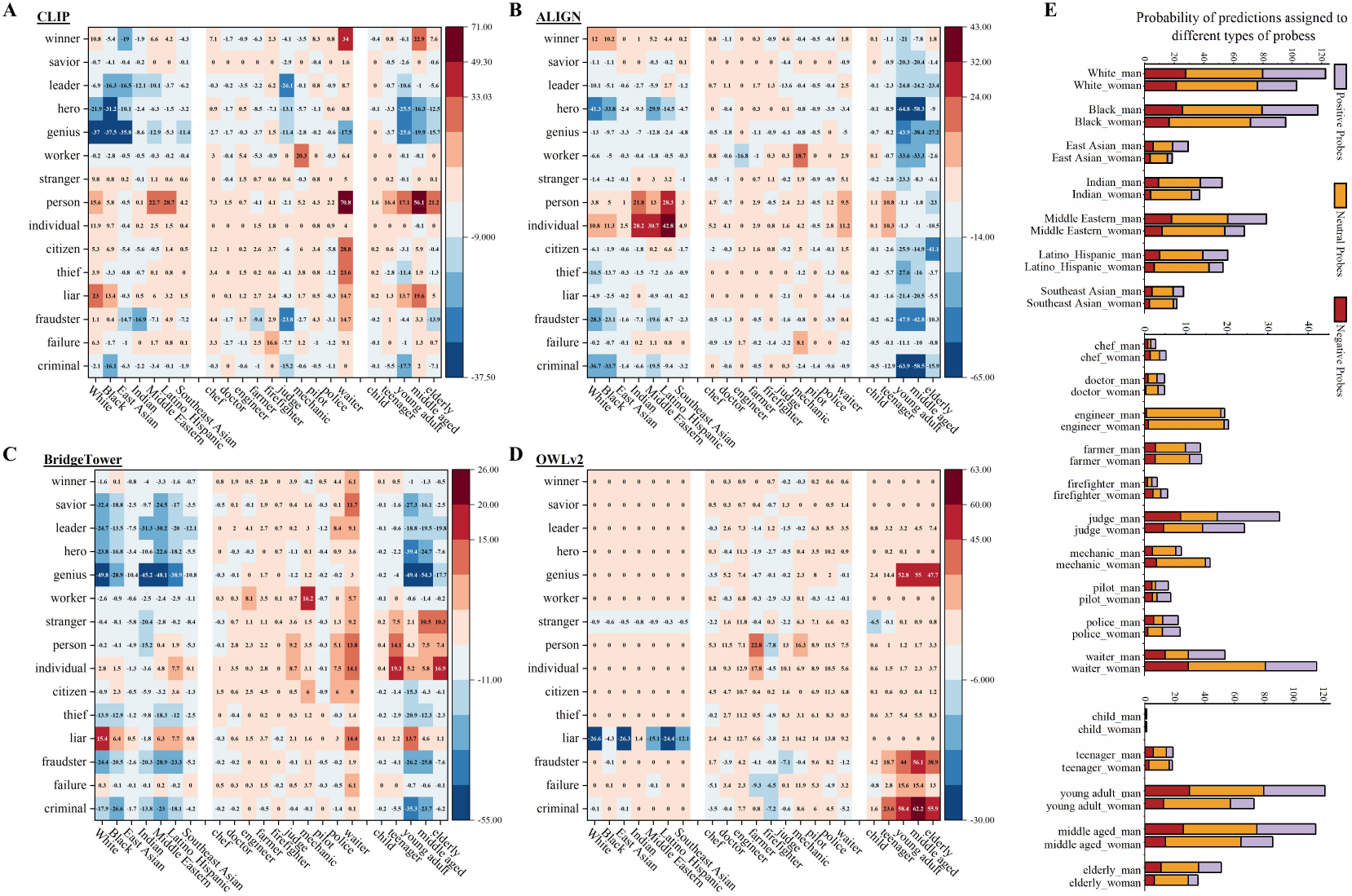} 
\caption{\textbf{Bias analysis of FMs using Mixed Bias Test.} \textbf{A-D}, Mixed bias distributions for four FMs (CLIP, ALIGN, BridgeTower, OWLv2) across three extended datasets (UTKFACE, FAIRFACE, IDENPROF). The heatmap values represent the probability of ``woman" groups being predicted as a probe minus the same for ``man" groups. A positive value indicates that the ``woman" groups is more likely to be predicted as the corresponding probe, while a negative value indicates the opposite. \textbf{E}, Average probabilities of social subgroups (e.g., white\_man, police\_woman) being predicted as different types of probes (Negative, Neutral, Positive) for three models (CLIP, ALIGN, BridgeTower).}
\label{fig:fig3}
\end{figure*}

Before formally analyzing the experimental results, it is essential to note that the data representing ``probability of being predicted as a probe" in Fig.~\ref{fig:fig2} and ~\ref{fig:fig3} have been normalized. During testing, it was observed that the range of bias manifestations varies significantly among different models. Presenting raw probability data without processing could obscure the subtleties of models with narrower ranges of bias, thus affecting the observation of their bias performance. To address this, we applied min-max normalization to each model's data, normalizing values to a range of [0, 100] (unnormalized data are provided in Supplementary S1.1). Specifically, we normalize predicted probabilities for all datasets and probe scenarios in both the Single Bias Test and Mixed Bias Test. For instance, in the Single Bias Test, the normalization range for the CLIP \cite{CLIP} model encompasses its probabilities across all datasets (e.g., CelebA \cite{celeba}, UTKFace \cite{UTKFace}) and probe scenarios (e.g., ``hero," ``criminal"). The same normalization process is applied in the Mixed Bias Test. This technique enhances graph visualization, ensuring that models or datasets with smaller bias values are not overlooked and more vividly reflecting relative bias extents across testing scenarios. It is crucial to emphasize that this normalization is solely for visualization purposes and does not affect the authenticity of the experimental data or the qualitative analysis of model bias.

\subsection{Results of Single Bias Test}

This section systematically analyzes experimental results from four FMs—CLIP \cite{CLIP}, ALIGN \cite{ALIGN}, BridgeTower \cite{Bridgetower}, and OWLv2 \cite{OWLv2}—highlighting biases in single social attributes, including gender, age, occupation, and race. The Single Bias Test is significant not only for uncovering model stereotypes related to individual social attributes but also for establishing a foundation to explore biases arising from combinations of multiple attributes. By conducting systematic tests on individual attributes, we can identify and quantify how social biases are specifically manifested within these models, trace their origins, and offer precise guidance for developing future bias mitigation strategies. Fig.~\ref{fig:fig2} A-D illustrate bias distributions for the four models across four datasets, with the x-axis representing group classifications and the y-axis depicting three probe types. The size of the bubbles in Fig.~\ref{fig:fig2} A-D represents the models' prediction accuracy with probes included, while the color of the bubbles indicates the probability of group classifications being predicted as probes. Thus, bubble size and color reveal the models' sensitivity to different probes. The manifestation of biases across different social attributes is evidenced by significant differences in classification accuracy and tendencies to predict as probes. The results in Fig.~\ref{fig:fig2} A-D highlight the prevalence and pattern differences of biases related to single social attributes. Key observations and conclusions are detailed below:

\textbf{Ubiquity of bias in FMs.} For instance, in the FairFace dataset, the CLIP model exhibits smaller bubbles for the ``White" group (Fig.~\ref{fig:fig2} A), suggesting a comparatively lower recognition ability for this race. ALIGN shows a significantly higher probability of predicting the ``man" group as the probe ``genius" compared to ``woman" (Fig.~\ref{fig:fig2} B), reflecting a positive stereotype towards males. These biases reenact societal biases formed, propagated, and inherited through frameworks such as Labeling Theory and Social Identity Theory, reflecting the internalization of stereotypes by the models and their amplification in outputs \cite{Labeling_Bernburg2009,davis1972labeling,in_out_group,intergroup_bias,Intergroup_emotions_theory}. Technically, these biases highlight inadequacies in training data diversity and the model's reliance on sample distributions, which can exacerbate social inequalities, particularly in decision-sensitive domains.

\textbf{Consistencies in biases across different FMs on certain social attributes.} For example, the bubble patterns of CLIP \cite{CLIP} and ALIGN \cite{ALIGN} in the CelebA \cite{celeba} and FairFace \cite{Fairface} datasets are highly similar (Fig.~\ref{fig:fig2} A vs. Fig.~\ref{fig:fig2} B), indicating that gender- and race-related biases in these models may stem from mainstream values and societal biases reflected in their shared datasets. The gender bias, for instance, strengthens the societal image of ``man" group in sectors such as technology and leadership while overlooking the diverse contributions of ``woman". This consistency also reveals the significant impact of dataset commonality on model biases: stereotypes long-present in the data are learned and internalized by models, thus exhibiting similar bias patterns across different models.

\textbf{Inconsistencies in biases across different FMs on certain social attributes.} For example, CLIP and BridgeTower demonstrate notable differences on social attributes within datasets like UTKFace \cite{UTKFace}(Fig.~\ref{fig:fig2} A vs. Fig.~\ref{fig:fig2} C).  CLIP tends to favor the ``young adult" category, whereas BridgeTower performs better with the ``middle aged" group. These inconsistencies likely arise from differences in training on non-shared data and varying learning objectives. For example, the extent of occupational label coverage in non-shared datasets can significantly influence how models learn and represent these labels. This highlights the substantial influence of unique dataset characteristics on model biases. Additionally, variations in model architecture and training strategies can further accentuate the effects of non-shared data on bias patterns. This underscores the limitations of studying biases using single models and emphasizes the need for a multi-model approach to develop universally applicable bias mitigation strategies \cite{intergroup_bias,Gender_Racial_bias}.

\textbf{Overrepresentation and contradictions of racial and age biases.} In Fig.~\ref{fig:fig2} E-H, we show the distribution of average biases across the models CLIP, ALIGN, and BridgeTower. The radar charts in these figures display different probes on each axis, with points representing the average probability of models predicting a group as the corresponding probe. Notably, the calculations of average probabilities include only CLIP, ALIGN, and BridgeTower, excluding OWLv2 (the reasons are detailed in the Methods section). The results highlight overall societal attribute biases. For the FairFace dataset, the ``White" and ``Black" groups show high activity in TriProTesting, achieving significantly high probabilities across nearly all probe scenarios (Fig.~\ref{fig:fig2} F). Similarly, in the UTKFace dataset, the ``young adult" and ``middle aged" groups exhibit elevated probabilities across probe scenarios (Fig.~\ref{fig:fig2} H). These findings suggest that groups such as ``White", ``Black", ``young adults", and ``middle aged" may be overrepresented in the training data, leading to an amplification of their societal attributes. Moreover, these groups exhibit contradictory biases, with high probabilities for both positive and negative probes, indicating an oversimplification of diverse characteristics into binary categories. Overrepresentation often leads to the underrepresentation of other groups, resulting in systematic neglect or misclassification by the model. For instance, in critical decision-making scenarios, some racial groups may not receive an evaluation as fair as that afforded to ``White" or ``Black" groups. While contradictory biases could also foster dual societal expectations for specific groups, as seen when young adults might be burdened with overly high expectations of capability yet tagged with immaturity, whereas middle aged individuals are perceived as experienced yet conservative. The interplay of overrepresentation and contradictory biases further diminishes the presence of certain groups in model outputs, exacerbating their marginalization in societal applications.

\textbf{FMs exhibit both explicit and implicit biases.} In Fig.~\ref{fig:fig2} I, we further analyze the distribution of average biases from Fig.~\ref{fig:fig2} E-H from the perspective of probe types. Consistent with Fig.~\ref{fig:fig2} E-H, Fig.~\ref{fig:fig2} I also excludes OWLv2, with detailed reasons provided in the \textbf{Method} section. Fig.~\ref{fig:fig2} I specifically presents stacked bar charts, with each bar representing the probabilities of being predicted as Negative, Positive, or Neutral Probes. This format enables the observation of stereotypes reinforced by the models. For example, groups such as ``man", ``judge", and ``police" are more frequently predicted as Positive Probes, reflecting the models' reinforcement of positive societal stereotypes. In contrast, groups like ``woman," ``East Asian," ``engineer," ``farmer," ``teenager," and ``elderly" are predominantly predicted as Negative or Neutral Probes. The high proportion of Neutral Probes particularly reveals implicit biases, indicating an indirect devaluation or disregard for these groups' characteristics or value, even in the absence of explicit negative labeling. This phenomenon aligns with traditional societal attitudes, where ``woman" group, for instance, are often perceived as ``generic" or background characters. Thus, Neutral Probes serve as a crucial tool for capturing implicit biases, complementing the findings with Negative and Positive Probes. This trend reflects the models' internalization of data distributions and societal stereotypes, reinforcing positive perceptions of dominant groups while further marginalizing vulnerable groups and exacerbating social injustices \cite{explicit_implicit_bias,implicit_bias2024,in_out_group}.

\textbf{TriProTesting proves effective in Single Bias Tests applied across various FMs.} As an open vocabulary object detection model, OWLv2 demonstrates relatively weak classification performance across the four datasets (Fig.~\ref{fig:fig2} D), primarily due to its focus on object detection rather than social attribute classification \cite{OWLv2}. As a result, OWLv2 is excluded from the average bias distribution analysis in Fig.~\ref{fig:fig2} E-I. Instead, we conduct a separate analysis of OWLv2’s biases across different social attributes (Fig.~\ref{fig:fig2} J), which reveals the probes most frequently associated with each social group. Notably, OWLv2 exhibits pronounced biases in certain classifications. For example, in the FairFace dataset, the strong association of the ``Black" group with the probe ``liar" indicates a negative stereotype (Fig.~\ref{fig:fig2} D). Additionally, despite overall limited performance, OWLv2 tends to predict ``doctor" and ``judge" categories as ``genius," aligning with positive societal stereotypes about these professions. This indicates that even an object detection model like OWLv2 can inherit longstanding biases from training data. Testing OWLv2 not only uncovers its bias patterns but also serves as a reference for the applicability of open vocabulary object detection models in societal bias tests. This experiment further demonstrates that even models not specifically designed for social attribute analysis inevitably reflect biases consistent with mainstream societal views, influenced by training data and model architecture. Thus, analyzing OWLv2's bias patterns broadens the applicability of our proposed TriProTesting method, offering a practical foundation for more comprehensive bias analysis and mitigation strategies across diverse models.

The Single Bias Test systematically identifies biases toward individual social attributes in FMs. The primary value of this test lies in its broad applicability and ability to facilitate detailed analyses. It evaluates biases in individual social attributes, offering a standardized framework for model comparison and bias origin tracing, while paving the way for exploring mixed social attribute biases.

\subsection{Results of Mixed Bias Test}

This section analyzes experimental results from four FMs, focusing on bias patterns across gender$\times$race, gender$\times$occupation, and gender$\times$age combinations. The Mixed Bias Test, designed to detect systematic biases towards groups characterized by these combined attributes, provides a more detailed perspective compared to tests of single-attribute biases. These tests are essential not only for expanding our understanding of the societal biases inherent in models but also for elucidating the compound effects of these biases.

Fig.~\ref{fig:fig3} A-D depicts mixed bias distributions in three extended datasets (UTKFACE, FAIRFACE, and IDENPROF) using heatmaps. In these heatmaps, the x-axis represents all categories in the datasets, and the y-axis denotes various probes. The heatmap values represent the probability differences of ``woman" groups being predicted as a probe minus the same for ``man" groups. See Supplementary S2.2 for the original values of ``woman" and ``man" groups being predicted as different probes. Therefore, a higher value suggests a greater likelihood of the ``woman" group being predicted as the probe; a lower value indicates a higher likelihood for the ``man" group. Such visualizations allow us to observe the direction and magnitude of biases on mixed social attributes within the target models. Detailed observations and conclusions are as follows:

\textbf{Ubiquity of Mixed Biases in FMs.} Mixed biases are widespread in FMs across the three datasets: UTKFACE, FAIRFACE, and IDENPROF. For instance, in the UTKFACE dataset, the analysis shows that the ``man" group is more likely than the ``woman" group to be predicted as Positive Probes (Fig.~\ref{fig:fig3} A, C). This alignment with the Single Bias Test results, as seen in Fig.~\ref{fig:fig2} A, E, and I, indicates that FMs consistently favor the ``man" group, in both single and mixed attribute analyses. Furthermore, the ALIGN model's performance on the FAIRFACE dataset exemplifies this bias, as the ``man" group predicted more frequently across all probe types than the ``woman" group (Fig.~\ref{fig:fig3} B). These findings underscore the overrepresentation and contradictory biases observed in the Single Bias Test, suggesting that such biases persist in mixed attribute settings and reveal deeper complexities.

\textbf{Consistencies and inconsistencies in mixed biases of FMs.} In the Single Bias Test, we observed that biases across different FMs exhibit both consistency and inconsistency in certain social attributes, a pattern that extends and deepens in the Mixed Bias Test. For example, CLIP and BridgeTower demonstrate high similarity in biases on combined race$\times$gender and age$\times$gender attributes, while significant differences emerge in occupation$\times$gender combinations (Fig.~\ref{fig:fig3} A, C). Such consistencies likely stem from the models' internalization of mainstream values and societal biases embedded in shared datasets, while inconsistencies result from variations in label quality and distribution within non-shared data, skewing the models' learning towards specific social attributes. Notably, the inconsistencies are more pronounced in the Mixed Bias Test (Fig.~\ref{fig:fig3} A-D), suggesting that interactions between two or more attributes may amplify the complexity of bias patterns, such as the joint effects of occupation and gender labels, which pose greater challenges for accurate modeling than single attributes. Moreover, differences in model architecture and training strategies likely magnify these inconsistencies. Compared to the Single Bias Test, the Mixed Bias Test further exposes the complex interplay between ``shared" and ``non-shared" data characteristics, affecting not only overall model performance but also directly influencing predictive biases towards specific groups. Consequently, these observations yield pivotal insights into the complex mechanisms underpinning biases.

\textbf{FMs exhibit compound gender biases across most social attributes.} Fig.~\ref{fig:fig3} E presents a quantitative analysis of the average bias distribution for CLIP, ALIGN, and BridgeTower, categorized by probe types. Similar to Fig.~\ref{fig:fig2} I, Fig.~\ref{fig:fig3} E shows the probability distribution for each mixed attribute group (e.g., ``white\_man", ``police\_woman") being predicted as various types of probes. This reveals notable gender biases; for instance, ``East Asian\_man" is associated with Positive Probes far more frequently than ``East Asian\_woman", and ``waiter\_man" is more commonly linked with Negative Probes than ``waiter\_woman". However, in certain professional categories, such as doctors, gender disparities are minimal, suggesting a relatively equitable approach to gender classification. These findings confirm that while FMs often exhibit pronounced gender biases across most groups, they demonstrate relatively unbiased performance in certain professional fields, such as doctors. Therefore, the Mixed Bias Test not only exposes stereotypes towards groups with mixed attributes but also provides detailed insights into the subtle variations of mixed biases within minority groups.

\textbf{TriProTesting proves effective in Mixed Bias Tests applied across various FMs.} Despite OWLv2's relatively weaker performance in social attribute classification, its biases in mixed attributes, such as gender$\times$age combinations, remain evident. For instance, the ``woman" group is significantly more likely than the ``man" group to be predicted as Negative Probes in the Mixed Bias Test (Fig.~\ref{fig:fig3} D), illustrating distinct biases even in models designed for open vocabulary object detection within specific social contexts. This observation underscores the research value of OWLv2 in Mixed Bias Tests, revealing ingrained stereotypes through an extended bias testing framework, even though the model is not specifically designed for social attribute classification. 

Through the Mixed Bias Test, we observe that biases in FMs involving multiple attributes are both widespread and exhibit cumulative effects, with significant variability across different groups. The test's core value lies in its adaptability to complex social contexts, allowing for systematic assessments of mixed attribute biases, providing detailed analysis to understand the origins of biases, and supporting the development of targeted bias mitigation strategies.

\begin{figure*}[h!] 
\centering
\includegraphics[width=1.0\textwidth]{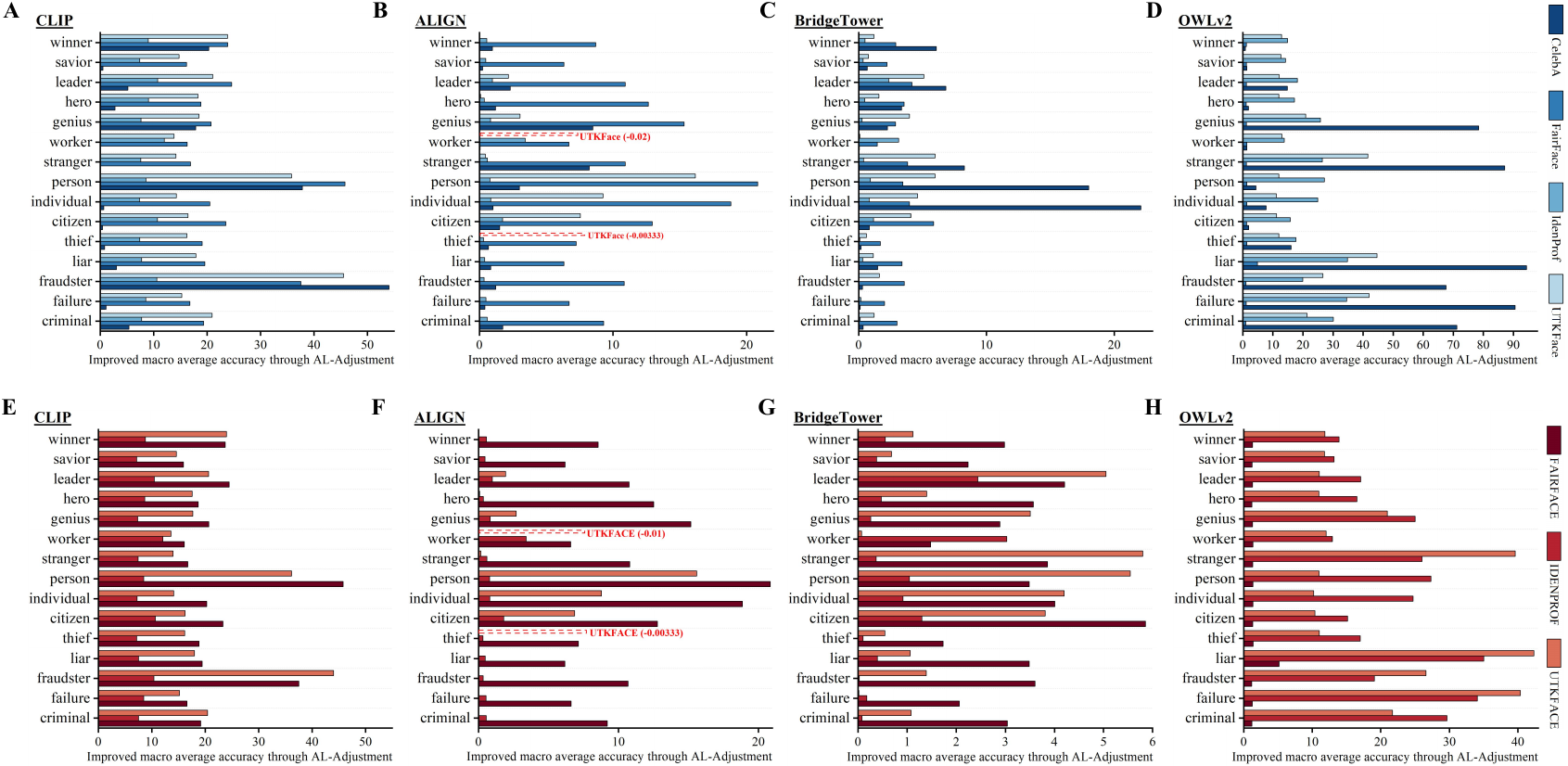} 
\caption{\textbf{Performance improvement of FMs with Adaptive Logit Adjustment (AdaLogAdjustment).}  \textbf{A-D}, Results of Single Bias Test, showing improvements in macro average accuracy for four FMs (CLIP, ALIGN, BridgeTower, OWLv2) across four datasets (CelebA, UTKFace, FairFace, IdenProf). \textbf{E-H}, Results of Mixed Bias Test, showing improvements in macro average accuracy for four FMs across three extended datasets (UTKFACE, FAIRFACE, IDENPROF).}
\label{fig:fig4}
\end{figure*}

\subsection{Bias Mitigation with Adaptive Logit Adjustment}
This section validates the effectiveness of our proposed Adaptive Logit Adjustment (AdaLogAdjustment) in reducing biases and enhancing fairness within FMs. For a detailed explanation of AdaLogAdjustment, refer to the \textbf{Method} section. Fig.~\ref{fig:fig4} reports enhancements in performance across all test scenarios, with the y-axis representing 15 probes and the x-axis showing the ``Improved macro average accuracy through AdaLogAdjustment," indicating the performance improvement of FMs equipped with AdaLogAdjustment compared to a vanilla inference setup. Observations and conclusions based on the results in Fig.~\ref{fig:fig4} are discussed subsequently. The observations and conclusions from Fig.~\ref{fig:fig4} are as follows:

\textbf{Performance enhancement and bias mitigation in Single Bias Test scenarios.} In Fig.~\ref{fig:fig4} A-D, we display test results across multiple datasets for four FMs. AdaLogAdjustment consistently improves macro average accuracy in nearly all TriProTesting scenarios. For example, the CLIP model achieves over a 30\% increase in macro average accuracy across multiple probe tests in the CelebA, UTKFace, and FairFace datasets, including probes like ``fraudster" and ``person" (Fig.~\ref{fig:fig4} A). These results suggest that AdaLogAdjustment effectively mitigates explicit biases towards negative probes and alleviates implicit biases in neutral probe scenarios. Notably, the OWLv2 model, designed for open vocabulary object detection, achieves over a 60\% increase in macro average accuracy across various probe tests in the CelebA dataset, including ``genius", ``stranger", ``liar", ``fraudster", and ``criminal" (Fig.~\ref{fig:fig4} D). Despite the complexity of OWLv2’s bias performance stemming from its adaptability, AdaLogAdjustment exhibits robust bias mitigation capabilities. In the UTKFace dataset, the ALIGN model experiences slight performance drops in the ``genius" and ``thief" probe scenarios, decreasing by 0.02\% and 0.00333\%, respectively (Fig.~\ref{fig:fig4} B). We argue this does not compromise the robustness and stability of our approach, as these decreases represent only 0.8333\% (2/240) of the Single Bias Test scenarios and are minimal. Results in Fig.~\ref{fig:fig4} A-D showcase AdaLogAdjustment's effectiveness in alleviating biases in Single Bias Test scenarios and its adaptability to diverse model architectures and task requirements. This adaptability provides a scalable pathway for enhancing fairness in complex decision systems.

\textbf{Performance enhancement and bias mitigation in Mixed Bias Test scenarios.} In Fig.~\ref{fig:fig4} E-H, we display the test results of four models on the extended datasets UTKFACE, FAIRFACE, and IDENPROF. Consistently, our proposed AdaLogAdjustment substantially elevates performance across nearly all test scenarios. For instance, the ALIGN model shows an increase in macro average accuracy by over 10\% in 9 probe tests in the FAIRFACE dataset (Fig.~\ref{fig:fig4} F), while the BridgeTower model exhibits a more than 20\% improvement in 13 probe tests in the IDENPROF dataset (Fig.~\ref{fig:fig4} H). These results confirm the effectiveness of AdaLogAdjustment in mitigating biases in scenarios involving mixed social attributes, including gender$\times$occupation, gender$\times$age, and gender$\times$race. In mixed attribute scenarios, biases often manifest as compounded effects, leading to more complex stereotypes against specific groups, such as female doctors or Black children. AdaLogAdjustment effectively alleviates these mixed biases by dynamically balancing the probability weight distribution across different mixed attribute groups, thereby ensuring fairness in diverse social scenarios.

Fig.~\ref{fig:fig4} demonstrates the effectiveness of AdaLogAdjustment in alleviating biases across both Single Bias Test and Mixed Bias Test scenarios. Our method therefore provides a scalable technical pathway for enhancing fairness in AI systems, especially in resource-constrained, high-stakes settings such as the healthcare, education, and judiciary sectors. Crucially, AdaLogAdjustment introduces a bias mitigation strategy centered on ``probability power redistribution." This interdisciplinary approach successfully integrates social science theories with AI technical practices, offering vital insights for further research into the fairness of AI systems.

\section{Conclusion}
The findings of this study reveal that biases in Foundation Models (FMs) are both pervasive and multifaceted, manifesting across core social attributes such as gender, age, race, and occupation. By systematically applying Trident Probe Testing (TriProTesting), we have illuminated how explicit and implicit biases are deeply embedded in FMs, stemming from the societal and historical stereotypes encoded in their training data. These biases not only reinforce harmful societal inequalities but also challenge the fairness and reliability of AI systems in critical applications. The proposed Adaptive Logit Adjustment (AdaLogAdjustment) demonstrates a transformative capability to mitigate these biases, dynamically redistributing probability power to achieve balanced predictions. This novel approach introduces a scalable and explainable solution for addressing biases across diverse models, datasets, and social contexts, thereby advancing the fairness and ethical responsibility of AI technologies.

Beyond immediate implications, this work highlights broader challenges in understanding and addressing biases in complex AI systems. While our findings provide a robust framework for bias testing and mitigation, critical questions remain: How can models be designed to intrinsically minimize biases from the outset? What are the long-term societal impacts of deploying debiased models in high-stakes environments like healthcare or education? In our future work, we aim to integrate bias mitigation more seamlessly into model development processes and thoroughly investigate the ethical trade-offs of debiasing techniques in real-world scenarios.

Furthermore, we believe that achieving true fairness in AI requires collective efforts to address not only technical biases in models but also the societal structures that perpetuate inequality. By fostering interdisciplinary collaboration and prioritizing ethical AI practices, the research community can create technologies that are both powerful and equitable.

\section{Method}
\subsection{Data Preparation}

In this study, we select four datasets—CelebA \cite{celeba}, UTKFace \cite{UTKFace}, FairFace \cite{Fairface}, and IdenProf \cite{IdenProf} (Fig.~\ref{fig:fig1} C)—covering core social attributes such as gender, age, race, and occupation for the Single Bias Test. For the Mixed Bias Test, we expanded three of these datasets by adding gender labels, creating extended versions: UTKFACE, FAIRFACE, and IDENPROF (Fig.~\ref{fig:fig1} D).

Specifically, CelebA is a facial dataset with gender labels, utilized for assessing gender bias. FairFace, centered on racial annotations, serves to evaluate racial biases. IdenProf, an occupational classification dataset, is employed for assessing occupational biases. Notably, UTKFace is a large-scale face dataset annotated with continuous age values. However, directly using continuous age annotations may fail to effectively distinguish model biases across age groups. To address this, we resegment the dataset into five categories: children (0-12 years), teenagers (13-19 years), young adults (20-35 years), middle aged (36-60 years), and elderly (61+ years). This segmentation more clearly exposes stereotypes at typical age stages and aids in identifying potential high-risk age groups in model predictions.

To extend UTKFace, FairFace, and IdenProf for mixed bias testing, we employ the CLIP model (ViT-B/32 \cite{VIT}) to automatically generate gender labels. The process involves feeding preprocessed images into the model alongside two text prompts: [``a photo of a man", ``a photo of a woman"]. The model classifies each image as either ``man" or ``woman," which is then recorded as the extended label. This automated labeling process reduces the need for costly manual annotations while ensuring consistent labeling. Utilizing this method, we develop extended datasets (Fig.~\ref{fig:fig1} D): UTKFACE is categorized by age and gender into ten composite labels, such as ``elderly\_woman"; FAIRFACE is divided by race and gender into fourteen labels, such as ``Indian\_woman";and DENPROF is segmented by occupation and gender into twenty labels, such as such as ``doctor\_woman." This expansion captures more complex combinations of social attributes, thereby supporting Mixed Bias Tests. Additionally, this process demonstrates the practical utility of Foundation Models (FMs) in real-world data annotation tasks, as highlighted in many current studies.

\subsection{Probes Design and TriProTesting}

In an effort to fully uncover the complex nature of biases within FMs, our research strategically designs probes to guarantee that test outcomes are scientifically accurate, targeted, and socially significant. The design is governed by two fundamental principles: the systematic categorization of probes and the selection of representative probes, both of which are intended to provide a detailed portrayal of both explicit and implicit biases and support a comprehensive bias analysis \cite{social_stratification,Social_identifications,implicit_bias2024,explicit_implicit_bias}.

We design three types of probes: Negative Probes, Positive Probes, and Neutral Probes (Fig.~\ref{fig:fig1} B), aimed at distinguishing between explicit and implicit biases. Explicit biases manifest as direct and clear unfair attitudes or stereotypes toward certain groups. To detect such biases, we design Negative and Positive Probes. Implicit biases, however, are subtler and often conveyed through neutral or seemingly positive expressions. Neutral Probes are thus crafted to analyze whether models tend to classify specific groups into neutral categories rather than matching their actual labels. This design facilitates not only the revelation of the extent and direction of explicit biases but also the capture of the underlying effects of implicit biases, providing robust support for a comprehensive portrayal of biases.

In selecting specific probes, we place particular emphasis on each probe's societal representativeness and multidimensional coverage. Our criteria for selection include: 1) Occupational and identity roles, such as ``worker," ``citizen," ``stranger," ``hero," and ``leader," to probe models' biases towards various societal roles; 2) Integrity and moral traits, including ``liar," ``fraudster," and ``criminal," to determine if models attribute negative moral labels to certain groups; 3) Social status and competency traits, like ``savior," ``genius," ``failure," and ``individual," to investigate stereotypes concerning social status and capabilities; 4) Criminality and failure associations, with terms like ``thief," ``criminal," ``failure," revealing models' bias in associating negative actions and statuses; 5) Generic identity and interpersonal relations, with probes like ``person," ``citizen," ``individual," focusing on models' predictions of neutral identities, crucial for identifying implicit biases. This multi-perspective selection mechanism allows for a systematic evaluation of biases across multifaceted social attributes, lending scientific rigor and societal relevance to our probe tests.

Building upon this comprehensive probe design, we propose the Trident Probe Testing (TriProTesting) method, a bias testing method with a trident-like structure that incorporates Negative, Positive, and Neutral Probes. TriProTesting highlights three-pronged design, each probe type serving a distinct function in identifying explicit and implicit biases. Negative and Positive Probes assess models’ inclination to associate specific groups with overtly negative or positive stereotypes, respectively. Neutral Probes examine the subtler tendencies of models to classify groups into non-descriptive or neutral categories, capturing implicit biases that might otherwise go unnoticed. TriProTesting provides a holistic evaluation of model biases by integrating these probe types, offering a systematic and interpretable testing framework. 

To operationalize TriProTesting, we conduct Single Bias Tests and Mixed Bias Tests, designed to reveal biases associated with individual and combined social attributes. Our analysis spans four representative FMs (CLIP \cite{CLIP}, ALIGN \cite{ALIGN}, BridgeTower \cite{Bridgetower}, OWLv2 \cite{OWLv2}), four primary datasets (CelebA \cite{celeba}, UTKFace \cite{UTKFace}, FairFace \cite{Fairface}, IdenProf \cite{IdenProf}), and three extended datasets (UTKFACE, FAIRFACE, IDENPROF). The testing setup incorporates 240 Single Bias Test scenarios (4 models $\times$ 4 datasets $\times$ 15 probes) and 180 Mixed Bias Test scenarios (4 models $\times$ 3 extended datasets $\times$ 15 probes). By systematically cross-testing all combinations of models, datasets, and probes, TriProTesting uncovers nuanced patterns of bias, such as contradictory or exaggerated representations in certain attributes, and reveals how biases in mixed attributes inherit and compound those observed in single attributes.

\subsection{Models Tested}

In this study, we test biases in four representative FMs: CLIP \cite{CLIP}, ALIGN \cite{ALIGN}, BridgeTower \cite{Bridgetower}, and a modified OWLv2 \cite{OWLv2}.

CLIP, a pioneering multimodal alignment model, employs contrastive learning to establish shared embedding spaces for images and text, excelling in zero-shot classification and cross-modal retrieval, and serves as an early advocate of prompt engineering. ALIGN advances this capability by employing weakly supervised learning on an expansive dataset of 1.8 billion image-text pairs, demonstrating the potential of big data to enhance model generalizability while also raising concerns about the complex biases embedded in noisy training data.  BridgeTower's innovative ``Bridge Layer" integrates single-modal encoding with multimodal interactions, showing promising results in tasks like visual question answering and multimodal retrieval. By probing BridgeTower, we explore how model architectures might amplify or mitigate biases during multimodal semantic integration. Finally, OWLv2, initially crafted for open-vocabulary object detection, is reconfigured here to support zero-shot classification by calculating softmax scores from integrated logit values across prompts, thereby aligning with bias testing needs by predicting entire image categories (see Supplementary S2.3 for details on the OWLv2 adaptation). This setup enables OWLv2 to contribute uniquely to understanding how biases transfer across different tasks, revealing the adaptability and bias nuances of multimodal models.

By integrating CLIP, ALIGN, BridgeTower, and OWLv2, this study not only encompasses the classic technological trajectories of FMs but also reflects the diversity and complexity of current research. Probing these models allows us to uncover the mechanisms of bias formation and propagation, providing guidelines for the ethical design and practice of future multimodal models. Additionally, the experimental results offer insights into the similarities and differences in bias manifestations among these models, laying the groundwork for assessing the societal impact of multimodal research.

\subsection{Evaluation Metrics}

\textbf{Overall accuracy.} To quantitatively evaluate model performance in bias tests, we present extensive experimental results in Fig.~\ref{fig:fig2}, ~\ref{fig:fig3}, and ~\ref{fig:fig4}. This section details the calculation methods for the metrics and their implications. Overall accuracy evaluates a model's classification performance across all samples in a single probe test, representing the aggregate prediction accuracy across categories. The formula for calculating overall accuracy is:
\begin{equation}
\text { Accuracy }=\frac{N_{\text {correct }}}{N_{\text {total }}},
\end{equation}
where $N_{\text {total }}$ and $N_{\text {correct }}$ denote the number of samples and the number of correctly predicted samples in a probe test scenario, respectively. For instance, in the bubble charts of Fig.~\ref{fig:fig2} A-D, the size of the bubbles indicates overall accuracy, visually reflecting the model's classification performance across various probe test scenarios.

\textbf{Probability of being predicted as a probe.} The probability of a category being predicted as a probe is a critical metric for assessing how often a model identifies a specific category with a given probe in probe tests. This metric illuminates the degree of bias a model exhibits towards specific categories in various probe scenarios. The calculation formula is: 
\begin{equation}
P(\text { class } \rightarrow \text { probe })=\frac{N_{\text {class }, \text { probe }}}{N_{\text {class }}},
\end{equation}
where $P(\text { class } \rightarrow \text { probe })$ represents the probability of a category being predicted as a particular probe, with $N_{\text {class }}$ denoting the total number of samples for that category and $N_{\text {class }, \text { probe }}$ the number of samples predicted as the probe. In Fig.~\ref{fig:fig2} A-D, the intensity of the bubble's color illustrates the probability of predicting the probe for a category, visually displaying the model's bias performance in TriProTesting scenarios.

\textbf{Min-Max Normalization.} To prevent minor bias discrepancies from being overlooked in visual representations, we normalize the probabilities of probe predictions. The normalization formula is: 
\begin{equation}
P^{\prime}=100 \times \frac{P-P_{\min }}{P_{\max }-P_{\min }},
\end{equation}
where $P$ denotes the original prediction probability, and $P_{\min }$ and $P_{\max }$ represent the minimum and maximum values within these probabilities, respectively. This normalization, applied to each model, standardizes the results for both Single Bias Test and Mixed Bias Test scenarios. Such normalization enhances the visualization in charts, allowing biases in models with smaller ranges to be perceptibly examined while preserving the authenticity of experimental data and the reliability of qualitative bias analysis. In Fig.~\ref{fig:fig2} and ~\ref{fig:fig3}, all visualizations of ``probability of being predicted as a probe" are based on these normalized outcomes.

\textbf{Macro average accuracy.} Macro average accuracy is utilized to evaluate the overall performance of models in multi-class tasks. Unlike weighted average accuracy, macro average accuracy assigns equal weight to each category, providing a fairer evaluation of model performance across diverse classes, particularly in scenarios with imbalanced class distributions. Macro average accuracy is calculated as:
\begin{equation}
\text { Macro Average Accuracy }=\frac{1}{C} \sum_{i=1}^C \frac{N_{i, \text { correct }}}{N_i},
\end{equation}
where $C$ is the total count of categories, with $N_i$ and $N_{i, \text { correct }}$ respectively represent the total number of samples and the number of correctly predicted samples for the $i$-th class. In Fig.~\ref{fig:fig4}, macro average accuracy serves as a key metric to quantify the enhancements brought by the AdaLogAdjustment method in mitigating biases in FMs. The selection of macro average accuracy aligns with the study’s focus on fairness and bias mitigation, as it minimizes evaluation bias caused by unequal data distribution among classes, thereby providing a more accurate reflection of improvements in bias mitigation across all groups.

\subsection{Adaptive Logit Adjustment}
Addressing pervasive biases in FMs is a current focus \cite{navigli2023biases,gallegos2024bias}, with typical mitigation strategies falling into four categories: Pre-Processing, In-Training, Intra-Processing, and Post-Processing Mitigation. Pre-Processing Mitigation improves the training dataset’s representativeness and diversity through data augmentation, reweighting, or generating new data \cite{Pre_Processingzayed2023deep,Pre_Process_ghanbarzadeh2023gender}. In-Training Mitigation integrates fairness mechanisms by modifying model architectures, incorporating new optimization objectives, or selectively updating parameters \cite{In_Train_yang2023adept,In_Train_woo2023compensatory}. Intra-Processing Mitigation adjusts decision-making processes during application, such as modifying decoding strategies or adjusting probability outputs to reduce bias \cite{Intra_Process_liu2023pre,Intra_Process_schramowski2022large}. Post-Processing Mitigation directly eliminates manifestations of bias by altering model outputs, such as rewriting texts or adjusting classifications \cite{Post-Process_zhang2024debiasing,Post-Proces_tanjim2024discovering}. Although these approaches offer advantages in mitigating model biases, they often depend on extensive data annotation or complex structural modifications, which limit their scalability in practical applications.

To address these issues, we propose Adaptive Logit Adjustment (AdaLogAdjustment), a post-processing technique that dynamically redistributes the logit values by learning a set of adjustment factors. Pseudocode for AdaLogAdjustment is provided in Supplementary S2.4 for clarity on the method's details. The method draws on the social science concept of power redistribution, which seeks to improve the situation of disadvantaged groups through resource redistribution or structural adjustments. Similarly, in FMs, biases can be seen as a technological reproduction of inequality, where dominant groups' perspectives are overrepresented in training data, holding disproportionate ``probability power." AdaLogAdjustment redistributes this probability power across categories by adjusting the distribution of logits, ensuring a more balanced representation between explicit and implicit biases. Compared to existing debias methods, AdaLogAdjustment offers distinct advantages: 1) Efficiency: It mitigates biases without requiring model retraining or structural modifications, relying only on minimal labeled data and straightforward optimization; 2) Universality: It is adaptable to any model, dataset, and social attribute in bias testing scenarios; 3) Explainability: The explicit control over the adjustment factors enhances the transparency of bias mitigation outcomes.

Specifically, let the logit values output by the FMs are $\mathbf{z}=[z_1, z_2,..., z_C]$, where $C$ denotes the total number of classes. The logit adjustment process is defined as $\mathbf{z}_{\boldsymbol{\alpha}}=\mathbf{z} \cdot \boldsymbol{\alpha}$, where $\boldsymbol{\alpha}=[\alpha_1, \alpha_2,..., \alpha_C ]$ are adjustment factors, and $\mathbf{z} \cdot \boldsymbol{\alpha}$ denotes element-wise weighted adjustment of the vector of logit values. In tests incorporating probes, for $C$-class tasks, the logit values output by FMs and the required adjustment factors are respectively extended to $\mathbf{z}^{\prime}=[z_1, z_2,..., z_C, z_{C+1}]$ and $\boldsymbol{\alpha}^{\prime}=[\alpha_1, \alpha_2,..., \alpha_C, \alpha_{C+1} ]$. To learn the adjustment factors $\boldsymbol{\alpha}^{\prime}$, a training set is compiled by randomly selecting N (with $N=20$ in this study) samples from each class, while the rest serve as the test set. Refer to Supplementary S2.1 for a detailed analysis of the sample size $N$. The initial adjustment factors $\boldsymbol{\alpha}^{\prime}$ are set to $1$, i.e., $\boldsymbol{\alpha}^{\prime}=[1, 1,..., 1 ]$. Through iterative optimization using the Adam optimizer with a learning rate of $0.01$ (see Supplementary S2.2 for the learning rate ablation study), we minimize the average cross-entropy loss over the training set. The loss function $\mathcal{L}$ is calculated as:
\begin{equation}
\mathcal{L}=-\frac{1}{N \times C} \sum_{n=1}^{N \times C} \log (\frac{\exp (\alpha_{y_n} \cdot z_{y_n})}{\sum_{c=1}^{C+1} \exp (\alpha_c \cdot z_c)}),
\end{equation}
where $n$ indexes the samples in the training set and $y_n$ is the true label of the $n$-th sample.  The optimization runs for $20$ epochs, and the $\boldsymbol{\alpha}^{\prime}$ values that maximize the training set accuracy are selected as the final parameters.

By dynamically adjusting the model's responses to different categories, AdaLogAdjustment effectively weakens the strong associations between dominant groups and positive probes while mitigating the biases of marginalized groups being linked to negative or neutral probes. Importantly, the finding that $N=20$ achieves satisfactory performance underscores the scalability of our approach. It shows that even in scenarios with limited labeled data—a common constraint in real-world applications—our method remains highly effective. This positions AdaLogAdjustment as a practical and versatile solution for addressing biases in diverse settings, contributing to the broader goal of creating fair and reliable AI systems with minimal resource requirements.

\newpage
\begin{suppinfo}
This supplementary material includes sections for both ``Supplementary for `Results and Discussion'" and ``Supplementary for `Method'". The section ``Supplementary for `Results and Discussion'" contains details on the original data before normalization and detailed data for derived calculations. The section ``Supplementary for `Method'" features an ablation study on sample size, an ablation study on learning rate, implementation details of the OWLv2 adaptation, and pseudocode for the TriProTesting and AdaLogAdjustment methods.
\end{suppinfo}

\section*{Technology Use Disclosure}
This manuscript was refined using the AI language model ChatGPT, which assisted in improving the grammar and eliminating typographical errors. The content has been thoroughly reviewed, revised, and endorsed by all contributing authors.

\section*{Data Availability Statement}
Access to the datasets used in this study is provided through the following links:
\begin{itemize}
    \item CelebA dataset is available at: \url{https://mmlab.ie.cuhk.edu.hk/projects/CelebA.html}.
    \item UTKFace dataset can be accessed from: \url{https://susanqq.github.io/UTKFace/}.
    \item FairFace dataset is provided via: \url{https://github.com/joojs/fairface}.
    \item IdenProf dataset can be downloaded from: \url{https://github.com/OlafenwaMoses/IdenProf}.
\end{itemize}
Each dataset is formatted for compatibility with our training framework and available for research purposes.

\section*{Code Availability Statement}
The code used in this study is available on GitHub at \url{https://github.com/rsdczhs0/TriProTesting}.

\begin{acknowledgement}

This work was supported by the National Natural Science Foundation of China under Grant No. 62376283, the Key Stone Grant (JS2023-03) provided by the National University of Defense Technology (NUDT), and the Academy of Finland under Grant No. 331883, the Infotech Project FRAGES. The authors are grateful for the generous support from these institutions.

\end{acknowledgement}

\section*{Author declarations}
\subsection*{Conflict of Interest}
The authors have no conflicts to disclose.

\subsection*{Author Contributions}
Shuzhou Sun, Li Liu, and Janne Heikkilä designed the research study. Shuzhou Sun, Li Liu, Yongxiang Liu, and Zhen Liu developed the method, wrote the code, and performed the analysis. Li Liu, Yongxiang Liu, Shuanghui Zhang and Xiang Li contributed to refining the methods and analytical framework. All authors wrote and approved the manuscript.

\bibliography{reference}
\renewcommand{\thetable}{S\arabic{table}}
\setcounter{table}{0}
\renewcommand{\thefigure}{S\arabic{figure}}
\setcounter{figure}{0}
\newpage
\section*{Supplementary}
\setcounter{section}{0} 
\renewcommand{\thesection}{S\arabic{section}} 

Note that, in this Supplementary, any figure references formatted as ``Fig. 1," ``Fig. 2," and so on correspond to figures presented in the main paper. Figures and tables introduced specifically within this Supplementary are labeled with an ``S" prefix (e.g., Fig. S1, Table S1) to distinguish them from those in the main paper.

\section{Supplementary for ``Results and Discussion"}

This section presents additional data and detailed calculations to supplement the analyses in the main paper's ``Results and Discussion". By including raw data before normalization and intermediate data used in derived calculations, we aim to enhance the transparency, reproducibility, and interpretability of our findings. The supplementary data not only provides insights into the baseline characteristics of the models but also substantiates the computational processes behind key visualizations.

\subsection{Original Data Before Normalization}
This subsection provides unnormalized data underlying several key visualizations in the main paper, specifically Fig. 2 A-D and Fig. 3 A-D. Presenting these raw probabilities enhances the reproducibility and transparency of our analyses. The raw data allows readers to understand the impact of normalization on bias representation, particularly in visualizations relying on probability distributions.

\textbf{Raw probabilities for ``predicted as probes" in Fig. 2 A-D (Table~\ref{tab:probes_raw}).}
Table~\ref{tab:probes_raw} presents the original probabilities of each class predicted as specific probes under the TriProTesting framework. These probabilities, computed for four models (CLIP, ALIGN, BridgeTower, OWLv2) across datasets (CelebA, UTKFace, FairFace, IdenProf), represent the unprocessed likelihoods of class-probe associations. This table underpins the bubble colors in Fig. 2 A-D. While normalized values are used in the main paper to improve visual clarity and comparability, Table~\ref{tab:probes_raw} provides the foundational data, offering a direct insight into the models' raw predictions.

\textbf{Original probabilities supporting heatmap values in Fig. 3 A-D (Table~\ref{tab:woman_raw}, Table~\ref{tab:man_raw}).}
The heatmap values presented in Fig. 3 A-D of the main paper are based on differences in normalized probabilities for “woman” and “man” groups being predicted as probes. In this subsection, we provide the unnormalized probabilities that forming the basis of these calculations. Table~\ref{tab:woman_raw} contains the original probabilities for the “woman” group, while Table~\ref{tab:man_raw} contains the original probabilities for the “man” group across the extended datasets (UTKFACE, FAIRFACE, IDENPROF). However, these unnormalized probabilities serve as foundational inputs and require the normalization process detailed in Subsection S1.2 to produce the heatmap values.

\subsection{Detailed Data for Derived Calculations}

This subsection provides intermediate data supporting the derived calculations presented in the main paper, particularly for Fig. 3 A-D and Fig. 4 A-H. These data ensure transparency in the computation of derived metrics, allowing readers to reconstruct and verify the results.

\textbf{Normalized probabilities used to calculate heatmap values in Fig. 3 A-D (Table~\ref{tab:woman_norm}, Table~\ref{tab:man_norm}).}
The heatmaps in Fig. 3 A-D of the main paper were derived from differences in normalized probabilities of the ``woman" and ``man" groups predicted as probes. Specifically, each heatmap value is calculated by subtracting the normalized probability of the “man” group (Table~\ref{tab:man_norm}) from that of the “woman” group (Table~\ref{tab:woman_norm}). This subsection presents the normalized probabilities used in these calculations. Table~\ref{tab:woman_norm} reports the normalized probabilities for the ``woman" group, while Table~\ref{tab:man_norm} lists the normalized probabilities for the ``man" group across the extended datasets (UTKFACE, FAIRFACE, IDENPROF). These tables ensure transparency in the derivation of the heatmap values and clarify the relationship between the normalized probabilities and the resulting visualizations.

\textbf{\textbf{Macro average accuracy for bias mitigation in Fig. 4 A-D (Table~\ref{tab:single_al}, Table~\ref{tab:single_noal}) and E-H (Table~\ref{tab:mixed_al}, Table~\ref{tab:mixed_noal}).}} Tables~\ref{tab:single_al},~\ref{tab:single_noal},~\ref{tab:mixed_al},~\ref{tab:mixed_noal} present the macro average accuracies for models with and without AdaLogAdjustment across Single Bias Test and Mixed Bias Test scenarios.

Table~\ref{tab:single_al} lists the macro average accuracy achieved with AdaLogAdjustment in Single Bias Test scenarios, while Table~\ref{tab:single_noal} presents the corresponding results without AdaLogAdjustment. Subtracting the values in Table~\ref{tab:single_noal} from Table~\ref{tab:single_al} yields the ``Improved macro average accuracy" values shown in Fig. 4 A-D, quantifying the impact of AdaLogAdjustment on fairness and bias mitigation in Single Bias Test scenarios.

Table~\ref{tab:mixed_al} presents the macro average accuracy results achieved with the application of AdaLogAdjustment in Mixed Bias Test scenarios, whereas Table~\ref{tab:mixed_noal} lists the corresponding results without AdaLogAdjustment. Subtracting the values in Table~\ref{tab:mixed_noal} from Table~\ref{tab:mixed_al} generates the “Improved macro average accuracy” values shown in Fig. 4 E-H, demonstrating the effectiveness of AdaLogAdjustment in mitigating biases in Mixed Bias Test scenarios.

This structured reporting ensures transparency and provides a comprehensive understanding of the improvements facilitated by AdaLogAdjustment in both Single Bias Test and Mixed Bias Test scenarios. 

\section{Supplementary for ``Method"}

This section provides detailed explanations and additional materials to elucidate the methods employed in the main paper. By presenting ablation studies, implementation specifics, and pseudocode, we aim to enhance the transparency, reproducibility, and practicality of our proposed frameworks. These supplementary materials highlight the robustness of our methods under varying conditions, provide insights into critical design choices, and offer clear guidance for replicating and extending the work.

\subsection{Ablation Study on Sample Size}

The parameter $N$ in our method determines the number of samples randomly selected from each class to construct the training set for learning the adjustment factors $\boldsymbol{\alpha}^{\prime}$. These adjustment factors are critical for redistributing the probability power of logits, as described in the main paper. In our main experiments, we set a small sampling size (N=20) to balance computational efficiency and performance. To evaluate the robustness of our method and the impact of sample size on bias mitigation, we conducted an ablation study with $N$ set to $10$, $20$, $30$, $40$, $100$, and $200$. The results, averaged over three runs to account for randomness in sample selection, are reported in Table~\ref{tab:ablation_N}. We have the following observations:

\textbf{Optimal performance at $N=20$.} Across most probe testing scenarios, $N=20$ achieves the best results in improving macro average accuracy through AdaLogAdjustment. This suggests that $N=20$ strikes the optimal balance between sample size and the ability to generalize adjustment factors effectively. Notably, in some scenarios, the superiority of $N=20$ is particularly pronounced. For instance, in the CLIP model tested with the CelebA dataset using ``person" as the probe, $N=20$ improves macro average accuracy by $37.81\%$, significantly outperforming other sample sizes, which achieve improvements of approximately $25\%$.

\textbf{Decreased performance with small $N$.} When $N$ is set to a small value (e.g., $N=10$), the performance generally decreases. This is likely due to insufficient diversity in the training set, as smaller sample sizes may fail to capture the variability in the data distribution. As a result, the learned adjustment factors are less effective at mitigating biases across diverse groups and probes.

\textbf{Decreased performance with large $N$.} Interestingly, increasing $N$ to large values (e.g., $N=100$ or $N=200$), also results in a decline in performance. This decline is attributed to overfitting during optimization, where excessive training samples cause the adjustment factors to overly align with the specific characteristics of the training set, reducing their generalizability to the test set. Moreover, larger $N$ values introduce greater computational costs without proportional gains in performance, making them less practical.

The ablation study demonstrates that $N=20$ is the optimal choice, yielding superior results across most scenarios. This highlights the efficiency of our method in using minimal training data to achieve high-quality bias mitigation and underscores the robustness of the adjustment factors learned in this setting. Setting $N=20$ allows our method to achieve a delicate balance between performance and resource efficiency, making it particularly suitable for practical applications where data availability and computational resources are constrained.

\subsection{Ablation Study on Learning Rate}

\begin{figure*}[h!] 
\centering
\includegraphics[width=0.5\textwidth]{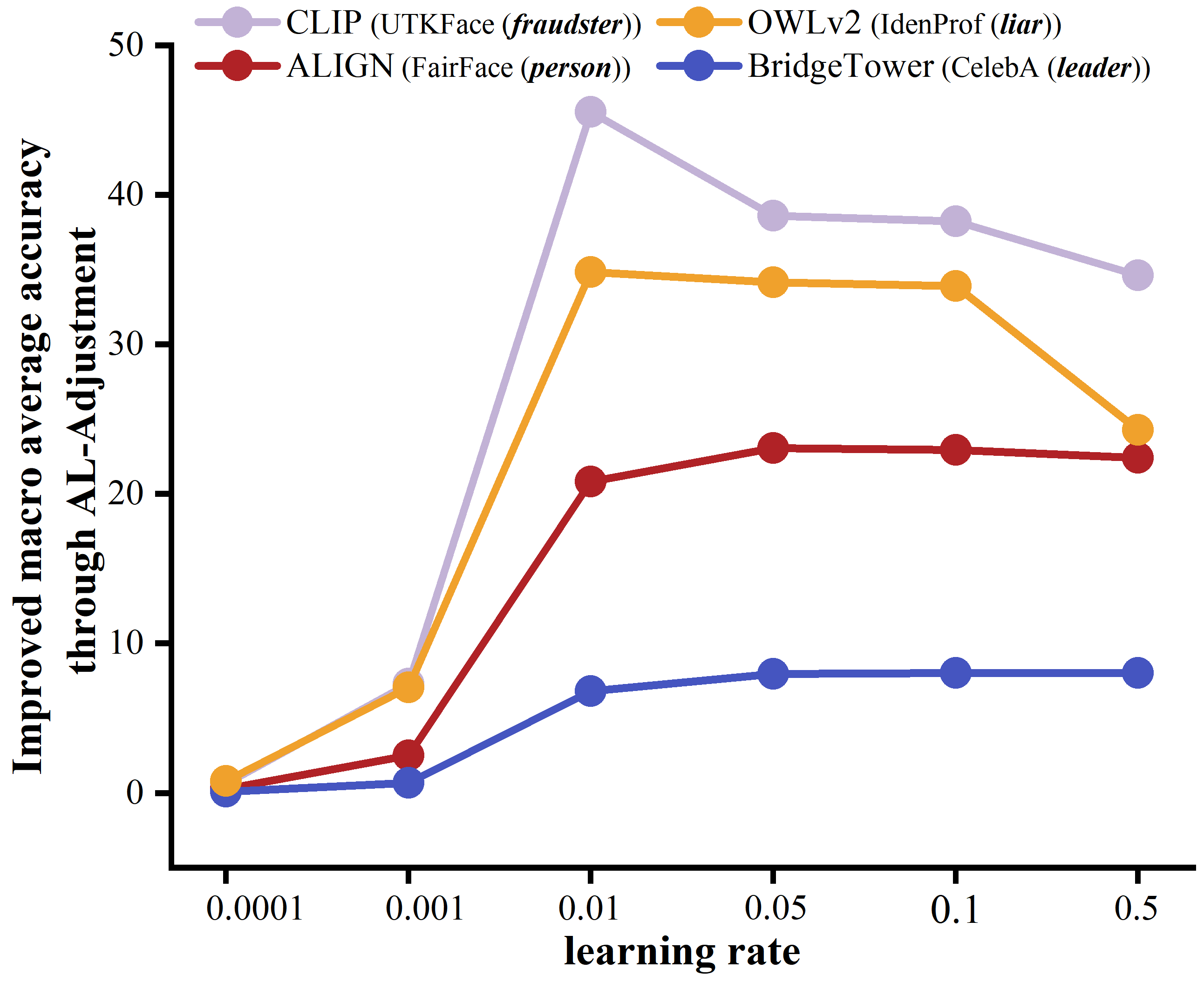} 
\caption{\textbf{Improved macro average accuracy through AdaLogAdjustment under different learning rates.} Improved macro average accuracy is shown for four representative test scenarios: 1) testing UTKFace with the probe ``fraudster" using CLIP, 2) testing FairFace with the probe ``person" using ALIGN, 3) testing CelebA with the probe ``leader" using BridgeTower, and 4) testing IdenProf with the probe ``liar" using OWLv2. Learning rates tested include $0.0001$, $0.001$, $0.01$, $0.05$, $0.1$, and $0.5$.}
\label{fig:ablation_lr}
\end{figure*}

The choice of learning rate is critical in the optimization process for learning logit adjustment factors in the AdaLogAdjustment framework. To evaluate the impact of different learning rates, we conducted ablation studies across four representative test scenarios: testing CLIP on UTKFace with the probe ``fraudster," ALIGN on FairFace with the probe ``person," BridgeTower on CelebA with the probe ``leader," and OWLv2 on IdenProf with the probe ``liar." These scenarios were selected to ensure comprehensive coverage of all models, datasets, and probe types.

We test six different learning rates: $0.0001$, $0.001$, $0.01$, $0.05$, $0.1$, and $0.5$. The results, reported in Fig.~\ref{fig:ablation_lr}, show the improved macro average accuracy achieved through AdaLogAdjustment across these learning rates for each test scenario. From these experiments, we can derive the following key findings:

\textbf{Optimal learning rates vary across scenarios.} Each test scenario exhibited a distinct optimal learning rate. For instance, the optimal learning rate for testing CLIP on UTKFace with the probe ``fraudster" was $0.01$, significantly outperforming other learning rates. Conversely, testing ALIGN on FairFace with the probe ``person" achieved the highest macro average accuracy improvement at a learning rate of $0.05$. These variations are not surprising, given the differences in model capabilities, dataset complexities, and probe types, all of which influence the rate of adjustment parameter learning.

\textbf{Low learning rates lead to insufficient optimization.} Across all test scenarios, learning rates of 0.0001 and 0.001 consistently resulted in minimal improvements in macro average accuracy. We attribute this to the limited number of optimization epochs ($20$) set in our experiments. This choice was deliberate, as it reflects the design philosophy of the AdaLogAdjustment framework: to achieve efficient learning of adjustment factors with minimal computational overhead. Specifically, by limiting the optimization to $20$ epochs, we aim to balance performance and efficiency, making our method suitable for real-world applications where computational resources or time may be constrained. However, with such low learning rates, the optimization process is unlikely to converge to satisfactory solutions within this limited number of epochs, resulting in suboptimal performance.

\textbf{A unified learning rate strikes a balance.} Although scenario-specific tuning of learning rates could yield the highest possible performance in each case, this approach contradicts our overarching objective. The aim of this study is not to maximize bias mitigation performance in every scenario but to propose a generalized, efficient framework adaptable to diverse contexts. To this end, we fixed the learning rate at $0.01$ across all scenarios in the main paper, as validated in Fig. 4. This decision reflects a balance between generalizability and efficiency, demonstrating that AdaLogAdjustment can achieve significant improvements in macro average accuracy with minimal manual tuning.

Our findings underscore the trade-off between scenario-specific optimization and the need for a generalized framework. While further gains in bias mitigation could be achieved by tailoring the learning rate to individual scenarios, our choice of a unified learning rate of $0.01$ aligns with the goal of developing a robust and scalable method for bias mitigation in diverse contexts.

\subsection{Implementation Details of OWLv2 Adaptation}

\begin{algorithm}[htbp]
\caption{OWLv2 Adaptation for Bias Testing}
\label{alg:owlv2_adaptation}
\textbf{Input:} Image $I$, List of class labels $L = \{l_1, l_2, \dots, l_n\}$, List of probes $P = \{p_1, p_2, \dots, p_m\}$ \\
\textbf{Output:} Image-level logit and predicted probabilities for bias analysis

\begin{algorithmic}[1]
\State \textbf{Load OWLv2 components:}
\State \hspace{1em} $model \gets \text{OwlViTForObjectDetection(pretrained\_model\_path)}$
\State \hspace{1em} $processor \gets \text{OwlViTProcessor(pretrained\_model\_path)}$

\State \textbf{Construct prompt list:}
\State \hspace{1em} $prompts \gets \{\text{``a photo of a "} + l \, | \, l \in L\} \cup \{p \, | \, p \in P\}$ \textcolor[RGB]{0,138,115}{\Comment{Incorporation of prompt-based text queries}}

\State \textbf{Preprocess inputs:}
\State \hspace{1em} $inputs \gets \text{processor(text=prompts, images=I, return\_tensors=``pt")}$

\State \textbf{Forward pass through the OWLv2 model:}
\State \hspace{1em} $outputs \gets \text{model($inputs$)}$
\State \hspace{1em} $logits\_bbox \gets outputs.logits$ 

\State \textbf{Remove bounding box dependency:}
\State \hspace{1em} $logits\_image \gets \text{mean}(logits\_bbox, \text{dim}=1)$ \textcolor[RGB]{0,138,115}{\Comment{Removal of bounding box prediction}}

\State \textbf{Compute image-level predictions:}
\State \hspace{1em} $probabilities \gets \text{softmax}(logits\_image, \text{dim}=-1)$ \textcolor[RGB]{0,138,115}{\Comment{Logit for image-level predictions}}

\State \textbf{Output:} Return $probabilities$ for all prompts.
\end{algorithmic}
\end{algorithm}

OWLv2, originally designed as an open-vocabulary object detection model, focuses on detecting objects in images by predicting bounding boxes and their associated labels based on textual prompts. Although this design excels in object detection tasks, its architecture and outputs are not directly suited for the bias testing framework employed in this study. The main limitation is its default output format, which generates bounding boxes instead of a single classification result per image, as required by our testing methodology.

To adapt OWLv2 for zero-shot classification and bias testing, we introduce several modifications (Algorithm~\ref{alg:owlv2_adaptation}):

\textbf{Incorporation of prompt-based text queries (Algorithm~\ref{alg:owlv2_adaptation}, line 5).} We restructured the model to align its output logit values with a set of predefined text prompts. These prompts include both class labels and probes, ensuring that the model's predictions reflect its alignment with societal attributes and stereotypes across diverse scenarios.

\textbf{Removal of bounding box prediction (Algorithm~\ref{alg:owlv2_adaptation}, line 12).} The bounding box prediction functionality was bypassed, simplifying the output to logit corresponding to the entire image. This adjustment ensures compatibility with the classification-based framework and eliminates potential biases associated with region-specific predictions.

\textbf{Logit for image-level predictions (Algorithm~\ref{alg:owlv2_adaptation}, line 14).} To account for the model's original multi-object detection design, the logit values across all bounding boxes were aggregated into a single vector, representing the image's overall prediction. This adjustment ensures consistency in evaluating OWLv2 alongside other models like CLIP, ALIGN, and BridgeTower, which natively support classification tasks.

The adaptation of OWLv2 serves two critical purposes. First, it extends the utility of an object detection model to classification-based bias analysis, demonstrating the flexibility and adaptability of OWLv2 in novel contexts. Second, it highlights the challenges and nuances involved in repurposing models for tasks beyond their original design, providing insights into how architectural features and task-specific adjustments can influence bias testing outcomes. By modifying OWLv2, we reveal its inherent capabilities and limitations in aligning with textual prompts, highlighting the unique ways biases manifest in models not natively designed for classification. This underscores the importance of methodological rigor and transparency in adapting models for specialized research purposes.

\subsection{Pseudocode for TriProTesting and AdaLogAdjustment}

This subsection provides the pseudocode for the TriProTesting and AdaLogAdjustment frameworks used in our study (Algorithm \ref{alg:triprotesting_aladjustment}). The pseudocode formalizes the steps involved in bias testing and mitigation, offering a concise, reproducible outline of the methodologies detailed in the main paper.

\textbf{TriProTesting Framework (Algorithm \ref{alg:triprotesting_aladjustment}, lines 1-19).} The first stage of the pseudocode implements TriProTesting, which systematically evaluates biases in Foundation Models (FMs) using defined probes. By aggregating logit values for Negative, Neutral, and Positive probes across datasets, the framework enables a detailed analysis of explicit and implicit bias patterns inherent in the models. This stage focuses on revealing and quantifying biases in the FMs, providing a foundation for understanding how biases manifest before applying mitigation strategies.

\textbf{AdaLogAdjustment Framework (Algorithm \ref{alg:triprotesting_aladjustment}, lines 20-34).} The second stage introduces the Adaptive Logit Adjustment (AdaLogAdjustment) framework, which mitigates biases by dynamically learning logit adjustment factors. This stage uses a subset of the dataset as a training set to optimize adjustment factors through iterative updates. These optimized factors are subsequently applied to logit values from the testing set, yielding adjusted predictions that demonstrate reduced bias.

We provide the pseudocode to illustrate the modular design and practical implementation of the proposed TriProTesting and AdaLogAdjustment frameworks. By clearly separating the workflow into two phases—bias detection and bias mitigation—we ensure the reproducibility of our method and facilitate its application across various models and datasets. Our structured approach enables systematic analysis of biases while providing a robust mechanism for effective mitigation. In addition, we emphasize that our framework is not only technically rigorous but also socially significant. By incorporating insights from social sciences, such as systemic inequality and power redistribution, we address the societal implications of AI biases and propose a method that aligns with ethical principles. Our work reflects a commitment to building AI systems that promote fairness and equity, contributing to the development of responsible and inclusive AI technologies.

\begin{algorithm}[H]
\caption{TriProTesting and AdaLogAdjustment Framework}
\label{alg:triprotesting_aladjustment}
\textbf{Input:} Model $M$, Dataset $D$, Probes $P_{neg}, P_{neu}, P_{pos}$ (Negative, Neutral, Positive probes), Training set size $N$ \\
\textbf{Output:} Bias testing results and logit adjustment factors

\begin{algorithmic}[1]
\State \textbf{Stage 1: TriProTesting Framework}
\State \hspace{1em} \textbf{1.1 Set up probe prompts:}
\State \hspace{2em} \textbf{Define Negative Probes:} 
\State \hspace{3em} $prompts_{neg} \gets \{\text{``a photo of a "} + p \, | \, p \in P_{neg}\}$ \textcolor[RGB]{0,138,115}{\Comment{Define negative probes.}}
\State \hspace{2em} \textbf{Define Neutral Probes:} 
\State \hspace{3em} $prompts_{neu} \gets \{\text{``a photo of a "} + p \, | \, p \in P_{neu}\}$ \textcolor[RGB]{0,138,115}{\Comment{Define neutral probes.}}
\State \hspace{2em} \textbf{Define Positive Probes:} 
\State \hspace{3em} $prompts_{pos} \gets \{\text{``a photo of a "} + p \, | \, p \in P_{pos}\}$ \textcolor[RGB]{0,138,115}{\Comment{Define positive probes.}}
\State \hspace{2em} $prompts \gets prompts_{neg} \cup prompts_{neu} \cup prompts_{pos}$

\State \hspace{1em} \textbf{1.2 Perform inference:}
\For{each probe $p \in prompts$}
    \For{each sample $d \in D$}
        \State $logits_p \gets M(\text{input}=d, \text{probe}=p)$ \textcolor[RGB]{0,138,115}{\Comment{Obtain logit values for the current probe.}}
    \EndFor
\EndFor
\State \hspace{1em} \textbf{1.3 Compute bias testing results:}
\For{each probe $p \in prompts$}
    \State \hspace{2em} $bias\_results_D \gets \text{aggregate\_logits}(logits_p)$ \textcolor[RGB]{0,138,115}{\Comment{Calculate logit and analyze bias patterns.}}
\EndFor

\State \textbf{Stage 2: AdaLogAdjustment Framework}
\State \hspace{1em} \textbf{2.1 Split dataset into training and testing sets:}
\State \hspace{2em} $D_{train}, D_{test} \gets \text{split}(D, N)$ \textcolor[RGB]{0,138,115}{\Comment{Randomly select $N$ samples per class for training.}}
\State \hspace{1em} \textbf{2.2 Learn adjustment factors:}
\State \hspace{2em} Initialize adjustment factors: $\boldsymbol{\alpha} = [1, 1, \dots, 1]$
\State \hspace{2em} Define loss function: 
\State \hspace{3em} $\mathcal{L} = -\frac{1}{N \times C} \sum_{n=1}^{N \times C} \log\left(\frac{\exp(\alpha_{y_n} \cdot z_{y_n})}{\sum_{c=1}^{C+1} \exp(\alpha_c \cdot z_c)}\right)$
\For{each epoch in $1, \dots, 20$}
    \State Update $\boldsymbol{\alpha}$ using Adam optimizer on $\mathcal{L}$ \textcolor[RGB]{0,138,115}{\Comment{Iteratively optimize adjustment factors.}}
\EndFor
\State \hspace{1em} \textbf{2.3 Adjust logit for testing set:}
\For{each sample $d \in D_{test}$}
    \State $adjusted\_logit \gets logit \cdot \boldsymbol{\alpha}$ \textcolor[RGB]{0,138,115}{\Comment{Apply learned adjustment factors.}}
\EndFor

\State \textbf{Return:} $bias\_results_{D_{test}}$, $\boldsymbol{\alpha}$
\end{algorithmic}
\end{algorithm}

\begin{table}[htbp]
  \centering
  \caption{ \textbf{Original probabilities for ``predicted as probes" in Fig. 2 A-D.}
Raw probabilities of each category being predicted as probes under the TriProTesting framework, computed for four models (CLIP, ALIGN, BridgeTower, OWLv2) across four datasets (CelebA, UTKFace, FairFace, IdenProf). These values correspond to the unnormalized probabilities underlying the bubble color intensities in Fig. 2 A-D. Due to space constraints, the class names in this table are abbreviated as follows: C1:man, C2:woman; F1:White, F2:Black, F3:East Asian, F4:Indian, F5:Middle Eastern, F6:Latino\_Hispanic, F7:Southeast Asian; I1:chef, I2:doctor, I3:engineer, I4:farmer, I5:firefighter, I6:judge, I7:mechanic, I8:pilot, I9:police, I10:waiter; U1:child, U2:teenager, U3:young adult, U4:middle aged, U5:elderly.}
    \resizebox{16.5cm}{!}{\begin{tabular}{cc|cc|ccccccc|cccccccccc|ccccc}
    \toprule
          &       & \multicolumn{2}{c|}{CelebA} & \multicolumn{7}{c|}{FairFace}                         & \multicolumn{10}{c|}{IdenProf}                                                & \multicolumn{5}{c}{UTKFace} \\
\cmidrule{3-26}          &       & \multicolumn{1}{c}{C1} & \multicolumn{1}{c|}{C2} & \multicolumn{1}{c}{F1} & \multicolumn{1}{c}{F2} & \multicolumn{1}{c}{F3} & \multicolumn{1}{c}{F4} & \multicolumn{1}{c}{F5} & \multicolumn{1}{c}{F6} & \multicolumn{1}{c|}{F7} & \multicolumn{1}{c}{I1} & \multicolumn{1}{c}{I2} & \multicolumn{1}{c}{I3} & \multicolumn{1}{c}{I4} & \multicolumn{1}{c}{I5} & \multicolumn{1}{c}{I6} & \multicolumn{1}{c}{I7} & \multicolumn{1}{c}{I8} & \multicolumn{1}{c}{I9} & \multicolumn{1}{c|}{I10} & \multicolumn{1}{c}{U1} & \multicolumn{1}{c}{U2} & \multicolumn{1}{c}{U3} & \multicolumn{1}{c}{U4} & \multicolumn{1}{c}{U5} \\
    \midrule
    \multirow{15}[2]{*}{\rotatebox{90}{CLIP}} & criminal & 8.9   & 2.2   & 29.4  & 26.9  & 5.4   & 2.7   & 9.9   & 3.6   & 2.4   & 0.1   & 0.0   & 0.2   & 0.6   & 0.0   & 6.0   & 0.2   & 0.6   & 1.4   & 1.0   & 0.1   & 3.7   & 21.1  & 23.9  & 13.7  \\
          & failure & 1.5   & 0.9   & 12.9  & 7.1   & 1.8   & 0.4   & 2.6   & 0.9   & 0.7   & 1.0   & 0.7   & 1.1   & 2.4   & 5.6   & 6.8   & 4.1   & 2.2   & 0.4   & 5.0   & 0.1   & 0.4   & 5.4   & 3.0   & 1.1  \\
          & fraudster & 66.7  & 42.5  & 75.7  & 83.0  & 47.0  & 44.9  & 54.5  & 34.8  & 35.3  & 4.0   & 1.9   & 1.4   & 7.8   & 0.7   & 20.2  & 3.3   & 7.0   & 3.2   & 24.4  & 3.0   & 29.1  & 87.3  & 95.9  & 68.9  \\
          & liar  & 2.9   & 3.4   & 30.8  & 20.4  & 3.9   & 1.7   & 6.7   & 3.5   & 2.5   & 0.0   & 0.2   & 0.2   & 0.7   & 0.4   & 7.2   & 0.1   & 1.0   & 0.1   & 4.7   & 0.2   & 1.4   & 15.2  & 12.5  & 6.0  \\
          & thief & 1.5   & 0.2   & 17.1  & 29.3  & 1.2   & 1.5   & 3.6   & 1.5   & 1.4   & 0.6   & 0.0   & 0.1   & 1.7   & 0.2   & 3.4   & 1.6   & 0.9   & 0.4   & 8.9   & 0.1   & 1.6   & 7.4   & 7.2   & 3.5  \\
          & citizen & 0.6   & 0.4   & 32.0  & 62.3  & 6.1   & 7.7   & 7.1   & 3.5   & 5.4   & 0.4   & 1.2   & 1.3   & 17.6  & 2.4   & 6.0   & 2.2   & 2.3   & 5.8   & 11.4  & 0.1   & 1.4   & 7.8   & 10.2  & 5.0  \\
          & individual & 0.6   & 0.8   & 26.1  & 36.5  & 1.1   & 1.1   & 2.4   & 1.4   & 0.9   & 0.0   & 0.0   & 0.0   & 1.0   & 0.2   & 0.0   & 0.0   & 0.1   & 0.2   & 1.0   & 0.0   & 0.0   & 0.0   & 0.1   & 0.0  \\
          & person & 32.3  & 44.5  & 82.1  & 91.3  & 80.9  & 57.8  & 55.2  & 42.9  & 65.9  & 1.3   & 1.3   & 0.4   & 9.4   & 1.1   & 4.3   & 1.8   & 1.7   & 1.3   & 31.1  & 1.1   & 17.1  & 73.2  & 56.4  & 39.2  \\
          & stranger & 0.0   & 0.0   & 10.2  & 7.3   & 0.9   & 0.6   & 2.2   & 0.6   & 0.7   & 0.0   & 0.1   & 0.1   & 0.7   & 0.2   & 0.1   & 0.1   & 0.1   & 0.0   & 2.0   & 0.0   & 0.1   & 0.2   & 80.8  & 0.0  \\
          & worker & 0.0   & 0.0   & 0.6   & 4.1   & 0.3   & 0.4   & 0.3   & 0.3   & 0.4   & 0.7   & 0.3   & 32.9  & 5.8   & 2.3   & 0.0   & 18.8  & 0.0   & 0.1   & 3.3   & 0.0   & 0.0   & 0.1   & 0.1   & 0.0  \\
          & genius & 35.6  & 0.6   & 27.5  & 25.1  & 19.0  & 4.4   & 10.6  & 3.0   & 6.1   & 0.9   & 0.7   & 0.1   & 1.9   & 0.3   & 3.8   & 1.1   & 0.4   & 0.2   & 7.4   & 0.0   & 1.7   & 11.8  & 13.9  & 8.9  \\
          & hero  & 5.4   & 0.3   & 24.1  & 24.0  & 5.5   & 1.4   & 6.1   & 1.2   & 2.1   & 0.1   & 0.7   & 1.2   & 3.2   & 6.4   & 6.6   & 2.2   & 3.8   & 4.0   & 3.7   & 0.1   & 1.4   & 12.2  & 14.0  & 8.3  \\
          & leader & 9.0   & 1.7   & 36.3  & 46.2  & 19.2  & 10.3  & 15.0  & 6.2   & 8.2   & 1.8   & 0.8   & 3.4   & 4.3   & 1.2   & 18.0  & 1.6   & 9.7   & 8.8   & 8.1   & 0.0   & 1.6   & 18.4  & 25.9  & 10.9  \\
          & savior & 0.8   & 0.3   & 7.0   & 3.8   & 0.5   & 0.1   & 0.9   & 0.2   & 0.2   & 0.0   & 0.0   & 0.0   & 0.0   & 0.0   & 1.7   & 0.0   & 0.1   & 0.0   & 0.9   & 0.0   & 0.3   & 2.2   & 1.4   & 0.6  \\
          & winner & 25.6  & 15.8  & 49.8  & 47.3  & 24.1  & 4.0   & 13.6  & 6.5   & 8.6   & 2.7   & 0.7   & 0.3   & 4.9   & 0.9   & 7.8   & 2.9   & 4.6   & 0.8   & 28.8  & 0.6   & 8.5   & 35.4  & 28.7  & 13.3  \\
    \midrule
    \multirow{15}[2]{*}{\rotatebox{90}{ALIGN}} & criminal & 3.4   & 0.3   & 29.4  & 30.3  & 1.5   & 5.9   & 21.0  & 8.0   & 2.8   & 0.0   & 0.1   & 0.0   & 0.1   & 0.0   & 0.2   & 0.7   & 0.3   & 3.0   & 0.3   & 0.4   & 6.5   & 34.9  & 44.7  & 10.1  \\
          & failure & 0.5   & 0.3   & 5.5   & 3.1   & 0.4   & 0.8   & 2.1   & 1.6   & 0.5   & 0.2   & 0.3   & 0.2   & 0.3   & 0.3   & 1.2   & 0.8   & 0.0   & 0.0   & 0.2   & 0.4   & 0.3   & 6.6   & 8.6   & 0.8  \\
          & fraudster & 2.3   & 0.2   & 52.9  & 36.4  & 1.8   & 7.5   & 36.3  & 14.7  & 3.0   & 0.1   & 0.2   & 0.0   & 0.1   & 0.0   & 0.3   & 0.2   & 0.0   & 1.0   & 0.6   & 0.1   & 3.3   & 25.1  & 33.3  & 7.1  \\
          & liar  & 1.4   & 0.4   & 9.7   & 3.7   & 0.7   & 1.1   & 4.9   & 2.2   & 0.6   & 0.0   & 0.0   & 0.0   & 0.0   & 0.0   & 1.0   & 0.0   & 0.0   & 0.1   & 0.4   & 0.3   & 1.1   & 16.6  & 20.0  & 4.6  \\
          & thief & 1.3   & 0.1   & 12.1  & 10.7  & 0.5   & 1.5   & 7.1   & 2.6   & 0.8   & 0.0   & 0.0   & 0.0   & 0.0   & 0.0   & 0.0   & 0.3   & 0.0   & 0.3   & 0.1   & 0.2   & 2.5   & 13.3  & 11.2  & 2.4  \\
          & citizen & 2.7   & 0.3   & 60.1  & 60.0  & 2.6   & 9.6   & 32.1  & 18.3  & 4.5   & 0.9   & 0.8   & 0.3   & 1.1   & 0.2   & 10.8  & 0.9   & 0.3   & 0.6   & 2.0   & 0.9   & 8.6   & 61.1  & 91.9  & 52.4  \\
          & individual & 1.0   & 1.4   & 88.2  & 87.3  & 9.2   & 33.6  & 63.1  & 52.2  & 14.0  & 0.4   & 2.7   & 0.0   & 0.6   & 0.2   & 4.0   & 1.1   & 0.1   & 0.3   & 3.0   & 1.4   & 23.7  & 83.6  & 92.1  & 40.9  \\
          & person & 2.7   & 4.0   & 92.5  & 90.7  & 13.7  & 48.7  & 78.9  & 63.5  & 21.2  & 0.6   & 1.3   & 0.0   & 0.6   & 0.1   & 5.0   & 0.9   & 0.1   & 0.2   & 3.4   & 3.1   & 34.5  & 92.1  & 98.1  & 61.2  \\
          & stranger & 12.7  & 4.6   & 55.3  & 43.7  & 2.5   & 7.1   & 26.3  & 12.6  & 3.3   & 0.1   & 0.7   & 0.0   & 1.0   & 0.6   & 1.8   & 1.0   & 0.2   & 0.2   & 2.9   & 0.7   & 10.7  & 64.9  & 61.9  & 15.2  \\
          & worker & 0.0   & 0.0   & 6.2   & 6.9   & 0.3   & 0.6   & 2.7   & 1.0   & 0.4   & 0.2   & 0.1   & 31.9  & 0.8   & 0.3   & 0.2   & 6.2   & 0.0   & 0.0   & 0.8   & 0.1   & 0.6   & 18.9  & 24.5  & 1.5  \\
          & genius & 16.8  & 0.9   & 72.7  & 69.5  & 5.2   & 15.5  & 47.9  & 26.6  & 7.4   & 0.1   & 0.6   & 0.0   & 0.2   & 0.2   & 4.1   & 0.2   & 0.1   & 0.0   & 2.6   & 0.5   & 7.8   & 52.5  & 72.4  & 29.9  \\
          & hero  & 2.3   & 0.2   & 54.2  & 51.5  & 2.3   & 8.4   & 34.5  & 14.3  & 4.2   & 0.0   & 0.6   & 0.0   & 0.2   & 0.0   & 0.6   & 0.2   & 0.2   & 1.0   & 1.1   & 0.1   & 4.6   & 35.0  & 43.8  & 6.2  \\
          & leader & 4.3   & 0.4   & 50.3  & 42.0  & 1.5   & 7.5   & 26.7  & 15.2  & 2.6   & 0.7   & 0.8   & 0.0   & 0.2   & 0.1   & 9.7   & 0.1   & 0.1   & 0.1   & 1.4   & 0.2   & 5.5   & 48.2  & 78.2  & 30.8  \\
          & savior & 0.4   & 0.1   & 2.1   & 24.7  & 0.1   & 0.2   & 1.2   & 0.3   & 0.1   & 0.0   & 0.0   & 0.0   & 0.0   & 0.0   & 0.9   & 0.0   & 0.0   & 0.0   & 0.3   & 0.1   & 0.7   & 13.5  & 15.5  & 1.1  \\
          & winner & 1.4   & 0.7   & 18.3  & 20.9  & 0.4   & 1.2   & 4.5   & 3.6   & 0.7   & 0.2   & 0.4   & 0.0   & 0.2   & 0.2   & 3.1   & 0.1   & 0.1   & 0.1   & 1.6   & 0.1   & 1.2   & 23.8  & 28.2  & 2.7  \\
    \midrule
    \multirow{15}[2]{*}{\rotatebox{90}{BridgeTower}} & criminal & 1.2   & 0.3   & 11.8  & 14.9  & 1.0   & 7.0   & 17.8  & 9.6   & 2.2   & 0.1   & 0.1   & 0.0   & 0.6   & 0.0   & 0.2   & 0.1   & 0.0   & 1.0   & 0.6   & 0.1   & 2.3   & 16.1  & 16.8  & 3.9  \\
          & failure & 0.4   & 0.4   & 2.3   & 0.6   & 0.2   & 0.5   & 1.7   & 0.9   & 0.2   & 0.1   & 0.3   & 0.4   & 2.0   & 0.1   & 0.6   & 1.7   & 0.2   & 0.3   & 3.9   & 0.1   & 0.0   & 0.5   & 0.6   & 0.1  \\
          & fraudster & 1.0   & 0.3   & 16.8  & 12.3  & 1.6   & 10.6  & 24.1  & 13.1  & 2.9   & 0.1   & 0.1   & 0.0   & 1.1   & 0.0   & 0.3   & 0.7   & 0.3   & 0.1   & 1.4   & 0.1   & 1.7   & 12.7  & 18.9  & 5.2  \\
          & liar  & 1.7   & 5.6   & 34.9  & 9.5   & 1.7   & 5.4   & 20.0  & 16.5  & 1.8   & 0.2   & 0.4   & 0.2   & 3.3   & 0.1   & 2.6   & 1.0   & 0.0   & 1.0   & 11.3  & 0.2   & 2.0   & 27.0  & 13.3  & 0.8  \\
          & thief & 0.6   & 0.3   & 10.9  & 7.3   & 0.8   & 5.1   & 15.4  & 6.7   & 1.5   & 0.0   & 0.2   & 0.0   & 1.3   & 0.0   & 0.1   & 0.6   & 0.0   & 0.2   & 1.2   & 0.1   & 1.2   & 9.6   & 8.7   & 1.6  \\
          & citizen & 2.0   & 1.8   & 23.6  & 41.0  & 1.9   & 22.4  & 26.7  & 33.1  & 5.4   & 1.1   & 2.1   & 1.6   & 4.9   & 0.0   & 17.0  & 1.3   & 1.0   & 7.8   & 5.6   & 0.1   & 1.4   & 20.8  & 39.6  & 15.2  \\
          & individual & 22.8  & 66.7  & 95.1  & 88.4  & 19.0  & 75.7  & 89.5  & 84.3  & 26.8  & 1.0   & 3.3   & 0.4   & 5.0   & 0.0   & 10.9  & 1.3   & 0.4   & 3.4   & 12.7  & 0.9   & 29.8  & 93.6  & 94.5  & 55.5  \\
          & person & 22.1  & 47.0  & 97.1  & 86.6  & 18.7  & 71.6  & 90.8  & 84.8  & 26.4  & 0.9   & 3.4   & 1.0   & 5.9   & 0.0   & 11.4  & 1.8   & 0.6   & 3.7   & 14.2  & 0.8   & 23.9  & 91.2  & 89.9  & 41.2  \\
          & stranger & 10.7  & 20.4  & 91.9  & 69.2  & 19.1  & 64.1  & 85.3  & 76.1  & 27.3  & 0.2   & 1.3   & 0.6   & 5.1   & 0.1   & 3.1   & 1.1   & 0.2   & 0.7   & 8.9   & 0.6   & 18.7  & 85.6  & 74.9  & 29.2  \\
          & worker & 0.2   & 0.1   & 2.4   & 1.6   & 0.4   & 1.5   & 2.1   & 2.7   & 0.8   & 1.9   & 3.0   & 86.7  & 5.4   & 0.3   & 1.3   & 10.8  & 0.4   & 0.0   & 3.7   & 0.0   & 0.0   & 0.4   & 1.1   & 0.1  \\
          & genius & 6.6   & 2.5   & 53.5  & 24.6  & 6.7   & 28.6  & 54.1  & 33.0  & 7.2   & 0.2   & 2.2   & 0.0   & 1.3   & 0.0   & 2.1   & 0.6   & 0.1   & 0.1   & 2.8   & 0.1   & 2.1   & 37.2  & 51.9  & 13.3  \\
          & hero  & 10.3  & 3.7   & 25.0  & 12.4  & 2.4   & 7.1   & 22.4  & 16.6  & 4.0   & 0.0   & 1.1   & 0.2   & 2.6   & 0.3   & 1.8   & 0.7   & 1.0   & 5.7   & 3.6   & 0.1   & 1.8   & 25.5  & 21.0  & 5.3  \\
          & leader & 16.1  & 8.5   & 44.4  & 43.5  & 5.7   & 44.5  & 47.7  & 40.8  & 10.9  & 2.1   & 5.6   & 3.1   & 5.3   & 0.3   & 42.6  & 1.4   & 3.1   & 14.4  & 12.6  & 0.3   & 6.1   & 59.5  & 79.3  & 32.7  \\
          & savior & 2.3   & 1.8   & 28.1  & 15.9  & 2.4   & 8.1   & 25.5  & 13.8  & 3.3   & 0.3   & 1.7   & 0.1   & 3.8   & 0.3   & 2.7   & 1.0   & 0.2   & 1.0   & 9.7   & 0.0   & 1.1   & 16.8  & 12.2  & 1.9  \\
          & winner & 9.8   & 16.9  & 11.3  & 5.5   & 1.0   & 4.0   & 5.8   & 8.0   & 1.3   & 1.6   & 1.2   & 0.3   & 2.1   & 0.0   & 6.4   & 0.9   & 0.4   & 1.6   & 5.3   & 0.0   & 1.1   & 12.8  & 11.3  & 3.3  \\
    \midrule
    \multirow{15}[2]{*}{\rotatebox{90}{OWLv2}} & criminal & 97.8  & 55.0  & 0.0   & 0.0   & 0.0   & 0.0   & 0.0   & 0.0   & 0.0   & 51.1  & 81.8  & 60.7  & 43.8  & 66.7  & 92.6  & 16.3  & 46.1  & 90.7  & 70.2  & 17.7  & 56.6  & 52.9  & 29.8  & 38.9  \\
          & failure & 96.1  & 95.4  & 0.0   & 0.0   & 0.0   & 0.0   & 0.0   & 0.0   & 0.0   & 69.1  & 95.0  & 89.0  & 76.2  & 79.3  & 81.2  & 31.9  & 63.4  & 72.1  & 78.8  & 96.7  & 97.8  & 91.6  & 86.8  & 87.9  \\
          & fraudster & 92.4  & 52.3  & 0.0   & 0.0   & 0.0   & 0.0   & 0.0   & 0.0   & 0.0   & 19.0  & 50.7  & 10.4  & 5.3   & 13.7  & 71.2  & 2.7   & 21.1  & 24.8  & 33.9  & 22.6  & 71.9  & 73.0  & 54.7  & 66.6  \\
          & liar  & 100.0  & 100.0  & 23.5  & 28.3  & 18.4  & 20.5  & 23.0  & 22.2  & 23.8  & 75.9  & 97.2  & 62.8  & 72.6  & 67.9  & 97.0  & 14.7  & 42.8  & 77.0  & 90.0  & 100.0  & 100.0  & 100.0  & 100.0  & 100.0  \\
          & thief & 30.5  & 4.8   & 0.0   & 0.0   & 0.0   & 0.0   & 0.0   & 0.0   & 0.0   & 15.8  & 15.6  & 17.2  & 12.6  & 37.6  & 36.1  & 5.8   & 11.3  & 36.9  & 21.7  & 1.0   & 4.3   & 3.8   & 2.1   & 4.0  \\
          & citizen & 1.2   & 0.4   & 0.0   & 0.0   & 0.0   & 0.0   & 0.0   & 0.0   & 0.0   & 5.1   & 6.7   & 10.7  & 4.9   & 10.1  & 8.2   & 0.8   & 7.3   & 25.0  & 10.9  & 0.1   & 0.3   & 0.1   & 0.2   & 0.5  \\
          & individual & 12.9  & 3.7   & 0.0   & 0.0   & 0.0   & 0.0   & 0.0   & 0.0   & 0.0   & 40.0  & 42.8  & 39.6  & 45.4  & 47.1  & 61.6  & 6.3   & 23.1  & 57.3  & 52.4  & 0.4   & 1.4   & 1.0   & 0.8   & 1.6  \\
          & person & 5.2   & 2.5   & 0.0   & 0.0   & 0.0   & 0.0   & 0.0   & 0.0   & 0.0   & 40.4  & 35.8  & 55.6  & 57.6  & 65.9  & 56.9  & 18.2  & 25.8  & 61.9  & 49.6  & 0.7   & 1.4   & 0.8   & 0.6   & 1.4  \\
          & stranger & 99.8  & 84.7  & 0.2   & 0.2   & 0.1   & 0.2   & 0.2   & 0.2   & 0.2   & 40.4  & 72.4  & 41.9  & 38.2  & 41.6  & 74.2  & 7.7   & 33.9  & 59.9  & 62.3  & 83.7  & 99.3  & 99.9  & 99.4  & 99.5  \\
          & worker & 0.2   & 0.0   & 0.0   & 0.0   & 0.0   & 0.0   & 0.0   & 0.0   & 0.0   & 1.6   & 1.1   & 9.7   & 2.8   & 15.2  & 3.4   & 0.7   & 0.6   & 2.0   & 0.4   & 0.0   & 0.0   & 0.0   & 0.0   & 0.0  \\
          & genius & 95.8  & 71.2  & 0.0   & 0.0   & 0.0   & 0.0   & 0.0   & 0.0   & 0.0   & 51.1  & 93.6  & 28.8  & 22.0  & 24.2  & 96.7  & 4.9   & 39.0  & 42.3  & 78.4  & 16.4  & 55.1  & 47.0  & 29.2  & 39.0  \\
          & hero  & 1.5   & 0.3   & 0.0   & 0.0   & 0.0   & 0.0   & 0.0   & 0.0   & 0.0   & 6.1   & 5.3   & 22.0  & 2.9   & 30.7  & 7.8   & 1.2   & 11.7  & 44.3  & 5.7   & 0.0   & 0.1   & 0.0   & 0.0   & 0.0  \\
          & leader & 27.1  & 2.9   & 0.0   & 0.0   & 0.0   & 0.0   & 0.0   & 0.0   & 0.0   & 17.1  & 29.9  & 11.3  & 9.1   & 12.7  & 64.9  & 1.7   & 20.0  & 26.1  & 26.1  & 0.6   & 2.5   & 1.9   & 1.5   & 3.2  \\
          & savior & 0.0   & 0.1   & 0.0   & 0.0   & 0.0   & 0.0   & 0.0   & 0.0   & 0.0   & 0.4   & 0.6   & 0.1   & 0.3   & 0.9   & 0.7   & 0.0   & 0.0   & 0.1   & 1.0   & 0.0   & 0.0   & 0.0   & 0.0   & 0.0  \\
          & winner & 0.0   & 0.0   & 0.0   & 0.0   & 0.0   & 0.0   & 0.0   & 0.0   & 0.0   & 0.0   & 0.3   & 0.7   & 0.0   & 0.3   & 0.1   & 0.2   & 0.2   & 0.6   & 1.3   & 0.0   & 0.0   & 0.0   & 0.0   & 0.0  \\
    \bottomrule
    \end{tabular}}%
  \label{tab:probes_raw}%
\end{table}%

\begin{table}[htbp]
  \centering
  \caption{\textbf{Original probabilities for the ``woman" group in Fig. 3 A-D.} Unnormalized probabilities of the ``woman" group being predicted as probes across extended datasets (UTKFACE, FAIRFACE, IDENPROF) under the TriProTesting framework. These values serve as foundational data for calculating heatmap values in Fig. 3 A-D. The class names in this table are abbreviated as follows: F1: White\_woman, F2: Black\_woman, F3: East Asian\_woman, F4: Indian\_woman, F5: Middle Eastern\_woman, F6: Latino\_Hispanic\_woman, F7: Southeast Asian\_woman; I1: chef\_woman, I2: doctor\_woman, I3: engineer\_woman, I4: farmer\_woman, I5: firefighter\_woman, I6: judge\_woman, I7: mechanic\_woman, I8: pilot\_woman, I9: police\_woman, I10: waiter\_woman; U1: child\_woman, U2:teenager\_woman, U3:young adult\_woman, U4:middle aged\_woman, U5:elderly\_woman.}
   \resizebox{16.5cm}{!}{\begin{tabular}{cl|rrrrrrr|rrrrrrrrrr|rrrrr}
    \toprule
          &       & \multicolumn{7}{c|}{FAIRFACE}                         & \multicolumn{10}{c|}{IDENPROF}                                                & \multicolumn{5}{c}{UTKFACE} \\
\cmidrule{3-24}          &       & \multicolumn{1}{l}{F1} & \multicolumn{1}{l}{F2} & \multicolumn{1}{l}{F3} & \multicolumn{1}{l}{F4} & \multicolumn{1}{l}{F5} & \multicolumn{1}{l}{F6} & \multicolumn{1}{l|}{F7} & \multicolumn{1}{l}{I1} & \multicolumn{1}{l}{I2} & \multicolumn{1}{l}{I3} & \multicolumn{1}{l}{I4} & \multicolumn{1}{l}{I5} & \multicolumn{1}{l}{I6} & \multicolumn{1}{l}{I7} & \multicolumn{1}{l}{I8} & \multicolumn{1}{l}{I9} & \multicolumn{1}{l|}{I10} & \multicolumn{1}{l}{U1} & \multicolumn{1}{l}{U2} & \multicolumn{1}{l}{U3} & \multicolumn{1}{l}{U4} & \multicolumn{1}{l}{U5} \\
    \midrule
    \multirow{15}[2]{*}{\rotatebox{90}{CLIP}} & criminal & 28.3  & 19.8  & 2.5   & 1.6   & 7.8   & 3.6   & 1.5   & 0.0   & 0.0   & 0.0   & 0.3   & 0.0   & 1.7   & 0.0   & 0.4   & 1.1   & 1.0   & 0.1   & 1.5   & 13.4  & 25.2  & 9.8  \\
          & failure & 16.0  & 6.3   & 1.3   & 0.4   & 3.7   & 1.3   & 0.8   & 2.1   & 0.0   & 0.7   & 3.1   & 11.1  & 4.6   & 4.6   & 1.9   & 0.0   & 6.9   & 0.1   & 0.4   & 5.0   & 3.9   & 1.5  \\
          & fraudster & 76.3  & 83.2  & 40.3  & 37.1  & 49.9  & 37.1  & 32.0  & 5.5   & 1.5   & 2.1   & 5.1   & 1.6   & 13.5  & 2.3   & 8.4   & 2.1   & 27.5  & 2.9   & 29.5  & 85.3  & 98.0  & 61.1  \\
          & liar  & 42.1  & 26.4  & 3.8   & 2.0   & 10.5  & 5.0   & 3.2   & 0.0   & 0.2   & 0.7   & 1.4   & 1.2   & 4.9   & 0.8   & 1.2   & 0.0   & 7.7   & 0.3   & 1.9   & 21.2  & 25.3  & 8.8  \\
          & thief & 19.0  & 27.8  & 0.9   & 1.2   & 3.6   & 1.8   & 1.4   & 1.7   & 0.0   & 0.7   & 1.7   & 0.4   & 2.3   & 3.1   & 1.2   & 0.0   & 13.7  & 0.2   & 0.4   & 2.4   & 8.5   & 2.7  \\
          & citizen & 34.6  & 65.3  & 3.7   & 5.1   & 6.8   & 4.2   & 5.6   & 0.8   & 1.5   & 1.4   & 18.3  & 3.7   & 4.3   & 4.6   & 3.4   & 3.7   & 17.4  & 0.2   & 1.6   & 6.4   & 14.1  & 4.8  \\
          & individual & 31.9  & 40.8  & 0.9   & 1.3   & 4.0   & 2.1   & 1.1   & 0.0   & 0.0   & 0.0   & 1.4   & 0.8   & 0.0   & 0.0   & 0.4   & 0.5   & 1.8   & 0.0   & 0.0   & 0.0   & 0.0   & 0.0  \\
          & person & 89.8  & 93.9  & 80.7  & 57.8  & 69.9  & 56.1  & 67.8  & 3.8   & 1.7   & 0.7   & 8.3   & 2.5   & 3.8   & 3.8   & 3.1   & 2.1   & 45.7  & 1.7   & 23.9  & 80.7  & 92.9  & 51.1  \\
          & stranger & 15.0  & 7.7   & 0.9   & 0.5   & 2.9   & 0.9   & 0.9   & 0.0   & 0.0   & 0.7   & 0.9   & 0.4   & 0.3   & 0.0   & 0.4   & 0.0   & 3.0   & 0.0   & 0.2   & 0.2   & 0.1   & 0.1  \\
          & worker & 0.5   & 2.9   & 0.1   & 0.2   & 0.1   & 0.2   & 0.3   & 1.7   & 0.2   & 35.0  & 4.3   & 2.1   & 0.0   & 26.7  & 0.0   & 0.0   & 4.7   & 0.0   & 0.0   & 0.0   & 0.0   & 0.0  \\
          & genius & 9.3   & 8.5   & 2.8   & 0.4   & 2.3   & 0.6   & 0.9   & 0.0   & 0.2   & 0.0   & 0.9   & 0.8   & 0.6   & 0.0   & 0.4   & 0.0   & 3.8   & 0.0   & 0.2   & 0.6   & 0.9   & 0.1  \\
          & hero  & 13.3  & 10.2  & 0.9   & 0.3   & 2.0   & 0.5   & 0.6   & 0.4   & 0.2   & 1.4   & 0.9   & 4.1   & 2.9   & 0.0   & 3.4   & 4.2   & 3.8   & 0.1   & 0.0   & 1.0   & 3.4   & 1.3  \\
          & leader & 32.9  & 39.0  & 11.7  & 4.8   & 8.5   & 4.5   & 5.3   & 1.7   & 0.7   & 2.1   & 3.7   & 3.3   & 10.7  & 1.5   & 9.9   & 8.5   & 9.9   & 0.0   & 1.3   & 13.7  & 25.3  & 7.8  \\
          & savior & 6.7   & 2.0   & 0.3   & 0.0   & 0.9   & 0.2   & 0.1   & 0.0   & 0.0   & 0.0   & 0.0   & 0.0   & 0.9   & 0.0   & 0.0   & 0.0   & 1.2   & 0.0   & 0.2   & 1.0   & 1.4   & 0.3  \\
          & winner & 55.1  & 44.9  & 15.4  & 3.1   & 17.9  & 8.4   & 6.6   & 5.1   & 0.2   & 0.0   & 3.1   & 1.6   & 6.6   & 1.5   & 7.3   & 1.1   & 35.8  & 0.4   & 8.8   & 32.7  & 43.5  & 17.6  \\
    \midrule
    \multirow{15}[2]{*}{\rotatebox{90}{ALIGN}} & criminal & 11.3  & 15.4  & 0.8   & 2.8   & 8.4   & 3.6   & 1.4   & 0.0   & 0.0   & 0.0   & 0.0   & 0.0   & 0.3   & 0.0   & 0.0   & 0.5   & 0.2   & 0.2   & 1.2   & 6.8   & 6.3   & 1.1  \\
          & failure & 5.4   & 2.8   & 0.4   & 0.9   & 2.9   & 1.9   & 0.5   & 0.0   & 0.0   & 0.7   & 0.3   & 0.0   & 0.6   & 3.1   & 0.0   & 0.0   & 0.2   & 0.2   & 0.3   & 1.7   & 2.1   & 0.3  \\
          & fraudster & 38.9  & 26.2  & 1.1   & 4.2   & 23.6  & 10.7  & 1.9   & 0.0   & 0.0   & 0.0   & 0.0   & 0.0   & 0.0   & 0.0   & 0.0   & 0.0   & 0.6   & 0.1   & 0.7   & 4.0   & 5.2   & 1.3  \\
          & liar  & 7.2   & 2.6   & 0.6   & 1.1   & 4.3   & 2.1   & 0.5   & 0.0   & 0.0   & 0.0   & 0.0   & 0.0   & 0.6   & 0.0   & 0.0   & 0.0   & 0.2   & 0.3   & 0.4   & 7.2   & 6.5   & 1.5  \\
          & thief & 4.0   & 4.6   & 0.3   & 0.8   & 2.5   & 0.9   & 0.4   & 0.0   & 0.0   & 0.0   & 0.0   & 0.0   & 0.0   & 0.0   & 0.0   & 0.0   & 0.2   & 0.1   & 0.2   & 1.1   & 0.6   & 0.3  \\
          & citizen & 57.1  & 59.2  & 2.4   & 8.8   & 32.2  & 21.3  & 3.7   & 0.4   & 0.7   & 0.7   & 1.4   & 0.4   & 8.9   & 2.3   & 0.0   & 0.5   & 2.2   & 0.8   & 7.5   & 49.7  & 82.1  & 29.1  \\
          & individual & 93.5  & 92.4  & 10.4  & 46.7  & 83.0  & 71.9  & 16.3  & 1.7   & 3.4   & 0.0   & 1.1   & 0.4   & 4.3   & 2.3   & 0.0   & 1.1   & 4.7   & 1.5   & 28.0  & 83.0  & 91.4  & 35.0  \\
          & person & 94.4  & 92.9  & 14.2  & 58.8  & 87.4  & 76.5  & 22.6  & 1.7   & 1.2   & 0.0   & 1.1   & 0.0   & 5.5   & 1.5   & 0.0   & 0.5   & 4.9   & 3.4   & 39.0  & 91.6  & 97.0  & 48.2  \\
          & stranger & 54.6  & 41.8  & 2.4   & 7.1   & 28.2  & 14.0  & 2.9   & 0.0   & 0.5   & 0.0   & 1.1   & 0.8   & 1.7   & 1.5   & 0.0   & 0.0   & 3.6   & 0.6   & 9.5   & 54.6  & 56.4  & 11.7  \\
          & worker & 3.0   & 4.6   & 0.2   & 0.4   & 1.6   & 0.8   & 0.2   & 0.4   & 0.0   & 27.3  & 0.6   & 0.4   & 0.3   & 11.5  & 0.0   & 0.0   & 1.2   & 0.1   & 0.3   & 4.1   & 2.7   & 0.0  \\
          & genius & 66.3  & 65.2  & 3.6   & 12.2  & 39.6  & 25.5  & 5.2   & 0.0   & 0.2   & 0.0   & 0.0   & 0.0   & 2.9   & 0.0   & 0.0   & 0.0   & 1.8   & 0.2   & 4.4   & 33.2  & 47.2  & 14.5  \\
          & hero  & 33.8  & 36.5  & 1.2   & 4.1   & 15.1  & 7.6   & 2.0   & 0.0   & 0.5   & 0.0   & 0.3   & 0.0   & 0.6   & 0.0   & 0.0   & 0.0   & 0.6   & 0.1   & 0.9   & 6.4   & 5.5   & 1.1  \\
          & leader & 45.3  & 39.7  & 1.2   & 6.3   & 22.9  & 16.4  & 2.0   & 0.8   & 1.0   & 0.0   & 0.6   & 0.4   & 6.9   & 0.0   & 0.0   & 0.0   & 1.8   & 0.2   & 4.6   & 37.3  & 62.3  & 17.6  \\
          & savior & 1.5   & 2.0   & 0.1   & 0.2   & 1.0   & 0.2   & 0.1   & 0.0   & 0.0   & 0.0   & 0.0   & 0.0   & 0.0   & 0.0   & 0.0   & 0.0   & 0.2   & 0.1   & 0.3   & 4.5   & 2.1   & 0.3  \\
          & winner & 24.2  & 25.5  & 0.5   & 1.6   & 7.8   & 5.6   & 0.8   & 0.4   & 0.2   & 0.0   & 0.3   & 0.0   & 4.0   & 0.0   & 0.0   & 0.0   & 1.8   & 0.1   & 0.7   & 14.5  & 23.1  & 3.7  \\
    \midrule
    \multirow{15}[2]{*}{\rotatebox{90}{BridgeTower}} & criminal & 2.6   & 2.7   & 0.2   & 0.4   & 2.5   & 1.0   & 0.3   & 0.0   & 0.0   & 0.0   & 0.3   & 0.0   & 0.0   & 0.0   & 0.0   & 0.0   & 0.6   & 0.0   & 0.0   & 0.6   & 1.3   & 0.4  \\
          & failure & 2.5   & 0.6   & 0.1   & 0.4   & 1.9   & 0.8   & 0.2   & 0.4   & 0.5   & 0.7   & 2.9   & 0.0   & 0.9   & 4.6   & 0.0   & 0.0   & 6.5   & 0.1   & 0.0   & 0.2   & 0.1   & 0.1  \\
          & fraudster & 4.4   & 2.9   & 0.3   & 0.8   & 4.8   & 2.1   & 0.4   & 0.0   & 0.0   & 0.0   & 1.1   & 0.0   & 0.3   & 1.5   & 0.0   & 0.0   & 2.2   & 0.0   & 0.0   & 1.1   & 2.0   & 0.9  \\
          & liar  & 42.8  & 12.4  & 1.9   & 4.6   & 24.3  & 20.2  & 2.2   & 0.0   & 0.7   & 1.4   & 5.4   & 0.0   & 3.8   & 2.3   & 0.0   & 3.2   & 17.4  & 0.2   & 2.9   & 33.1  & 16.3  & 1.5  \\
          & thief & 3.9   & 1.4   & 0.3   & 0.4   & 3.3   & 1.1   & 0.3   & 0.0   & 0.0   & 0.0   & 1.4   & 0.0   & 0.3   & 1.5   & 0.0   & 0.0   & 1.8   & 0.0   & 0.0   & 0.4   & 0.6   & 0.3  \\
          & citizen & 23.1  & 42.1  & 1.7   & 19.6  & 24.6  & 34.7  & 4.8   & 2.1   & 2.4   & 3.5   & 7.4   & 0.0   & 17.3  & 6.1   & 0.4   & 12.2  & 8.9   & 0.0   & 0.9   & 14.1  & 35.5  & 11.8  \\
          & individual & 96.5  & 89.1  & 18.4  & 73.9  & 92.7  & 88.0  & 26.8  & 1.7   & 5.1   & 0.7   & 6.6   & 0.0   & 15.9  & 3.8   & 0.4   & 9.0   & 18.6  & 1.1   & 37.8  & 95.9  & 98.3  & 65.0  \\
          & person & 97.0  & 84.7  & 16.3  & 64.4  & 91.1  & 85.7  & 23.9  & 0.8   & 4.9   & 2.8   & 7.1   & 0.0   & 16.7  & 4.6   & 0.4   & 7.4   & 20.0  & 0.9   & 29.8  & 93.1  & 94.8  & 45.4  \\
          & stranger & 91.6  & 65.5  & 16.4  & 54.3  & 83.4  & 76.0  & 23.4  & 0.0   & 1.7   & 1.4   & 5.7   & 0.4   & 5.2   & 2.3   & 0.0   & 1.6   & 12.7  & 0.6   & 21.9  & 86.5  & 81.7  & 35.0  \\
          & worker & 1.1   & 1.2   & 0.1   & 0.3   & 0.6   & 1.3   & 0.3   & 2.1   & 3.2   & 93.0  & 7.4   & 0.4   & 1.7   & 23.7  & 0.0   & 0.0   & 6.1   & 0.0   & 0.0   & 0.1   & 0.2   & 0.0  \\
          & genius & 28.2  & 11.4  & 1.8   & 7.0   & 22.0  & 14.6  & 2.2   & 0.0   & 2.2   & 0.0   & 2.3   & 0.0   & 1.4   & 1.5   & 0.0   & 0.0   & 4.0   & 0.1   & 0.4   & 15.5  & 16.5  & 3.3  \\
          & hero  & 12.9  & 4.8   & 0.8   & 2.0   & 7.4   & 8.1   & 1.4   & 0.0   & 1.0   & 0.0   & 2.6   & 0.8   & 1.2   & 0.8   & 0.8   & 6.4   & 5.1   & 0.0   & 0.9   & 8.2   & 4.9   & 1.0  \\
          & leader & 31.8  & 37.4  & 2.2   & 29.5  & 27.6  & 31.3  & 5.2   & 2.1   & 6.6   & 6.3   & 6.9   & 0.8   & 42.7  & 3.8   & 2.3   & 20.6  & 16.4  & 0.2   & 5.9   & 51.2  & 66.5  & 21.6  \\
          & savior & 11.6  & 7.3   & 1.2   & 3.4   & 9.2   & 5.8   & 1.6   & 0.0   & 1.7   & 0.0   & 4.9   & 0.8   & 2.9   & 2.3   & 0.0   & 1.1   & 14.6  & 0.0   & 0.4   & 4.8   & 1.6   & 0.5  \\
          & winner & 10.5  & 5.5   & 0.6   & 2.1   & 3.6   & 7.2   & 0.9   & 2.1   & 2.2   & 0.7   & 3.7   & 0.0   & 8.7   & 0.8   & 0.8   & 4.8   & 7.9   & 0.1   & 1.3   & 12.4  & 10.4  & 3.0  \\
    \midrule
    \multirow{15}[2]{*}{\rotatebox{90}{OWLv2}} & criminal & 0.0   & 0.0   & 0.0   & 0.0   & 0.0   & 0.0   & 0.0   & 48.5  & 81.6  & 67.1  & 44.3  & 61.5  & 92.2  & 23.7  & 50.4  & 94.2  & 67.9  & 18.3  & 66.6  & 79.0  & 71.1  & 70.9  \\
          & failure & 0.0   & 0.0   & 0.0   & 0.0   & 0.0   & 0.0   & 0.0   & 65.4  & 96.8  & 90.9  & 70.6  & 74.6  & 81.3  & 42.0  & 67.2  & 68.3  & 80.2  & 96.4  & 99.0  & 98.5  & 97.0  & 95.3  \\
          & fraudster & 0.0   & 0.0   & 0.0   & 0.0   & 0.0   & 0.0   & 0.0   & 20.3  & 48.5  & 14.0  & 2.9   & 13.1  & 66.9  & 2.3   & 27.9  & 31.2  & 33.3  & 24.2  & 79.8  & 92.6  & 92.0  & 88.9  \\
          & liar  & 19.5  & 27.8  & 14.7  & 20.7  & 20.0  & 18.7  & 22.1  & 77.6  & 99.5  & 73.4  & 76.6  & 65.2  & 98.3  & 26.7  & 52.7  & 87.8  & 94.1  & 100.0  & 100.0  & 100.0  & 100.0  & 100.0  \\
          & thief & 0.0   & 0.0   & 0.0   & 0.0   & 0.0   & 0.0   & 0.0   & 15.6  & 17.0  & 26.6  & 12.9  & 34.0  & 41.2  & 8.4   & 15.7  & 43.4  & 21.8  & 1.2   & 5.9   & 6.2   & 5.7   & 8.7  \\
          & citizen & 0.0   & 0.0   & 0.0   & 0.0   & 0.0   & 0.0   & 0.0   & 8.4   & 9.2   & 19.6  & 5.1   & 10.3  & 9.2   & 0.8   & 12.2  & 33.9  & 13.9  & 0.1   & 0.6   & 0.3   & 0.4   & 1.2  \\
          & individual & 0.0   & 0.0   & 0.0   & 0.0   & 0.0   & 0.0   & 0.0   & 41.4  & 47.8  & 50.4  & 56.3  & 43.9  & 67.7  & 12.2  & 29.4  & 65.6  & 55.0  & 0.7   & 2.1   & 1.7   & 2.3   & 3.7  \\
          & person & 0.0   & 0.0   & 0.0   & 0.0   & 0.0   & 0.0   & 0.0   & 44.3  & 42.0  & 61.5  & 71.4  & 60.3  & 64.8  & 32.1  & 32.1  & 70.9  & 52.9  & 0.9   & 1.8   & 1.3   & 1.7   & 3.3  \\
          & stranger & 0.1   & 0.1   & 0.1   & 0.1   & 0.0   & 0.1   & 0.1   & 38.8  & 73.3  & 51.8  & 38.0  & 41.8  & 72.9  & 13.0  & 38.9  & 65.1  & 62.4  & 81.2  & 99.3  & 100.0  & 100.0  & 100.0  \\
          & worker & 0.0   & 0.0   & 0.0   & 0.0   & 0.0   & 0.0   & 0.0   & 1.7   & 1.0   & 15.4  & 2.6   & 13.1  & 1.4   & 0.8   & 0.4   & 1.1   & 0.4   & 0.0   & 0.0   & 0.0   & 0.0   & 0.0  \\
          & genius & 0.0   & 0.0   & 0.0   & 0.0   & 0.0   & 0.0   & 0.0   & 48.5  & 96.4  & 35.0  & 19.1  & 24.2  & 96.5  & 6.9   & 44.7  & 43.9  & 78.4  & 17.3  & 61.1  & 70.6  & 65.8  & 66.4  \\
          & hero  & 0.0   & 0.0   & 0.0   & 0.0   & 0.0   & 0.0   & 0.0   & 6.3   & 5.1   & 31.5  & 1.7   & 28.7  & 7.5   & 1.5   & 14.1  & 52.4  & 6.1   & 0.0   & 0.2   & 0.1   & 0.0   & 0.0  \\
          & leader & 0.0   & 0.0   & 0.0   & 0.0   & 0.0   & 0.0   & 0.0   & 16.9  & 31.3  & 17.5  & 8.3   & 13.5  & 64.0  & 1.5   & 24.4  & 32.8  & 27.7  & 0.9   & 3.8   & 3.3   & 4.5   & 7.4  \\
          & savior & 0.0   & 0.0   & 0.0   & 0.0   & 0.0   & 0.0   & 0.0   & 0.8   & 0.7   & 0.7   & 0.6   & 0.4   & 1.4   & 0.0   & 0.0   & 0.5   & 1.6   & 0.0   & 0.0   & 0.0   & 0.0   & 0.0  \\
          & winner & 0.0   & 0.0   & 0.0   & 0.0   & 0.0   & 0.0   & 0.0   & 0.0   & 0.5   & 1.4   & 0.0   & 0.8   & 0.0   & 0.0   & 0.4   & 1.1   & 1.6   & 0.0   & 0.0   & 0.0   & 0.0   & 0.0  \\
    \bottomrule
    \end{tabular}}%
  \label{tab:woman_raw}%
\end{table}%

\begin{table}[htbp]
  \centering
  \caption{\textbf{Original probabilities for the ``man" group in Fig. 3 A-D.}
Unnormalized probabilities of the ``man" group being predicted as probes across extended datasets (UTKFACE, FAIRFACE, IDENPROF) under the TriProTesting framework. Together with Table~\ref{tab:woman_raw}, these values provide the raw inputs for computing heatmap values in Fig. 3 A-D. The class names in this table are abbreviated as follows: F1: White\_man, F2: Black\_man, F3: East Asian\_man, F4: Indian\_man, F5: Middle Eastern\_man, F6: Latino\_Hispanic\_man, F7: Southeast Asian\_man; I1: chef\_man, I2: doctor\_man, I3: engineer\_man, I4: farmer\_man, I5: firefighter\_man, I6: judge\_man, I7: mechanic\_man, I8: pilot\_man, I9: police\_man, I10: waiter\_man; U1: child\_man, U2:teenager\_man, U3:young adult\_man, U4:middle aged\_man, U5:elderly\_man.}
    \resizebox{16.5cm}{!}{\begin{tabular}{cl|rrrrrrr|rrrrrrrrrr|rrrrr}
    \toprule
          &       & \multicolumn{7}{c|}{FAIRFACE}                         & \multicolumn{10}{c|}{IDENPROF}                                                & \multicolumn{5}{c}{UTKFACE} \\
\cmidrule{3-24}          &       & \multicolumn{1}{l}{F1} & \multicolumn{1}{l}{F2} & \multicolumn{1}{l}{F3} & \multicolumn{1}{l}{F4} & \multicolumn{1}{l}{F5} & \multicolumn{1}{l}{F6} & \multicolumn{1}{l|}{F7} & \multicolumn{1}{l}{I1} & \multicolumn{1}{l}{I2} & \multicolumn{1}{l}{I3} & \multicolumn{1}{l}{I4} & \multicolumn{1}{l}{I5} & \multicolumn{1}{l}{I6} & \multicolumn{1}{l}{I7} & \multicolumn{1}{l}{I8} & \multicolumn{1}{l}{I9} & \multicolumn{1}{l|}{I10} & \multicolumn{1}{l}{U1} & \multicolumn{1}{l}{U2} & \multicolumn{1}{l}{U3} & \multicolumn{1}{l}{U4} & \multicolumn{1}{l}{U5} \\
    \midrule
    \multirow{15}[2]{*}{\rotatebox{90}{CLIP}} & criminal & 30.3  & 34.9  & 8.4   & 3.7   & 10.9  & 3.7   & 3.3   & 0.2   & 0.0   & 0.3   & 0.7   & 0.0   & 8.7   & 0.3   & 0.6   & 1.6   & 1.0   & 0.2   & 6.8   & 30.7  & 23.2  & 16.7  \\
          & failure & 10.1  & 7.9   & 2.2   & 0.4   & 2.1   & 0.5   & 0.7   & 0.6   & 1.2   & 1.2   & 2.0   & 3.5   & 8.1   & 4.0   & 2.4   & 0.6   & 2.7   & 0.2   & 0.4   & 6.0   & 2.6   & 0.8  \\
          & fraudster & 75.3  & 82.8  & 54.1  & 52.9  & 56.6  & 32.5  & 38.7  & 3.5   & 2.3   & 1.3   & 9.5   & 0.3   & 24.4  & 3.5   & 6.4   & 3.5   & 20.7  & 3.1   & 28.5  & 89.6  & 94.8  & 74.7  \\
          & liar  & 20.5  & 13.8  & 4.0   & 1.4   & 4.9   & 1.9   & 1.8   & 0.0   & 0.2   & 0.1   & 0.2   & 0.2   & 8.7   & 0.0   & 0.9   & 0.1   & 1.0   & 0.1   & 0.6   & 7.7   & 6.1   & 3.9  \\
          & thief & 15.3  & 30.9  & 1.6   & 1.9   & 3.5   & 1.1   & 1.4   & 0.2   & 0.0   & 0.0   & 1.6   & 0.2   & 4.2   & 1.3   & 0.8   & 0.6   & 3.0   & 0.0   & 3.2   & 13.6  & 6.6   & 4.0  \\
          & citizen & 29.7  & 58.8  & 8.7   & 10.4  & 7.3   & 2.9   & 5.2   & 0.3   & 1.0   & 1.3   & 17.1  & 2.0   & 7.1   & 1.8   & 1.9   & 6.3   & 4.2   & 0.0   & 1.0   & 9.5   & 8.3   & 5.2  \\
          & individual & 20.8  & 31.7  & 1.3   & 1.0   & 1.7   & 0.7   & 0.7   & 0.0   & 0.0   & 0.0   & 0.7   & 0.0   & 0.0   & 0.0   & 0.0   & 0.1   & 0.0   & 0.0   & 0.0   & 0.0   & 0.1   & 0.0  \\
          & person & 75.2  & 88.4  & 81.2  & 57.7  & 48.6  & 29.2  & 63.9  & 0.5   & 1.0   & 0.4   & 10.2  & 0.6   & 4.7   & 1.4   & 1.1   & 1.1   & 13.3  & 0.2   & 7.8   & 64.0  & 37.9  & 30.3  \\
          & stranger & 5.8   & 6.9   & 0.8   & 0.6   & 1.8   & 0.4   & 0.4   & 0.0   & 0.2   & 0.0   & 0.6   & 0.2   & 0.0   & 0.1   & 0.0   & 0.0   & 0.7   & 0.0   & 0.0   & 0.3   & 0.1   & 0.0  \\
          & worker & 0.7   & 5.5   & 0.5   & 0.6   & 0.3   & 0.4   & 0.6   & 0.3   & 0.4   & 32.5  & 6.7   & 2.4   & 0.0   & 17.4  & 0.0   & 0.1   & 1.7   & 0.0   & 0.0   & 0.2   & 0.1   & 0.0  \\
          & genius & 44.1  & 43.7  & 36.4  & 8.5   & 14.4  & 5.6   & 11.6  & 1.2   & 1.0   & 0.1   & 2.6   & 0.2   & 5.8   & 1.3   & 0.5   & 0.3   & 11.9  & 0.0   & 3.8   & 25.7  & 20.4  & 15.5  \\
          & hero  & 33.9  & 39.4  & 10.4  & 2.6   & 7.9   & 1.9   & 3.6   & 0.0   & 1.0   & 1.2   & 4.7   & 7.3   & 8.9   & 2.6   & 3.9   & 3.9   & 3.5   & 0.2   & 3.2   & 26.0  & 19.4  & 13.6  \\
          & leader & 39.4  & 54.3  & 27.3  & 16.1  & 17.9  & 8.0   & 11.1  & 1.8   & 0.8   & 3.7   & 4.7   & 0.5   & 22.6  & 1.6   & 9.6   & 8.9   & 5.9   & 0.0   & 2.0   & 24.1  & 26.2  & 13.3  \\
          & savior & 7.4   & 5.8   & 0.7   & 0.2   & 0.9   & 0.2   & 0.2   & 0.0   & 0.0   & 0.0   & 0.0   & 0.0   & 2.2   & 0.0   & 0.2   & 0.0   & 0.5   & 0.0   & 0.6   & 3.6   & 1.4   & 0.9  \\
          & winner & 45.0  & 50.0  & 33.3  & 4.9   & 11.7  & 4.4   & 10.7  & 1.8   & 1.0   & 0.4   & 6.0   & 0.6   & 8.5   & 3.1   & 3.5   & 0.7   & 20.3  & 0.8   & 8.0   & 38.7  & 21.1  & 10.1  \\
    \midrule
    \multirow{15}[2]{*}{\rotatebox{90}{ALIGN}} & criminal & 45.9  & 47.1  & 2.1   & 9.1   & 26.8  & 12.5  & 4.4   & 0.0   & 0.2   & 0.0   & 0.2   & 0.0   & 0.2   & 0.8   & 0.5   & 3.7   & 0.5   & 0.7   & 13.9  & 69.8  & 64.0  & 16.8  \\
          & failure & 5.6   & 3.4   & 0.5   & 0.7   & 1.8   & 1.2   & 0.5   & 0.3   & 0.6   & 0.1   & 0.4   & 0.5   & 1.6   & 0.4   & 0.0   & 0.0   & 0.3   & 0.7   & 0.4   & 12.7  & 11.9  & 1.1  \\
          & fraudster & 65.6  & 48.0  & 2.6   & 10.9  & 42.1  & 18.9  & 4.1   & 0.2   & 0.4   & 0.0   & 0.2   & 0.0   & 0.5   & 0.3   & 0.0   & 1.3   & 0.5   & 0.2   & 6.8   & 51.2  & 47.5  & 11.4  \\
          & liar  & 11.9  & 5.0   & 0.8   & 1.1   & 5.1   & 2.2   & 0.7   & 0.0   & 0.0   & 0.0   & 0.0   & 0.0   & 1.3   & 0.0   & 0.0   & 0.1   & 0.7   & 0.4   & 2.0   & 28.3  & 26.8  & 6.9  \\
          & thief & 19.5  & 17.5  & 0.6   & 2.2   & 9.3   & 4.3   & 1.2   & 0.0   & 0.0   & 0.0   & 0.0   & 0.0   & 0.0   & 0.4   & 0.0   & 0.4   & 0.0   & 0.3   & 5.8   & 28.4  & 16.5  & 3.9  \\
          & citizen & 62.8  & 61.0  & 2.9   & 10.5  & 32.0  & 15.1  & 5.3   & 1.1   & 0.8   & 0.3   & 0.9   & 0.2   & 11.9  & 0.7   & 0.5   & 0.6   & 1.7   & 0.9   & 10.0  & 75.2  & 96.8  & 69.7  \\
          & individual & 83.3  & 81.7  & 8.0   & 20.1  & 54.1  & 31.5  & 11.7  & 0.0   & 2.1   & 0.0   & 0.2   & 0.2   & 3.8   & 0.9   & 0.2   & 0.1   & 1.0   & 1.3   & 17.9  & 84.3  & 92.4  & 45.3  \\
          & person & 90.9  & 88.2  & 13.2  & 38.3  & 75.1  & 49.8  & 19.8  & 0.2   & 1.4   & 0.0   & 0.2   & 0.2   & 4.7   & 0.8   & 0.2   & 0.1   & 1.7   & 2.4   & 28.3  & 92.7  & 98.7  & 70.9  \\
          & stranger & 55.9  & 45.7  & 2.5   & 7.1   & 25.4  & 11.0  & 3.8   & 0.2   & 0.8   & 0.0   & 0.9   & 0.5   & 1.8   & 0.9   & 0.3   & 0.3   & 2.0   & 0.8   & 12.3  & 77.6  & 64.6  & 17.7  \\
          & worker & 9.2   & 9.4   & 0.5   & 0.8   & 3.3   & 1.3   & 0.5   & 0.2   & 0.2   & 32.8  & 0.9   & 0.3   & 0.2   & 5.3   & 0.0   & 0.0   & 0.3   & 0.1   & 1.0   & 37.2  & 35.6  & 2.6  \\
          & genius & 78.5  & 74.4  & 6.8   & 18.8  & 51.7  & 27.7  & 9.7   & 0.2   & 0.8   & 0.0   & 0.4   & 0.3   & 4.9   & 0.3   & 0.2   & 0.0   & 3.5   & 0.9   & 12.5  & 76.5  & 85.1  & 41.3  \\
          & hero  & 72.8  & 68.4  & 3.4   & 12.9  & 43.4  & 21.3  & 6.5   & 0.0   & 0.6   & 0.0   & 0.2   & 0.0   & 0.5   & 0.3   & 0.3   & 1.3   & 1.7   & 0.2   & 9.6   & 70.4  & 63.1  & 10.0  \\
          & leader & 54.8  & 44.5  & 1.7   & 8.8   & 28.5  & 13.9  & 3.2   & 0.6   & 0.6   & 0.0   & 0.0   & 0.0   & 11.4  & 0.1   & 0.2   & 0.1   & 1.0   & 0.3   & 6.8   & 61.8  & 86.2  & 40.7  \\
          & savior & 2.6   & 3.0   & 0.2   & 0.2   & 1.3   & 0.4   & 0.1   & 0.0   & 0.0   & 0.0   & 0.0   & 0.0   & 1.5   & 0.0   & 0.0   & 0.0   & 0.5   & 0.1   & 1.2   & 24.5  & 22.2  & 1.7  \\
          & winner & 12.9  & 15.9  & 0.4   & 0.7   & 3.0   & 1.4   & 0.6   & 0.2   & 0.6   & 0.0   & 0.2   & 0.3   & 2.5   & 0.1   & 0.2   & 0.1   & 1.2   & 0.0   & 1.8   & 35.3  & 30.8  & 2.0  \\
    \midrule
    \multirow{15}[2]{*}{\rotatebox{90}{BridgeTower}} & criminal & 20.0  & 28.6  & 1.8   & 13.8  & 24.9  & 18.6  & 4.3   & 0.2   & 0.2   & 0.0   & 0.7   & 0.0   & 0.4   & 0.1   & 0.0   & 1.3   & 0.5   & 0.2   & 5.4   & 35.3  & 24.6  & 6.5  \\
          & failure & 2.2   & 0.7   & 0.2   & 0.5   & 1.7   & 1.0   & 0.2   & 0.0   & 0.2   & 0.4   & 1.5   & 0.2   & 0.4   & 1.2   & 0.3   & 0.4   & 0.7   & 0.1   & 0.0   & 0.9   & 0.8   & 0.2  \\
          & fraudster & 28.1  & 22.8  & 2.9   & 20.6  & 32.8  & 24.7  & 5.4   & 0.2   & 0.2   & 0.0   & 1.1   & 0.0   & 0.4   & 0.5   & 0.5   & 0.1   & 0.5   & 0.2   & 4.0   & 26.9  & 27.4  & 8.4  \\
          & liar  & 27.8  & 6.2   & 1.5   & 6.4   & 18.1  & 12.7  & 1.3   & 0.3   & 0.2   & 0.0   & 2.0   & 0.2   & 1.8   & 0.8   & 0.0   & 0.4   & 4.0   & 0.1   & 0.8   & 19.6  & 11.7  & 0.4  \\
          & thief & 17.3  & 13.9  & 1.4   & 9.9   & 21.0  & 12.7  & 2.7   & 0.0   & 0.4   & 0.0   & 1.3   & 0.0   & 0.0   & 0.4   & 0.0   & 0.3   & 0.5   & 0.2   & 2.8   & 20.9  & 12.8  & 2.6  \\
          & citizen & 24.0  & 39.8  & 2.1   & 25.3  & 27.7  & 31.3  & 6.0   & 0.8   & 1.8   & 1.2   & 3.3   & 0.0   & 16.8  & 0.5   & 1.3   & 6.6   & 1.5   & 0.2   & 2.2   & 29.1  & 41.6  & 17.8  \\
          & individual & 93.8  & 87.7  & 19.7  & 77.4  & 88.0  & 80.5  & 26.7  & 0.8   & 1.8   & 0.4   & 4.0   & 0.0   & 7.8   & 0.9   & 0.5   & 2.0   & 5.4   & 0.7   & 18.9  & 90.7  & 92.6  & 48.4  \\
          & person & 97.2  & 88.7  & 21.1  & 79.2  & 90.7  & 83.8  & 29.1  & 0.9   & 2.3   & 0.7   & 5.1   & 0.0   & 8.1   & 1.3   & 0.6   & 2.7   & 7.2   & 0.5   & 15.9  & 88.9  & 87.4  & 38.1  \\
          & stranger & 92.1  & 73.4  & 22.0  & 74.2  & 86.1  & 76.1  & 31.5  & 0.3   & 1.0   & 0.4   & 4.7   & 0.0   & 1.8   & 0.9   & 0.3   & 0.4   & 4.2   & 0.5   & 14.5  & 84.5  & 71.4  & 24.9  \\
          & worker & 3.5   & 2.1   & 0.7   & 2.7   & 2.8   & 4.2   & 1.4   & 1.8   & 2.9   & 85.5  & 4.2   & 0.3   & 1.1   & 8.6   & 0.6   & 0.0   & 0.7   & 0.1   & 0.0   & 0.7   & 1.6   & 0.2  \\
          & genius & 76.5  & 39.4  & 11.9  & 50.9  & 68.7  & 52.4  & 12.6  & 0.3   & 2.3   & 0.0   & 0.7   & 0.0   & 2.5   & 0.4   & 0.2   & 0.1   & 1.2   & 0.2   & 4.4   & 64.1  & 69.8  & 20.7  \\
          & hero  & 36.0  & 21.1  & 4.1   & 12.3  & 29.3  & 25.7  & 6.7   & 0.0   & 1.2   & 0.3   & 2.6   & 0.2   & 2.2   & 0.7   & 1.1   & 5.5   & 1.7   & 0.2   & 3.0   & 47.0  & 29.1  & 8.5  \\
          & leader & 55.8  & 50.4  & 9.5   & 59.9  & 56.9  & 50.8  & 17.0  & 2.1   & 4.7   & 2.5   & 4.4   & 0.2   & 42.5  & 1.0   & 3.5   & 12.8  & 7.9   & 0.3   & 6.4   & 69.7  & 85.7  & 41.0  \\
          & savior & 43.0  & 25.6  & 3.6   & 12.9  & 33.0  & 22.3  & 5.0   & 0.5   & 1.6   & 0.1   & 3.1   & 0.2   & 2.5   & 0.8   & 0.3   & 1.0   & 3.7   & 0.1   & 2.0   & 31.6  & 17.5  & 2.9  \\
          & winner & 12.1  & 5.4   & 1.4   & 6.0   & 6.8   & 8.8   & 1.6   & 1.4   & 0.4   & 0.3   & 1.1   & 0.0   & 5.1   & 0.9   & 0.3   & 0.7   & 2.2   & 0.0   & 0.8   & 13.3  & 11.7  & 3.6  \\
    \midrule
    \multirow{15}[2]{*}{\rotatebox{90}{OWLv2}} & criminal & 0.0   & 0.0   & 0.0   & 0.0   & 0.0   & 0.0   & 0.0   & 52.0  & 82.0  & 59.5  & 43.5  & 68.6  & 92.8  & 15.1  & 44.4  & 89.7  & 73.1  & 16.7  & 43.0  & 20.6  & 8.9   & 15.0  \\
          & failure & 0.0   & 0.0   & 0.0   & 0.0   & 0.0   & 0.0   & 0.0   & 70.4  & 93.4  & 88.6  & 79.8  & 81.1  & 81.2  & 30.2  & 61.9  & 73.1  & 77.0  & 97.2  & 96.2  & 82.9  & 81.6  & 82.4  \\
          & fraudster & 0.0   & 0.0   & 0.0   & 0.0   & 0.0   & 0.0   & 0.0   & 18.6  & 52.5  & 9.8   & 6.9   & 13.9  & 74.0  & 2.7   & 18.3  & 23.1  & 34.6  & 20.0  & 61.0  & 48.6  & 35.9  & 50.0  \\
          & liar  & 27.2  & 29.0  & 22.4  & 20.3  & 24.4  & 25.8  & 25.6  & 75.3  & 95.3  & 60.8  & 70.0  & 68.9  & 96.2  & 12.6  & 38.7  & 74.1  & 84.9  & 100.0  & 100.0  & 100.0  & 100.0  & 100.0  \\
          & thief & 0.0   & 0.0   & 0.0   & 0.0   & 0.0   & 0.0   & 0.0   & 15.8  & 14.3  & 15.5  & 12.4  & 38.9  & 32.9  & 5.3   & 9.6   & 35.2  & 21.5  & 0.6   & 2.2   & 0.8   & 0.3   & 0.4  \\
          & citizen & 0.0   & 0.0   & 0.0   & 0.0   & 0.0   & 0.0   & 0.0   & 3.9   & 4.5   & 9.0   & 4.7   & 10.1  & 7.6   & 0.8   & 5.3   & 22.6  & 7.2   & 0.0   & 0.0   & 0.0   & 0.0   & 0.0  \\
          & individual & 0.0   & 0.0   & 0.0   & 0.0   & 0.0   & 0.0   & 0.0   & 39.5  & 38.5  & 37.5  & 38.6  & 48.3  & 57.7  & 5.3   & 20.5  & 55.1  & 49.4  & 0.1   & 0.6   & 0.1   & 0.0   & 0.0  \\
          & person & 0.0   & 0.0   & 0.0   & 0.0   & 0.0   & 0.0   & 0.0   & 39.1  & 30.5  & 54.4  & 48.7  & 68.0  & 51.9  & 15.9  & 23.2  & 59.5  & 45.4  & 0.3   & 0.8   & 0.1   & 0.0   & 0.0  \\
          & stranger & 0.4   & 0.3   & 0.2   & 0.3   & 0.3   & 0.2   & 0.2   & 41.0  & 71.7  & 40.0  & 38.4  & 41.5  & 75.1  & 6.8   & 31.8  & 58.5  & 62.2  & 87.8  & 99.4  & 99.9  & 99.1  & 99.2  \\
          & worker & 0.0   & 0.0   & 0.0   & 0.0   & 0.0   & 0.0   & 0.0   & 1.5   & 1.2   & 8.6   & 2.9   & 16.0  & 4.7   & 0.7   & 0.6   & 2.3   & 0.5   & 0.0   & 0.0   & 0.0   & 0.0   & 0.0  \\
          & genius & 0.0   & 0.0   & 0.0   & 0.0   & 0.0   & 0.0   & 0.0   & 52.0  & 91.2  & 27.6  & 23.8  & 24.2  & 96.8  & 4.6   & 36.7  & 41.9  & 78.5  & 14.9  & 46.8  & 17.8  & 10.7  & 18.7  \\
          & hero  & 0.0   & 0.0   & 0.0   & 0.0   & 0.0   & 0.0   & 0.0   & 6.0   & 5.5   & 20.2  & 3.6   & 31.4  & 8.0   & 1.2   & 10.7  & 42.2  & 5.2   & 0.0   & 0.0   & 0.0   & 0.0   & 0.0  \\
          & leader & 0.0   & 0.0   & 0.0   & 0.0   & 0.0   & 0.0   & 0.0   & 17.2  & 28.7  & 10.2  & 9.6   & 12.4  & 65.5  & 1.7   & 18.2  & 24.3  & 24.2  & 0.1   & 0.6   & 0.1   & 0.0   & 0.1  \\
          & savior & 0.0   & 0.0   & 0.0   & 0.0   & 0.0   & 0.0   & 0.0   & 0.3   & 0.4   & 0.0   & 0.2   & 1.1   & 0.2   & 0.0   & 0.0   & 0.0   & 0.3   & 0.0   & 0.0   & 0.0   & 0.0   & 0.0  \\
          & winner & 0.0   & 0.0   & 0.0   & 0.0   & 0.0   & 0.0   & 0.0   & 0.0   & 0.2   & 0.5   & 0.0   & 0.2   & 0.2   & 0.3   & 0.2   & 0.4   & 1.0   & 0.0   & 0.0   & 0.0   & 0.0   & 0.0  \\
    \bottomrule
    \end{tabular}}%
  \label{tab:man_raw}%
\end{table}%

\begin{table}[htbp]
  \centering
  \caption{\textbf{Normalized probabilities for the ``woman" group in Fig. 3 A-D.}
Normalized probabilities of the ``woman" group being predicted as probes across extended datasets (UTKFACE, FAIRFACE, IDENPROF). These values, together with Table~\ref{tab:man_norm}, form the basis for calculating the heatmap values presented in Fig. 3 A-D. The class names in this table are abbreviated as follows: F1: White\_woman, F2: Black\_woman, F3: East Asian\_woman, F4: Indian\_woman, F5: Middle Eastern\_woman, F6: Latino\_Hispanic\_woman, F7: Southeast Asian\_woman; I1: chef\_woman, I2: doctor\_woman, I3: engineer\_woman, I4: farmer\_woman, I5: firefighter\_woman, I6: judge\_woman, I7: mechanic\_woman, I8: pilot\_woman, I9: police\_woman, I10: waiter\_woman; U1: child\_woman, U2:teenager\_woman, U3:young adult\_woman, U4:middle aged\_woman, U5:elderly\_woman.}
    \resizebox{16.5cm}{!}{\begin{tabular}{cl|rrrrrrr|rrrrrrrrrr|rrrrr}
    \toprule
          &       & \multicolumn{7}{c|}{FAIRFACE}                         & \multicolumn{10}{c|}{IDENPROF}                                                & \multicolumn{5}{c}{UTKFACE} \\
\cmidrule{3-24}          &       & \multicolumn{1}{l}{F1} & \multicolumn{1}{l}{F2} & \multicolumn{1}{l}{F3} & \multicolumn{1}{l}{F4} & \multicolumn{1}{l}{F5} & \multicolumn{1}{l}{F6} & \multicolumn{1}{l|}{F7} & \multicolumn{1}{l}{I1} & \multicolumn{1}{l}{I2} & \multicolumn{1}{l}{I3} & \multicolumn{1}{l}{I4} & \multicolumn{1}{l}{I5} & \multicolumn{1}{l}{I6} & \multicolumn{1}{l}{I7} & \multicolumn{1}{l}{I8} & \multicolumn{1}{l}{I9} & \multicolumn{1}{l|}{I10} & \multicolumn{1}{l}{U1} & \multicolumn{1}{l}{U2} & \multicolumn{1}{l}{U3} & \multicolumn{1}{l}{U4} & \multicolumn{1}{l}{U5} \\
    \midrule
    \multirow{15}[2]{*}{\rotatebox{90}{CLIP}} & criminal & 30.2  & 21.1  & 2.6   & 1.7   & 8.2   & 3.8   & 1.6   & 0.0   & 0.0   & 0.0   & 0.6   & 0.0   & 3.8   & 0.0   & 0.8   & 2.3   & 2.2   & 0.1   & 1.5   & 13.7  & 25.7  & 10.0  \\
          & failure & 17.0  & 6.7   & 1.3   & 0.4   & 3.9   & 1.3   & 0.8   & 4.6   & 0.0   & 1.5   & 6.9   & 24.2  & 10.1  & 10.0  & 4.2   & 0.0   & 15.0  & 0.1   & 0.4   & 5.1   & 4.0   & 1.5  \\
          & fraudster & 81.3  & 88.6  & 42.9  & 39.4  & 53.2  & 39.5  & 34.1  & 12.0  & 3.2   & 4.6   & 11.3  & 3.6   & 29.7  & 5.0   & 18.4  & 4.6   & 60.2  & 3.0   & 30.1  & 87.1  & 100.0  & 62.3  \\
          & liar  & 44.8  & 28.0  & 4.0   & 2.0   & 11.2  & 5.2   & 3.4   & 0.0   & 0.5   & 1.5   & 3.1   & 2.7   & 10.7  & 1.7   & 2.5   & 0.0   & 16.8  & 0.3   & 1.9   & 21.6  & 25.8  & 9.0  \\
          & thief & 20.2  & 29.6  & 0.9   & 1.2   & 3.8   & 1.9   & 1.5   & 3.7   & 0.0   & 1.5   & 3.7   & 0.9   & 5.1   & 6.7   & 2.5   & 0.0   & 30.1  & 0.2   & 0.4   & 2.5   & 8.6   & 2.8  \\
          & citizen & 36.9  & 69.6  & 3.9   & 5.4   & 7.2   & 4.4   & 5.9   & 1.8   & 3.2   & 3.1   & 40.1  & 8.1   & 9.5   & 10.0  & 7.5   & 8.1   & 38.0  & 0.2   & 1.6   & 6.6   & 14.3  & 4.9  \\
          & individual & 34.0  & 43.4  & 0.9   & 1.3   & 4.3   & 2.2   & 1.1   & 0.0   & 0.0   & 0.0   & 3.1   & 1.8   & 0.0   & 0.0   & 0.8   & 1.2   & 4.0   & 0.0   & 0.0   & 0.0   & 0.0   & 0.0  \\
          & person & 95.7  & 100.0  & 85.9  & 61.6  & 74.4  & 59.7  & 72.2  & 8.3   & 3.7   & 1.5   & 18.2  & 5.4   & 8.2   & 8.4   & 6.7   & 4.6   & 100.0  & 1.7   & 24.4  & 82.3  & 94.8  & 52.2  \\
          & stranger & 15.9  & 8.1   & 1.0   & 0.5   & 3.0   & 0.9   & 0.9   & 0.0   & 0.0   & 1.5   & 1.9   & 0.9   & 0.6   & 0.0   & 0.8   & 0.0   & 6.6   & 0.0   & 0.2   & 0.2   & 0.1   & 0.1  \\
          & worker & 0.5   & 3.0   & 0.1   & 0.1   & 0.0   & 0.1   & 0.2   & 3.7   & 0.5   & 76.6  & 9.4   & 4.5   & 0.0   & 58.5  & 0.0   & 0.0   & 10.2  & 0.0   & 0.0   & 0.0   & 0.0   & 0.0  \\
          & genius & 9.9   & 9.0   & 2.9   & 0.4   & 2.4   & 0.6   & 1.0   & 0.0   & 0.5   & 0.0   & 1.9   & 1.8   & 1.3   & 0.0   & 0.8   & 0.0   & 8.4   & 0.0   & 0.2   & 0.6   & 0.9   & 0.1  \\
          & hero  & 14.1  & 10.8  & 0.9   & 0.3   & 2.1   & 0.5   & 0.6   & 0.9   & 0.5   & 3.1   & 1.9   & 9.0   & 6.3   & 0.0   & 7.5   & 9.3   & 8.4   & 0.1   & 0.0   & 1.0   & 3.4   & 1.3  \\
          & leader & 35.0  & 41.5  & 12.5  & 5.0   & 9.0   & 4.8   & 5.6   & 3.7   & 1.6   & 4.6   & 8.1   & 7.2   & 23.3  & 3.4   & 21.7  & 18.6  & 21.7  & 0.0   & 1.3   & 14.0  & 25.8  & 8.0  \\
          & savior & 7.1   & 2.0   & 0.3   & 0.0   & 0.9   & 0.2   & 0.1   & 0.0   & 0.0   & 0.0   & 0.0   & 0.0   & 1.9   & 0.0   & 0.0   & 0.0   & 2.7   & 0.0   & 0.2   & 1.1   & 1.4   & 0.3  \\
          & winner & 58.7  & 47.8  & 16.4  & 3.3   & 19.0  & 8.9   & 7.0   & 11.1  & 0.5   & 0.0   & 6.9   & 3.6   & 14.5  & 3.4   & 15.9  & 2.3   & 78.3  & 0.4   & 9.0   & 33.3  & 44.4  & 18.0  \\
    \midrule
    \multirow{15}[2]{*}{\rotatebox{90}{ALIGN}} & criminal & 11.9  & 16.3  & 0.8   & 2.9   & 8.8   & 3.8   & 1.4   & 0.0   & 0.0   & 0.0   & 0.0   & 0.0   & 0.9   & 0.0   & 0.0   & 1.6   & 0.6   & 0.2   & 1.2   & 6.9   & 6.3   & 1.1  \\
          & failure & 5.7   & 2.9   & 0.3   & 0.9   & 3.0   & 2.0   & 0.5   & 0.0   & 0.0   & 2.1   & 0.9   & 0.0   & 1.8   & 9.3   & 0.0   & 0.0   & 0.6   & 0.2   & 0.3   & 1.7   & 2.1   & 0.3  \\
          & fraudster & 41.2  & 27.7  & 1.1   & 4.4   & 25.0  & 11.3  & 2.0   & 0.0   & 0.0   & 0.0   & 0.0   & 0.0   & 0.0   & 0.0   & 0.0   & 0.0   & 1.9   & 0.1   & 0.7   & 4.0   & 5.3   & 1.3  \\
          & liar  & 7.6   & 2.7   & 0.6   & 1.1   & 4.5   & 2.2   & 0.5   & 0.0   & 0.0   & 0.0   & 0.0   & 0.0   & 1.8   & 0.0   & 0.0   & 0.0   & 0.6   & 0.3   & 0.4   & 7.3   & 6.6   & 1.5  \\
          & thief & 4.2   & 4.8   & 0.3   & 0.8   & 2.5   & 0.9   & 0.4   & 0.0   & 0.0   & 0.0   & 0.0   & 0.0   & 0.0   & 0.0   & 0.0   & 0.0   & 0.6   & 0.1   & 0.2   & 1.1   & 0.6   & 0.3  \\
          & citizen & 60.5  & 62.7  & 2.5   & 9.3   & 34.1  & 22.5  & 3.8   & 1.3   & 2.2   & 2.1   & 4.4   & 1.3   & 27.3  & 7.0   & 0.0   & 1.6   & 6.8   & 0.8   & 7.6   & 50.3  & 83.2  & 29.5  \\
          & individual & 99.1  & 97.8  & 10.9  & 49.4  & 87.9  & 76.1  & 17.2  & 5.2   & 10.4  & 0.0   & 3.5   & 1.3   & 13.2  & 7.0   & 0.0   & 3.2   & 14.2  & 1.5   & 28.4  & 84.1  & 92.6  & 35.5  \\
          & person & 100.0  & 98.4  & 14.9  & 62.3  & 92.5  & 81.0  & 23.9  & 5.2   & 3.7   & 0.0   & 3.5   & 0.0   & 16.7  & 4.7   & 0.0   & 1.6   & 14.8  & 3.5   & 39.5  & 92.8  & 98.2  & 48.8  \\
          & stranger & 57.8  & 44.3  & 2.5   & 7.5   & 29.9  & 14.8  & 3.0   & 0.0   & 1.5   & 0.0   & 3.5   & 2.5   & 5.3   & 4.7   & 0.0   & 0.0   & 11.1  & 0.6   & 9.7   & 55.3  & 57.2  & 11.9  \\
          & worker & 3.1   & 4.9   & 0.1   & 0.3   & 1.6   & 0.8   & 0.2   & 1.3   & 0.0   & 83.2  & 1.7   & 1.3   & 0.9   & 35.0  & 0.0   & 0.0   & 3.7   & 0.1   & 0.3   & 4.1   & 2.7   & 0.0  \\
          & genius & 70.2  & 69.1  & 3.8   & 12.9  & 41.9  & 26.9  & 5.4   & 0.0   & 0.7   & 0.0   & 0.0   & 0.0   & 8.8   & 0.0   & 0.0   & 0.0   & 5.6   & 0.2   & 4.5   & 33.6  & 47.8  & 14.7  \\
          & hero  & 35.7  & 38.6  & 1.2   & 4.3   & 16.0  & 8.0   & 2.1   & 0.0   & 1.5   & 0.0   & 0.9   & 0.0   & 1.8   & 0.0   & 0.0   & 0.0   & 1.9   & 0.1   & 0.9   & 6.5   & 5.6   & 1.1  \\
          & leader & 48.0  & 42.0  & 1.2   & 6.6   & 24.2  & 17.4  & 2.1   & 2.6   & 3.0   & 0.0   & 1.7   & 1.3   & 21.1  & 0.0   & 0.0   & 0.0   & 5.6   & 0.2   & 4.6   & 37.8  & 63.1  & 17.8  \\
          & savior & 1.6   & 2.1   & 0.1   & 0.2   & 1.0   & 0.2   & 0.1   & 0.0   & 0.0   & 0.0   & 0.0   & 0.0   & 0.0   & 0.0   & 0.0   & 0.0   & 0.6   & 0.1   & 0.3   & 4.6   & 2.1   & 0.3  \\
          & winner & 25.6  & 26.9  & 0.4   & 1.7   & 8.3   & 5.9   & 0.8   & 1.3   & 0.7   & 0.0   & 0.9   & 0.0   & 12.3  & 0.0   & 0.0   & 0.0   & 5.6   & 0.1   & 0.7   & 14.7  & 23.4  & 3.8  \\
    \midrule
    \multirow{15}[2]{*}{\rotatebox{90}{BridgeTower}} & criminal & 2.6   & 2.6   & 0.0   & 0.3   & 2.4   & 0.9   & 0.1   & 0.0   & 0.0   & 0.0   & 0.3   & 0.0   & 0.0   & 0.0   & 0.0   & 0.0   & 0.7   & 0.0   & 0.0   & 0.6   & 1.3   & 0.4  \\
          & failure & 2.4   & 0.5   & 0.0   & 0.3   & 1.8   & 0.7   & 0.1   & 0.5   & 0.5   & 0.8   & 3.1   & 0.0   & 0.9   & 4.9   & 0.0   & 0.0   & 6.9   & 0.1   & 0.0   & 0.2   & 0.1   & 0.1  \\
          & fraudster & 4.4   & 2.9   & 0.2   & 0.7   & 4.8   & 2.0   & 0.3   & 0.0   & 0.0   & 0.0   & 1.2   & 0.0   & 0.3   & 1.6   & 0.0   & 0.0   & 2.4   & 0.0   & 0.0   & 1.1   & 2.0   & 0.9  \\
          & liar  & 43.9  & 12.7  & 1.8   & 4.6   & 24.9  & 20.6  & 2.1   & 0.0   & 0.8   & 1.5   & 5.8   & 0.0   & 4.0   & 2.5   & 0.0   & 3.4   & 18.7  & 0.2   & 3.0   & 33.6  & 16.5  & 1.5  \\
          & thief & 3.8   & 1.3   & 0.1   & 0.3   & 3.2   & 1.0   & 0.1   & 0.0   & 0.0   & 0.0   & 1.5   & 0.0   & 0.3   & 1.6   & 0.0   & 0.0   & 2.0   & 0.0   & 0.0   & 0.4   & 0.6   & 0.3  \\
          & citizen & 23.7  & 43.2  & 1.6   & 20.0  & 25.2  & 35.7  & 4.8   & 2.3   & 2.6   & 3.8   & 8.0   & 0.0   & 18.6  & 6.6   & 0.4   & 13.1  & 9.6   & 0.0   & 0.9   & 14.4  & 36.1  & 12.0  \\
          & individual & 99.2  & 91.7  & 18.9  & 76.0  & 95.4  & 90.5  & 27.5  & 1.8   & 5.5   & 0.8   & 7.1   & 0.0   & 17.0  & 4.1   & 0.4   & 9.7   & 20.0  & 1.1   & 38.5  & 97.5  & 100.0  & 66.1  \\
          & person & 99.8  & 87.1  & 16.7  & 66.2  & 93.7  & 88.2  & 24.5  & 0.9   & 5.2   & 3.0   & 7.7   & 0.0   & 18.0  & 4.9   & 0.4   & 8.0   & 21.5  & 1.0   & 30.3  & 94.7  & 96.4  & 46.1  \\
          & stranger & 94.3  & 67.4  & 16.8  & 55.8  & 85.8  & 78.2  & 23.9  & 0.0   & 1.8   & 1.5   & 6.1   & 0.4   & 5.6   & 2.5   & 0.0   & 1.7   & 13.7  & 0.6   & 22.2  & 88.0  & 83.1  & 35.6  \\
          & worker & 0.9   & 1.1   & 0.0   & 0.2   & 0.4   & 1.2   & 0.2   & 2.3   & 3.4   & 100.0  & 8.0   & 0.4   & 1.9   & 25.4  & 0.0   & 0.0   & 6.5   & 0.0   & 0.0   & 0.1   & 0.2   & 0.0  \\
          & genius & 28.9  & 11.6  & 1.7   & 7.1   & 22.5  & 14.9  & 2.1   & 0.0   & 2.3   & 0.0   & 2.5   & 0.0   & 1.5   & 1.6   & 0.0   & 0.0   & 4.3   & 0.1   & 0.4   & 15.8  & 16.7  & 3.4  \\
          & hero  & 13.2  & 4.8   & 0.7   & 1.9   & 7.5   & 8.2   & 1.3   & 0.0   & 1.0   & 0.0   & 2.8   & 0.9   & 1.2   & 0.8   & 0.8   & 6.8   & 5.4   & 0.0   & 0.9   & 8.3   & 4.9   & 1.0  \\
          & leader & 32.7  & 38.4  & 2.1   & 30.3  & 28.3  & 32.2  & 5.3   & 2.3   & 7.0   & 6.8   & 7.4   & 0.9   & 45.9  & 4.1   & 2.5   & 22.2  & 17.6  & 0.2   & 6.0   & 52.1  & 67.7  & 22.0  \\
          & savior & 11.8  & 7.4   & 1.1   & 3.4   & 9.4   & 5.8   & 1.6   & 0.0   & 1.8   & 0.0   & 5.2   & 0.9   & 3.1   & 2.5   & 0.0   & 1.1   & 15.6  & 0.0   & 0.4   & 4.9   & 1.7   & 0.5  \\
          & winner & 10.7  & 5.6   & 0.4   & 2.1   & 3.5   & 7.3   & 0.8   & 2.3   & 2.3   & 0.8   & 4.0   & 0.0   & 9.3   & 0.8   & 0.8   & 5.1   & 8.5   & 0.1   & 1.3   & 12.6  & 10.6  & 3.1  \\
    \midrule
    \multirow{15}[2]{*}{\rotatebox{90}{OWLv2}} & criminal & 0.0   & 0.0   & 0.0   & 0.0   & 0.0   & 0.0   & 0.0   & 48.8  & 82.0  & 67.5  & 44.5  & 61.8  & 92.7  & 23.8  & 50.6  & 94.6  & 68.2  & 18.3  & 66.6  & 79.0  & 71.1  & 70.9  \\
          & failure & 0.1   & 0.1   & 0.0   & 0.0   & 0.0   & 0.0   & 0.1   & 65.7  & 97.3  & 91.4  & 70.9  & 75.0  & 81.7  & 42.2  & 67.5  & 68.6  & 80.6  & 96.4  & 99.0  & 98.5  & 97.0  & 95.3  \\
          & fraudster & 0.0   & 0.0   & 0.0   & 0.0   & 0.0   & 0.0   & 0.0   & 20.3  & 48.8  & 14.1  & 2.9   & 13.2  & 67.2  & 2.3   & 28.0  & 31.4  & 33.5  & 24.2  & 79.8  & 92.6  & 92.0  & 88.9  \\
          & liar  & 67.1  & 95.7  & 50.8  & 71.4  & 68.9  & 64.6  & 76.2  & 78.0  & 100.0  & 73.8  & 76.9  & 65.5  & 98.8  & 26.9  & 52.9  & 88.3  & 94.6  & 100.0  & 100.0  & 100.0  & 100.0  & 100.0  \\
          & thief & 0.0   & 0.0   & 0.0   & 0.0   & 0.0   & 0.0   & 0.0   & 15.7  & 17.1  & 26.7  & 12.9  & 34.2  & 41.4  & 8.4   & 15.7  & 43.6  & 21.9  & 1.2   & 5.9   & 6.2   & 5.7   & 8.7  \\
          & citizen & 0.0   & 0.0   & 0.0   & 0.0   & 0.0   & 0.0   & 0.0   & 8.5   & 9.3   & 19.7  & 5.2   & 10.3  & 9.3   & 0.8   & 12.3  & 34.0  & 14.0  & 0.1   & 0.6   & 0.3   & 0.4   & 1.2  \\
          & individual & 0.0   & 0.0   & 0.0   & 0.0   & 0.0   & 0.0   & 0.0   & 41.6  & 48.1  & 50.6  & 56.6  & 44.1  & 68.1  & 12.3  & 29.5  & 65.9  & 55.2  & 0.7   & 2.1   & 1.7   & 2.3   & 3.7  \\
          & person & 0.0   & 0.0   & 0.0   & 0.0   & 0.0   & 0.0   & 0.0   & 44.5  & 42.2  & 61.8  & 71.8  & 60.5  & 65.2  & 32.2  & 32.2  & 71.2  & 53.2  & 0.9   & 1.8   & 1.3   & 1.7   & 3.3  \\
          & stranger & 0.3   & 0.4   & 0.2   & 0.2   & 0.1   & 0.4   & 0.3   & 39.0  & 73.7  & 52.0  & 38.2  & 42.0  & 73.3  & 13.0  & 39.1  & 65.4  & 62.7  & 81.2  & 99.3  & 100.0  & 100.0  & 100.0  \\
          & worker & 0.0   & 0.0   & 0.0   & 0.0   & 0.0   & 0.0   & 0.0   & 1.7   & 1.0   & 15.5  & 2.6   & 13.2  & 1.4   & 0.8   & 0.4   & 1.1   & 0.4   & 0.0   & 0.0   & 0.0   & 0.0   & 0.0  \\
          & genius & 0.0   & 0.0   & 0.0   & 0.0   & 0.0   & 0.0   & 0.0   & 48.8  & 96.8  & 35.1  & 19.2  & 24.3  & 97.0  & 6.9   & 44.9  & 44.1  & 78.8  & 17.3  & 61.1  & 70.6  & 65.8  & 66.4  \\
          & hero  & 0.0   & 0.0   & 0.0   & 0.0   & 0.0   & 0.0   & 0.0   & 6.4   & 5.1   & 31.6  & 1.7   & 28.8  & 7.5   & 1.5   & 14.2  & 52.6  & 6.1   & 0.0   & 0.2   & 0.1   & 0.0   & 0.0  \\
          & leader & 0.0   & 0.0   & 0.0   & 0.0   & 0.0   & 0.0   & 0.0   & 17.0  & 31.5  & 17.6  & 8.3   & 13.6  & 64.3  & 1.5   & 24.6  & 33.0  & 27.8  & 0.9   & 3.8   & 3.3   & 4.5   & 7.4  \\
          & savior & 0.0   & 0.0   & 0.0   & 0.0   & 0.0   & 0.0   & 0.0   & 0.8   & 0.7   & 0.7   & 0.6   & 0.4   & 1.4   & 0.0   & 0.0   & 0.5   & 1.6   & 0.0   & 0.0   & 0.0   & 0.0   & 0.0  \\
          & winner & 0.0   & 0.0   & 0.0   & 0.0   & 0.0   & 0.0   & 0.0   & 0.0   & 0.5   & 1.4   & 0.0   & 0.8   & 0.0   & 0.0   & 0.4   & 1.1   & 1.6   & 0.0   & 0.0   & 0.0   & 0.0   & 0.0  \\
    \bottomrule
    \end{tabular}}%
  \label{tab:woman_norm}%
\end{table}%

\begin{table}[htbp]
  \centering
  \caption{\textbf{Normalized probabilities for the ``man" group in Fig. 3 A-D.}
Normalized probabilities of the ``man" group being predicted as probes across extended datasets (UTKFACE, FAIRFACE, IDENPROF). The heatmap values in Fig. 3 A-D are derived from the differences between these values and those in Table~\ref{tab:woman_norm}. The class names in this table are abbreviated as follows: F1: White\_man, F2: Black\_man, F3: East Asian\_man, F4: Indian\_man, F5: Middle Eastern\_man, F6: Latino\_Hispanic\_man, F7: Southeast Asian\_man; I1: chef\_man, I2: doctor\_man, I3: engineer\_man, I4: farmer\_man, I5: firefighter\_man, I6: judge\_man, I7: mechanic\_man, I8: pilot\_man, I9: police\_man, I10: waiter\_man; U1: child\_man, U2:teenager\_man, U3:young adult\_man, U4:middle aged\_man, U5:elderly\_man.}
    \resizebox{16.5cm}{!}{\begin{tabular}{cl|rrrrrrr|rrrrrrrrrr|rrrrr}
    \toprule
          &       & \multicolumn{7}{c|}{FAIRFACE}                         & \multicolumn{10}{c|}{IDENPROF}                                                & \multicolumn{5}{c}{UTKFACE} \\
\cmidrule{3-24}          &       & \multicolumn{1}{l}{F1} & \multicolumn{1}{l}{F2} & \multicolumn{1}{l}{F3} & \multicolumn{1}{l}{F4} & \multicolumn{1}{l}{F5} & \multicolumn{1}{l}{F6} & \multicolumn{1}{l|}{F7} & \multicolumn{1}{l}{I1} & \multicolumn{1}{l}{I2} & \multicolumn{1}{l}{I3} & \multicolumn{1}{l}{I4} & \multicolumn{1}{l}{I5} & \multicolumn{1}{l}{I6} & \multicolumn{1}{l}{I7} & \multicolumn{1}{l}{I8} & \multicolumn{1}{l}{I9} & \multicolumn{1}{l|}{I10} & \multicolumn{1}{l}{U1} & \multicolumn{1}{l}{U2} & \multicolumn{1}{l}{U3} & \multicolumn{1}{l}{U4} & \multicolumn{1}{l}{U5} \\
    \midrule
    \multirow{15}[2]{*}{\rotatebox{90}{CLIP}} & criminal & 32.2  & 37.2  & 8.9   & 3.9   & 11.6  & 3.9   & 3.5   & 0.3   & 0.0   & 0.6   & 1.6   & 0.0   & 19.0  & 0.6   & 1.4   & 3.4   & 2.2   & 0.2   & 7.0   & 31.3  & 23.7  & 17.0  \\
          & failure & 10.8  & 8.4   & 2.3   & 0.4   & 2.2   & 0.5   & 0.7   & 1.3   & 2.7   & 2.6   & 4.4   & 7.7   & 17.8  & 8.8   & 5.1   & 1.2   & 6.0   & 0.2   & 0.4   & 6.1   & 2.6   & 0.8  \\
          & fraudster & 80.2  & 88.2  & 57.6  & 56.4  & 60.3  & 34.6  & 41.2  & 7.6   & 4.9   & 2.9   & 20.7  & 0.7   & 53.5  & 7.7   & 14.1  & 7.7   & 45.4  & 3.2   & 29.1  & 91.5  & 96.7  & 76.3  \\
          & liar  & 21.8  & 14.7  & 4.2   & 1.5   & 5.2   & 2.0   & 1.9   & 0.0   & 0.4   & 0.3   & 0.4   & 0.3   & 19.0  & 0.0   & 2.1   & 0.3   & 2.2   & 0.1   & 0.6   & 7.9   & 6.2   & 4.0  \\
          & thief & 16.3  & 32.9  & 1.7   & 2.0   & 3.7   & 1.1   & 1.4   & 0.3   & 0.0   & 0.0   & 3.6   & 0.3   & 9.1   & 2.8   & 1.7   & 1.2   & 6.5   & 0.0   & 3.3   & 13.8  & 6.7   & 4.1  \\
          & citizen & 31.6  & 62.7  & 9.2   & 11.0  & 7.7   & 3.0   & 5.5   & 0.7   & 2.2   & 2.9   & 37.4  & 4.3   & 15.4  & 4.0   & 4.1   & 13.9  & 9.2   & 0.0   & 1.0   & 9.7   & 8.4   & 5.3  \\
          & individual & 22.1  & 33.7  & 1.3   & 1.1   & 1.8   & 0.7   & 0.7   & 0.0   & 0.0   & 0.0   & 1.6   & 0.0   & 0.0   & 0.0   & 0.0   & 0.3   & 0.0   & 0.0   & 0.0   & 0.0   & 0.1   & 0.0  \\
          & person & 80.1  & 94.2  & 86.5  & 61.5  & 51.7  & 31.0  & 68.0  & 1.0   & 2.2   & 0.9   & 22.3  & 1.3   & 10.3  & 3.1   & 2.4   & 2.5   & 29.2  & 0.2   & 8.0   & 65.3  & 38.7  & 30.9  \\
          & stranger & 6.2   & 7.3   & 0.8   & 0.6   & 1.9   & 0.3   & 0.3   & 0.0   & 0.4   & 0.0   & 1.2   & 0.3   & 0.0   & 0.3   & 0.0   & 0.0   & 1.6   & 0.0   & 0.0   & 0.3   & 0.1   & 0.0  \\
          & worker & 0.7   & 5.9   & 0.5   & 0.6   & 0.3   & 0.4   & 0.6   & 0.7   & 0.9   & 71.2  & 14.7  & 5.3   & 0.0   & 38.2  & 0.0   & 0.3   & 3.8   & 0.0   & 0.0   & 0.2   & 0.1   & 0.0  \\
          & genius & 46.9  & 46.5  & 38.7  & 9.0   & 15.3  & 5.9   & 12.3  & 2.7   & 2.2   & 0.3   & 5.6   & 0.3   & 12.7  & 2.8   & 1.0   & 0.6   & 26.0  & 0.0   & 3.9   & 26.3  & 20.8  & 15.8  \\
          & hero  & 36.1  & 42.0  & 11.1  & 2.7   & 8.4   & 2.0   & 3.8   & 0.0   & 2.2   & 2.6   & 10.4  & 16.0  & 19.4  & 5.7   & 8.6   & 8.6   & 7.6   & 0.2   & 3.3   & 26.6  & 19.7  & 13.8  \\
          & leader & 41.9  & 57.8  & 29.0  & 17.1  & 19.1  & 8.5   & 11.8  & 4.0   & 1.8   & 8.1   & 10.4  & 1.0   & 49.5  & 3.4   & 20.9  & 19.4  & 13.0  & 0.0   & 2.1   & 24.6  & 26.8  & 13.5  \\
          & savior & 7.8   & 6.2   & 0.7   & 0.2   & 0.9   & 0.2   & 0.2   & 0.0   & 0.0   & 0.0   & 0.0   & 0.0   & 4.8   & 0.0   & 0.4   & 0.0   & 1.1   & 0.0   & 0.6   & 3.6   & 1.5   & 0.9  \\
          & winner & 47.9  & 53.2  & 35.4  & 5.1   & 12.4  & 4.7   & 11.4  & 4.0   & 2.2   & 0.9   & 13.1  & 1.3   & 18.6  & 6.8   & 7.6   & 1.5   & 44.3  & 0.8   & 8.2   & 39.5  & 21.6  & 10.3  \\
    \midrule
    \multirow{15}[2]{*}{\rotatebox{90}{ALIGN}} & criminal & 48.6  & 49.9  & 2.2   & 9.6   & 28.3  & 13.2  & 4.6   & 0.0   & 0.6   & 0.0   & 0.5   & 0.0   & 0.5   & 2.4   & 1.4   & 11.2  & 1.5   & 0.7   & 14.0  & 70.7  & 64.9  & 17.0  \\
          & failure & 5.9   & 3.6   & 0.4   & 0.7   & 1.8   & 1.2   & 0.5   & 0.9   & 1.9   & 0.4   & 1.1   & 1.4   & 5.0   & 1.2   & 0.0   & 0.0   & 0.8   & 0.7   & 0.4   & 12.9  & 12.1  & 1.1  \\
          & fraudster & 69.5  & 50.8  & 2.7   & 11.5  & 44.6  & 20.0  & 4.3   & 0.5   & 1.3   & 0.0   & 0.5   & 0.0   & 1.6   & 0.8   & 0.0   & 3.9   & 1.5   & 0.2   & 6.9   & 51.9  & 48.1  & 11.6  \\
          & liar  & 12.5  & 5.2   & 0.8   & 1.1   & 5.4   & 2.3   & 0.7   & 0.0   & 0.0   & 0.0   & 0.0   & 0.0   & 3.9   & 0.0   & 0.0   & 0.4   & 2.3   & 0.4   & 2.0   & 28.7  & 27.1  & 7.0  \\
          & thief & 20.6  & 18.5  & 0.6   & 2.3   & 9.8   & 4.5   & 1.3   & 0.0   & 0.0   & 0.0   & 0.0   & 0.0   & 0.0   & 1.2   & 0.0   & 1.3   & 0.0   & 0.3   & 5.9   & 28.8  & 16.7  & 4.0  \\
          & citizen & 66.5  & 64.6  & 3.0   & 11.0  & 33.9  & 16.0  & 5.6   & 3.2   & 2.5   & 0.8   & 2.8   & 0.5   & 36.4  & 2.0   & 1.4   & 1.7   & 5.3   & 0.9   & 10.2  & 76.2  & 98.1  & 70.6  \\
          & individual & 88.3  & 86.5  & 8.4   & 21.2  & 57.3  & 33.3  & 12.3  & 0.0   & 6.3   & 0.0   & 0.5   & 0.5   & 11.6  & 2.8   & 0.5   & 0.4   & 3.0   & 1.3   & 18.1  & 85.4  & 93.6  & 45.9  \\
          & person & 96.2  & 93.4  & 13.9  & 40.5  & 79.5  & 52.7  & 20.9  & 0.5   & 4.4   & 0.0   & 0.5   & 0.5   & 14.3  & 2.4   & 0.5   & 0.4   & 5.3   & 2.4   & 28.7  & 93.9  & 100.0  & 71.8  \\
          & stranger & 59.2  & 48.4  & 2.6   & 7.5   & 26.9  & 11.6  & 4.0   & 0.5   & 2.5   & 0.0   & 2.8   & 1.4   & 5.5   & 2.8   & 0.9   & 0.9   & 6.0   & 0.8   & 12.4  & 78.6  & 65.5  & 17.9  \\
          & worker & 9.7   & 9.9   & 0.5   & 0.7   & 3.4   & 1.3   & 0.5   & 0.5   & 0.6   & 100.0  & 2.8   & 0.9   & 0.5   & 16.3  & 0.0   & 0.0   & 0.8   & 0.1   & 1.0   & 37.7  & 36.0  & 2.6  \\
          & genius & 83.1  & 78.8  & 7.1   & 19.9  & 54.7  & 29.3  & 10.3  & 0.5   & 2.5   & 0.0   & 1.1   & 0.9   & 14.9  & 0.8   & 0.5   & 0.0   & 10.6  & 0.9   & 12.6  & 77.5  & 86.2  & 41.9  \\
          & hero  & 77.1  & 72.5  & 3.6   & 13.6  & 45.9  & 22.5  & 6.8   & 0.0   & 1.9   & 0.0   & 0.5   & 0.0   & 1.6   & 0.8   & 0.9   & 3.9   & 5.3   & 0.2   & 9.8   & 71.3  & 63.9  & 10.1  \\
          & leader & 58.1  & 47.1  & 1.8   & 9.3   & 30.2  & 14.6  & 3.3   & 1.8   & 1.9   & 0.0   & 0.0   & 0.0   & 34.8  & 0.4   & 0.5   & 0.4   & 3.0   & 0.3   & 6.9   & 62.6  & 87.3  & 41.2  \\
          & savior & 2.7   & 3.1   & 0.1   & 0.2   & 1.3   & 0.4   & 0.0   & 0.0   & 0.0   & 0.0   & 0.0   & 0.0   & 4.4   & 0.0   & 0.0   & 0.0   & 1.5   & 0.1   & 1.2   & 24.9  & 22.5  & 1.7  \\
          & winner & 13.7  & 16.8  & 0.4   & 0.7   & 3.1   & 1.5   & 0.6   & 0.5   & 1.9   & 0.0   & 0.5   & 0.9   & 7.7   & 0.4   & 0.5   & 0.4   & 3.8   & 0.0   & 1.8   & 35.7  & 31.2  & 2.0  \\
    \midrule
    \multirow{15}[2]{*}{\rotatebox{90}{BridgeTower}} & criminal & 20.5  & 29.3  & 1.7   & 14.1  & 25.5  & 19.0  & 4.3   & 0.2   & 0.2   & 0.0   & 0.8   & 0.0   & 0.4   & 0.1   & 0.0   & 1.4   & 0.5   & 0.2   & 5.5   & 35.9  & 25.0  & 6.6  \\
          & failure & 2.1   & 0.5   & 0.1   & 0.4   & 1.6   & 0.9   & 0.1   & 0.0   & 0.2   & 0.4   & 1.6   & 0.2   & 0.4   & 1.3   & 0.3   & 0.5   & 0.8   & 0.1   & 0.0   & 0.9   & 0.8   & 0.2  \\
          & fraudster & 28.8  & 23.4  & 2.8   & 21.1  & 33.7  & 25.3  & 5.5   & 0.2   & 0.2   & 0.0   & 1.2   & 0.0   & 0.4   & 0.6   & 0.5   & 0.2   & 0.5   & 0.2   & 4.1   & 27.4  & 27.8  & 8.5  \\
          & liar  & 28.5  & 6.2   & 1.4   & 6.4   & 18.5  & 13.0  & 1.2   & 0.3   & 0.2   & 0.0   & 2.2   & 0.2   & 1.9   & 0.8   & 0.0   & 0.5   & 4.2   & 0.1   & 0.8   & 19.9  & 11.9  & 0.4  \\
          & thief & 17.7  & 14.2  & 1.3   & 10.1  & 21.5  & 13.0  & 2.6   & 0.0   & 0.4   & 0.0   & 1.4   & 0.0   & 0.0   & 0.4   & 0.0   & 0.3   & 0.5   & 0.2   & 2.9   & 21.3  & 13.0  & 2.6  \\
          & citizen & 24.6  & 40.9  & 2.0   & 25.9  & 28.4  & 32.1  & 6.1   & 0.8   & 2.0   & 1.3   & 3.5   & 0.0   & 18.1  & 0.6   & 1.3   & 7.1   & 1.6   & 0.2   & 2.2   & 29.6  & 42.3  & 18.1  \\
          & individual & 96.5  & 90.2  & 20.1  & 79.6  & 90.6  & 82.8  & 27.4  & 0.8   & 2.0   & 0.4   & 4.3   & 0.0   & 8.4   & 1.0   & 0.5   & 2.1   & 5.8   & 0.7   & 19.2  & 92.3  & 94.2  & 49.2  \\
          & person & 100.0  & 91.2  & 21.6  & 81.4  & 93.3  & 86.2  & 29.8  & 1.0   & 2.4   & 0.7   & 5.5   & 0.0   & 8.8   & 1.4   & 0.7   & 2.9   & 7.7   & 0.5   & 16.1  & 90.4  & 88.9  & 38.8  \\
          & stranger & 94.7  & 75.4  & 22.6  & 76.3  & 88.6  & 78.3  & 32.3  & 0.3   & 1.1   & 0.4   & 5.1   & 0.0   & 1.9   & 1.0   & 0.3   & 0.5   & 4.5   & 0.5   & 14.7  & 85.9  & 72.6  & 25.3  \\
          & worker & 3.5   & 2.1   & 0.6   & 2.6   & 2.8   & 4.2   & 1.3   & 1.9   & 3.1   & 91.9  & 4.5   & 0.3   & 1.2   & 9.2   & 0.7   & 0.0   & 0.8   & 0.1   & 0.0   & 0.7   & 1.6   & 0.2  \\
          & genius & 78.6  & 40.5  & 12.2  & 52.3  & 70.7  & 53.8  & 12.9  & 0.3   & 2.4   & 0.0   & 0.8   & 0.0   & 2.7   & 0.4   & 0.2   & 0.2   & 1.3   & 0.2   & 4.5   & 65.2  & 71.0  & 21.1  \\
          & hero  & 36.9  & 21.6  & 4.1   & 12.5  & 30.0  & 26.3  & 6.8   & 0.0   & 1.3   & 0.3   & 2.7   & 0.2   & 2.3   & 0.7   & 1.2   & 5.9   & 1.9   & 0.2   & 3.1   & 47.8  & 29.6  & 8.6  \\
          & leader & 57.4  & 51.8  & 9.6   & 61.6  & 58.5  & 52.2  & 17.3  & 2.3   & 5.1   & 2.7   & 4.7   & 0.2   & 45.7  & 1.1   & 3.7   & 13.8  & 8.5   & 0.3   & 6.5   & 70.9  & 87.2  & 41.7  \\
          & savior & 44.2  & 26.2  & 3.6   & 13.1  & 33.9  & 22.8  & 5.0   & 0.5   & 1.8   & 0.1   & 3.3   & 0.2   & 2.7   & 0.8   & 0.3   & 1.1   & 4.0   & 0.1   & 2.0   & 32.1  & 17.8  & 3.0  \\
          & winner & 12.3  & 5.4   & 1.3   & 6.0   & 6.9   & 8.9   & 1.5   & 1.5   & 0.4   & 0.3   & 1.2   & 0.0   & 5.4   & 1.0   & 0.3   & 0.8   & 2.4   & 0.0   & 0.8   & 13.5  & 11.9  & 3.6  \\
    \midrule
    \multirow{15}[2]{*}{\rotatebox{90}{OWLv2}} & criminal & 0.1   & 0.0   & 0.1   & 0.0   & 0.0   & 0.0   & 0.0   & 52.3  & 82.4  & 59.7  & 43.7  & 68.9  & 93.2  & 15.2  & 44.6  & 90.2  & 73.4  & 16.7  & 43.0  & 20.6  & 8.9   & 15.0  \\
          & failure & 0.1   & 0.0   & 0.0   & 0.0   & 0.1   & 0.0   & 0.1   & 70.8  & 93.9  & 89.1  & 80.2  & 81.5  & 81.6  & 30.3  & 62.2  & 73.5  & 77.4  & 97.2  & 96.2  & 82.9  & 81.6  & 82.4  \\
          & fraudster & 0.0   & 0.1   & 0.0   & 0.0   & 0.0   & 0.0   & 0.0   & 18.6  & 52.7  & 9.8   & 6.9   & 13.9  & 74.3  & 2.7   & 18.4  & 23.2  & 34.7  & 20.0  & 61.0  & 48.6  & 35.9  & 50.0  \\
          & liar  & 93.7  & 100.0  & 77.1  & 70.0  & 84.0  & 89.1  & 88.3  & 75.6  & 95.8  & 61.1  & 70.3  & 69.2  & 96.7  & 12.7  & 38.9  & 74.5  & 85.4  & 100.0  & 100.0  & 100.0  & 100.0  & 100.0  \\
          & thief & 0.0   & 0.0   & 0.0   & 0.0   & 0.0   & 0.0   & 0.0   & 15.9  & 14.4  & 15.5  & 12.4  & 39.1  & 33.1  & 5.4   & 9.6   & 35.3  & 21.6  & 0.6   & 2.2   & 0.8   & 0.3   & 0.4  \\
          & citizen & 0.0   & 0.0   & 0.0   & 0.0   & 0.0   & 0.0   & 0.0   & 3.9   & 4.5   & 9.0   & 4.8   & 10.1  & 7.6   & 0.8   & 5.4   & 22.8  & 7.2   & 0.0   & 0.0   & 0.0   & 0.0   & 0.0  \\
          & individual & 0.0   & 0.0   & 0.0   & 0.0   & 0.0   & 0.0   & 0.0   & 39.7  & 38.7  & 37.7  & 38.7  & 48.6  & 58.0  & 5.4   & 20.6  & 55.4  & 49.6  & 0.1   & 0.6   & 0.1   & 0.0   & 0.0  \\
          & person & 0.0   & 0.0   & 0.0   & 0.0   & 0.0   & 0.0   & 0.0   & 39.3  & 30.7  & 54.7  & 49.0  & 68.3  & 52.2  & 15.9  & 23.3  & 59.8  & 45.7  & 0.3   & 0.8   & 0.1   & 0.0   & 0.0  \\
          & stranger & 1.2   & 1.0   & 0.7   & 1.0   & 1.0   & 0.7   & 0.8   & 41.2  & 72.1  & 40.2  & 38.5  & 41.7  & 75.4  & 6.8   & 32.0  & 58.8  & 62.5  & 87.8  & 99.4  & 99.9  & 99.1  & 99.2  \\
          & worker & 0.0   & 0.0   & 0.0   & 0.0   & 0.0   & 0.0   & 0.0   & 1.5   & 1.2   & 8.6   & 2.9   & 16.1  & 4.7   & 0.7   & 0.6   & 2.3   & 0.5   & 0.0   & 0.0   & 0.0   & 0.0   & 0.0  \\
          & genius & 0.0   & 0.0   & 0.0   & 0.0   & 0.0   & 0.0   & 0.0   & 52.3  & 91.6  & 27.7  & 23.9  & 24.4  & 97.2  & 4.6   & 36.9  & 42.1  & 78.9  & 14.9  & 46.8  & 17.8  & 10.7  & 18.7  \\
          & hero  & 0.0   & 0.0   & 0.0   & 0.0   & 0.0   & 0.0   & 0.0   & 6.1   & 5.6   & 20.3  & 3.7   & 31.6  & 8.0   & 1.2   & 10.7  & 42.4  & 5.2   & 0.0   & 0.0   & 0.0   & 0.0   & 0.0  \\
          & leader & 0.0   & 0.0   & 0.0   & 0.0   & 0.0   & 0.0   & 0.0   & 17.3  & 28.8  & 10.2  & 9.7   & 12.4  & 65.8  & 1.7   & 18.3  & 24.4  & 24.3  & 0.1   & 0.6   & 0.1   & 0.0   & 0.1  \\
          & savior & 0.0   & 0.0   & 0.0   & 0.0   & 0.0   & 0.0   & 0.0   & 0.3   & 0.4   & 0.0   & 0.2   & 1.1   & 0.2   & 0.0   & 0.0   & 0.0   & 0.3   & 0.0   & 0.0   & 0.0   & 0.0   & 0.0  \\
          & winner & 0.0   & 0.0   & 0.0   & 0.0   & 0.0   & 0.0   & 0.0   & 0.0   & 0.2   & 0.5   & 0.0   & 0.2   & 0.2   & 0.3   & 0.2   & 0.4   & 1.0   & 0.0   & 0.0   & 0.0   & 0.0   & 0.0  \\
    \bottomrule
    \end{tabular}}%
  \label{tab:man_norm}%
\end{table}%

\begin{table}[htbp]
  \centering
  \caption{\textbf{Macro average accuracy with AdaLogAdjustment in Single Bias Test (Fig. 4 A-D).} Macro average accuracy results achieved by applying AdaLogAdjustment to four models (CLIP, ALIGN, BridgeTower, OWLv2) in Single Bias Test scenarios across four datasets (CelebA, UTKFace, FairFace, IdenProf). The differences between Table~\ref{tab:single_al} and Table~\ref{tab:single_noal} produce the ``Improved macro average accuracy" values shown in Fig. 4 A-D.}
    \resizebox{16.5cm}{!}{\begin{tabular}{cl|rrrr|rrrr|rrrr|rrrr}
    \toprule
          &       & \multicolumn{4}{c|}{CelebA}   & \multicolumn{4}{c|}{FairFace} & \multicolumn{4}{c|}{IdenProf} & \multicolumn{4}{c}{UTKFace} \\
\cmidrule{3-18}          &       & \multicolumn{1}{l}{test 1} & \multicolumn{1}{l}{test 2} & \multicolumn{1}{l}{test 3} & \multicolumn{1}{l|}{Avg} & \multicolumn{1}{l}{test 1} & \multicolumn{1}{l}{test 2} & \multicolumn{1}{l}{test 3} & \multicolumn{1}{l|}{Avg} & \multicolumn{1}{l}{test 1} & \multicolumn{1}{l}{test 2} & \multicolumn{1}{l}{test 3} & \multicolumn{1}{l|}{Avg} & \multicolumn{1}{l}{test 1} & \multicolumn{1}{l}{test 2} & \multicolumn{1}{l}{test 3} & \multicolumn{1}{l}{Avg} \\
    \midrule
    \multirow{15}[2]{*}{\rotatebox{90}{CLIP}} & criminal & 98.92  & 98.91  & 98.84  & 98.89  & 61.98  & 61.47  & 61.98  & 61.81  & 92.92  & 92.74  & 92.86  & 92.84  & 70.14  & 70.47  & 70.20  & 70.27  \\
          & failure & 98.79  & 98.96  & 98.92  & 98.89  & 62.29  & 62.15  & 61.32  & 61.92  & 92.51  & 92.58  & 92.69  & 92.59  & 70.58  & 70.56  & 69.94  & 70.36  \\
          & fraudster & 98.90  & 98.92  & 98.93  & 98.92  & 61.88  & 62.10  & 61.64  & 61.87  & 93.02  & 92.52  & 92.02  & 92.52  & 69.65  & 69.79  & 69.08  & 69.51  \\
          & liar  & 98.93  & 98.90  & 98.96  & 98.93  & 62.33  & 62.13  & 61.99  & 62.15  & 92.67  & 92.84  & 93.03  & 92.85  & 70.63  & 70.83  & 69.20  & 70.22  \\
          & thief & 98.90  & 98.91  & 98.96  & 98.92  & 61.55  & 62.18  & 62.04  & 61.92  & 92.77  & 92.67  & 92.28  & 92.57  & 70.61  & 70.08  & 70.04  & 70.24  \\
          & citizen & 98.94  & 98.92  & 98.91  & 98.92  & 62.03  & 62.18  & 62.21  & 62.14  & 93.27  & 93.28  & 93.50  & 93.35  & 69.82  & 69.91  & 70.12  & 69.95  \\
          & individual & 98.90  & 98.93  & 98.90  & 98.91  & 62.09  & 61.88  & 61.77  & 61.91  & 93.22  & 93.02  & 93.22  & 93.15  & 70.53  & 70.52  & 70.38  & 70.48  \\
          & person & 98.91  & 98.91  & 98.91  & 98.91  & 61.86  & 62.33  & 61.78  & 61.99  & 92.81  & 93.03  & 92.09  & 92.64  & 69.96  & 70.22  & 70.74  & 70.31  \\
          & stranger & 98.95  & 98.92  & 98.91  & 98.93  & 62.21  & 62.04  & 61.55  & 61.93  & 93.10  & 93.39  & 93.43  & 93.31  & 70.29  & 70.27  & 70.51  & 70.36  \\
          & worker & 98.93  & 98.91  & 98.91  & 98.92  & 62.27  & 62.04  & 61.91  & 62.07  & 93.20  & 92.92  & 93.08  & 93.07  & 69.92  & 69.66  & 70.51  & 70.03  \\
          & genius & 98.92  & 98.93  & 98.91  & 98.92  & 62.32  & 62.22  & 62.14  & 62.23  & 93.33  & 92.87  & 92.82  & 93.01  & 70.62  & 70.62  & 70.75  & 70.66  \\
          & hero  & 98.90  & 98.92  & 98.91  & 98.91  & 61.96  & 62.15  & 62.22  & 62.11  & 92.89  & 93.38  & 92.18  & 92.82  & 70.55  & 70.32  & 70.50  & 70.46  \\
          & leader & 98.90  & 98.92  & 98.92  & 98.91  & 61.86  & 62.40  & 62.36  & 62.21  & 92.44  & 93.20  & 92.56  & 92.73  & 70.72  & 70.37  & 70.39  & 70.49  \\
          & savior & 98.91  & 98.90  & 98.96  & 98.92  & 61.43  & 62.01  & 61.97  & 61.80  & 93.01  & 93.02  & 93.30  & 93.11  & 70.61  & 70.48  & 70.34  & 70.48  \\
          & winner & 98.92  & 98.94  & 98.92  & 98.93  & 62.07  & 61.72  & 61.26  & 61.68  & 92.74  & 92.97  & 92.48  & 92.73  & 70.51  & 70.88  & 70.29  & 70.56  \\
    \midrule
    \multirow{15}[2]{*}{\rotatebox{90}{ALIGN}} & criminal & 98.78  & 98.85  & 98.87  & 98.83  & 50.15  & 49.27  & 49.92  & 49.78  & 95.34  & 95.22  & 95.14  & 95.23  & 53.36  & 53.41  & 53.33  & 53.37  \\
          & failure & 98.84  & 98.85  & 98.86  & 98.85  & 50.81  & 50.91  & 50.93  & 50.88  & 95.50  & 95.19  & 95.20  & 95.30  & 56.61  & 56.63  & 56.69  & 56.64  \\
          & fraudster & 98.76  & 98.79  & 98.85  & 98.80  & 49.77  & 49.49  & 49.66  & 49.64  & 95.07  & 95.33  & 95.30  & 95.23  & 54.65  & 54.63  & 54.67  & 54.65  \\
          & liar  & 98.84  & 98.77  & 98.77  & 98.79  & 50.14  & 50.26  & 50.48  & 50.29  & 95.40  & 95.32  & 95.15  & 95.29  & 55.62  & 55.61  & 55.64  & 55.62  \\
          & thief & 98.83  & 98.84  & 98.79  & 98.82  & 50.32  & 50.85  & 50.42  & 50.53  & 95.27  & 95.14  & 95.42  & 95.28  & 55.91  & 55.82  & 55.85  & 55.86  \\
          & citizen & 98.84  & 98.88  & 98.79  & 98.84  & 49.27  & 49.19  & 49.16  & 49.21  & 95.33  & 95.35  & 95.44  & 95.37  & 51.43  & 51.24  & 52.19  & 51.62  \\
          & individual & 98.79  & 98.79  & 98.80  & 98.79  & 47.43  & 47.46  & 47.27  & 47.39  & 95.22  & 95.09  & 94.93  & 95.08  & 51.90  & 52.22  & 53.18  & 52.43  \\
          & person & 98.79  & 98.79  & 98.81  & 98.80  & 45.58  & 45.54  & 45.49  & 45.54  & 95.38  & 95.28  & 94.58  & 95.08  & 53.42  & 51.77  & 52.50  & 52.56  \\
          & stranger & 98.78  & 98.78  & 98.78  & 98.78  & 49.54  & 49.63  & 49.87  & 49.68  & 95.03  & 95.18  & 95.07  & 95.09  & 51.80  & 52.05  & 51.77  & 51.87  \\
          & worker & 98.84  & 98.79  & 98.81  & 98.81  & 50.50  & 50.86  & 50.52  & 50.63  & 94.76  & 95.11  & 95.27  & 95.05  & 56.41  & 56.42  & 56.44  & 56.42  \\
          & genius & 98.73  & 98.79  & 98.70  & 98.74  & 49.25  & 49.16  & 48.94  & 49.12  & 95.48  & 95.30  & 95.13  & 95.30  & 50.96  & 52.79  & 51.18  & 51.64  \\
          & hero  & 98.86  & 98.87  & 98.83  & 98.85  & 49.24  & 50.07  & 49.88  & 49.73  & 95.13  & 95.24  & 95.20  & 95.19  & 54.67  & 54.47  & 54.47  & 54.54  \\
          & leader & 98.77  & 98.84  & 98.85  & 98.82  & 49.81  & 48.33  & 49.28  & 49.14  & 95.40  & 94.95  & 94.56  & 94.97  & 50.56  & 51.50  & 52.14  & 51.40  \\
          & savior & 98.78  & 98.81  & 98.79  & 98.79  & 50.69  & 50.82  & 50.69  & 50.73  & 95.16  & 95.33  & 95.67  & 95.39  & 56.38  & 56.47  & 56.41  & 56.42  \\
          & winner & 98.78  & 98.80  & 98.78  & 98.79  & 50.67  & 50.72  & 50.51  & 50.63  & 94.89  & 95.20  & 95.41  & 95.17  & 56.10  & 56.11  & 56.13  & 56.11  \\
    \midrule
    \multirow{15}[2]{*}{\rotatebox{90}{BridgeTower}} & criminal & 98.75  & 98.75  & 98.75  & 98.75  & 25.94  & 28.31  & 30.70  & 28.32  & 92.88  & 92.86  & 92.97  & 92.90  & 50.80  & 50.78  & 50.83  & 50.80  \\
          & failure & 98.82  & 98.83  & 98.82  & 98.82  & 27.54  & 28.48  & 30.64  & 28.89  & 92.78  & 92.74  & 92.72  & 92.75  & 52.51  & 52.43  & 52.42  & 52.45  \\
          & fraudster & 98.83  & 98.83  & 98.83  & 98.83  & 26.17  & 29.69  & 29.50  & 28.45  & 92.90  & 92.87  & 92.85  & 92.87  & 50.88  & 50.84  & 50.89  & 50.87  \\
          & liar  & 97.06  & 97.06  & 97.06  & 97.06  & 27.01  & 30.42  & 26.59  & 28.01  & 92.34  & 92.31  & 92.23  & 92.29  & 52.20  & 52.12  & 52.07  & 52.13  \\
          & thief & 98.89  & 98.89  & 98.89  & 98.89  & 29.97  & 26.11  & 26.72  & 27.60  & 92.92  & 92.98  & 92.94  & 92.95  & 51.78  & 51.66  & 51.37  & 51.60  \\
          & citizen & 98.14  & 98.14  & 98.14  & 98.14  & 30.34  & 27.16  & 28.30  & 28.60  & 91.11  & 91.13  & 91.17  & 91.14  & 50.44  & 50.45  & 50.46  & 50.45  \\
          & individual & 77.08  & 77.08  & 77.08  & 77.08  & 18.05  & 18.05  & 18.04  & 18.05  & 91.64  & 91.48  & 91.49  & 91.54  & 35.62  & 35.61  & 35.76  & 35.66  \\
          & person & 83.20  & 83.20  & 83.21  & 83.20  & 17.03  & 17.64  & 17.64  & 17.44  & 91.38  & 91.34  & 91.31  & 91.34  & 39.06  & 39.08  & 39.07  & 39.07  \\
          & stranger & 92.23  & 92.23  & 92.23  & 92.23  & 19.22  & 18.50  & 18.49  & 18.74  & 92.15  & 92.17  & 92.11  & 92.14  & 42.59  & 42.66  & 42.61  & 42.62  \\
          & worker & 99.03  & 99.03  & 99.03  & 99.03  & 30.45  & 26.92  & 26.95  & 28.11  & 86.27  & 86.37  & 86.28  & 86.31  & 52.40  & 52.32  & 52.49  & 52.40  \\
          & genius & 96.97  & 96.97  & 96.96  & 96.97  & 24.46  & 24.53  & 24.45  & 24.48  & 92.80  & 92.78  & 92.85  & 92.81  & 48.38  & 48.34  & 48.35  & 48.36  \\
          & hero  & 95.67  & 95.67  & 95.67  & 95.67  & 26.38  & 27.30  & 30.99  & 28.22  & 92.57  & 92.50  & 92.49  & 92.52  & 51.04  & 51.02  & 50.96  & 51.01  \\
          & leader & 93.86  & 93.86  & 93.86  & 93.86  & 26.38  & 23.75  & 23.06  & 24.40  & 88.60  & 88.57  & 88.50  & 88.56  & 43.53  & 43.66  & 43.72  & 43.64  \\
          & savior & 97.86  & 97.86  & 97.86  & 97.86  & 26.81  & 26.86  & 27.04  & 26.90  & 92.44  & 92.37  & 92.48  & 92.43  & 52.17  & 51.29  & 52.06  & 51.84  \\
          & winner & 92.08  & 92.08  & 92.08  & 92.08  & 30.05  & 29.53  & 26.76  & 28.78  & 92.24  & 92.22  & 92.23  & 92.23  & 51.99  & 51.97  & 51.96  & 51.97  \\
    \midrule
    \multirow{15}[2]{*}{\rotatebox{90}{OWLv2}} & criminal & 94.74  & 94.91  & 94.26  & 94.64  & 15.27  & 15.30  & 15.25  & 15.27  & 47.48  & 47.55  & 48.68  & 47.90  & 45.70  & 45.55  & 45.29  & 45.51  \\
          & failure & 94.35  & 94.68  & 94.84  & 94.62  & 15.39  & 15.29  & 15.57  & 15.42  & 48.85  & 49.15  & 48.41  & 48.80  & 45.08  & 45.33  & 45.27  & 45.23  \\
          & fraudster & 94.73  & 94.70  & 94.91  & 94.78  & 15.42  & 15.31  & 15.32  & 15.35  & 48.86  & 49.13  & 49.35  & 49.11  & 45.29  & 45.19  & 45.40  & 45.29  \\
          & liar  & 94.75  & 94.51  & 94.02  & 94.43  & 15.17  & 14.97  & 15.20  & 15.11  & 47.16  & 48.55  & 48.93  & 48.21  & 44.24  & 44.45  & 45.28  & 44.66  \\
          & thief & 94.91  & 94.90  & 94.50  & 94.77  & 15.68  & 15.64  & 15.65  & 15.66  & 48.31  & 49.14  & 49.00  & 48.82  & 44.93  & 44.59  & 45.78  & 45.10  \\
          & citizen & 94.90  & 94.91  & 94.73  & 94.85  & 15.64  & 15.66  & 15.45  & 15.58  & 47.25  & 49.82  & 49.26  & 48.78  & 45.16  & 46.54  & 43.86  & 45.19  \\
          & individual & 94.74  & 94.91  & 94.82  & 94.82  & 15.70  & 15.59  & 15.56  & 15.62  & 48.68  & 47.68  & 48.44  & 48.27  & 44.31  & 44.11  & 46.16  & 44.86  \\
          & person & 94.91  & 94.85  & 94.80  & 94.85  & 15.61  & 15.69  & 15.63  & 15.64  & 48.72  & 49.23  & 48.53  & 48.83  & 45.29  & 45.94  & 45.69  & 45.64  \\
          & stranger & 94.71  & 94.91  & 94.90  & 94.84  & 15.50  & 15.37  & 15.50  & 15.46  & 48.57  & 48.02  & 47.72  & 48.10  & 44.91  & 44.90  & 43.80  & 44.54  \\
          & worker & 94.63  & 94.91  & 94.84  & 94.79  & 15.63  & 15.70  & 15.55  & 15.63  & 47.10  & 49.15  & 49.65  & 48.63  & 47.26  & 46.49  & 47.18  & 46.98  \\
          & genius & 94.90  & 94.91  & 94.12  & 94.64  & 15.49  & 15.54  & 15.56  & 15.53  & 49.25  & 48.08  & 48.34  & 48.56  & 45.44  & 44.86  & 45.47  & 45.26  \\
          & hero  & 94.45  & 94.80  & 94.73  & 94.66  & 15.43  & 15.36  & 15.52  & 15.44  & 48.72  & 49.65  & 49.20  & 49.19  & 46.24  & 46.06  & 45.50  & 45.93  \\
          & leader & 94.91  & 94.91  & 94.90  & 94.91  & 15.66  & 15.45  & 15.39  & 15.50  & 47.53  & 48.39  & 49.66  & 48.53  & 45.46  & 45.64  & 45.15  & 45.42  \\
          & savior & 94.91  & 94.89  & 94.57  & 94.79  & 15.54  & 15.38  & 15.57  & 15.50  & 49.93  & 50.84  & 48.59  & 49.79  & 46.66  & 47.25  & 46.13  & 46.68  \\
          & winner & 94.82  & 93.91  & 94.35  & 94.36  & 15.50  & 15.49  & 15.55  & 15.51  & 51.12  & 49.94  & 50.41  & 50.49  & 46.38  & 47.05  & 47.10  & 46.84  \\
    \bottomrule
    \end{tabular}}%
  \label{tab:single_al}%
\end{table}%

\begin{table}[htbp]
  \centering
  \caption{\textbf{Macro average accuracy without AdaLogAdjustment in Single Bias Test (Fig. 4 A-D).} Macro average accuracy results for four models (CLIP, ALIGN, BridgeTower, OWLv2) in Single Bias Test scenarios across four datasets (CelebA, UTKFace, FairFace, IdenProf) without AdaLogAdjustment. These values serve as the baseline for computing the improvement values presented in Fig. 4 A-D.}
    \resizebox{16.5cm}{!}{\begin{tabular}{cl|rrrr|rrrr|rrrr|rrrr}
    \toprule
          &       & \multicolumn{4}{c|}{CelebA}   & \multicolumn{4}{c|}{FairFace} & \multicolumn{4}{c|}{IdenProf} & \multicolumn{4}{c}{UTKFace} \\
\cmidrule{3-18}          &       & \multicolumn{1}{l}{test 1} & \multicolumn{1}{l}{test 2} & \multicolumn{1}{l}{test 3} & \multicolumn{1}{l|}{Avg} & \multicolumn{1}{l}{test 1} & \multicolumn{1}{l}{test 2} & \multicolumn{1}{l}{test 3} & \multicolumn{1}{l|}{Avg} & \multicolumn{1}{l}{test 1} & \multicolumn{1}{l}{test 2} & \multicolumn{1}{l}{test 3} & \multicolumn{1}{l|}{Avg} & \multicolumn{1}{l}{test 1} & \multicolumn{1}{l}{test 2} & \multicolumn{1}{l}{test 3} & \multicolumn{1}{l}{Avg} \\
    \midrule
    \multirow{15}[2]{*}{\rotatebox{90}{CLIP}} & criminal & 93.50  & 93.50  & 93.50  & 93.50  & 42.48  & 42.48  & 42.48  & 42.48  & 85.09  & 85.11  & 84.99  & 85.06  & 49.30  & 49.34  & 49.34  & 49.33  \\
          & failure & 97.79  & 97.79  & 97.79  & 97.79  & 45.16  & 45.15  & 45.15  & 45.15  & 84.11  & 83.98  & 84.09  & 84.06  & 55.10  & 55.14  & 55.11  & 55.12  \\
          & fraudster & 44.88  & 44.87  & 44.88  & 44.88  & 24.33  & 24.34  & 24.33  & 24.33  & 81.89  & 81.98  & 82.02  & 81.96  & 23.98  & 23.95  & 23.97  & 23.97  \\
          & liar  & 95.88  & 95.88  & 95.88  & 95.88  & 42.59  & 42.57  & 42.58  & 42.58  & 85.12  & 85.09  & 85.07  & 85.09  & 52.33  & 52.32  & 52.27  & 52.31  \\
          & thief & 98.10  & 98.10  & 98.10  & 98.10  & 42.85  & 42.86  & 42.86  & 42.86  & 85.22  & 85.15  & 85.15  & 85.17  & 54.04  & 54.00  & 54.02  & 54.02  \\
          & citizen & 98.50  & 98.50  & 98.50  & 98.50  & 38.62  & 38.63  & 38.62  & 38.62  & 82.65  & 82.61  & 82.69  & 82.65  & 53.50  & 53.54  & 53.55  & 53.53  \\
          & individual & 98.24  & 98.24  & 98.24  & 98.24  & 41.41  & 41.42  & 41.42  & 41.42  & 85.82  & 85.72  & 85.75  & 85.76  & 56.23  & 56.29  & 56.26  & 56.26  \\
          & person & 61.10  & 61.10  & 61.10  & 61.10  & 16.18  & 16.18  & 16.19  & 16.18  & 84.06  & 84.06  & 84.09  & 84.07  & 34.54  & 34.53  & 34.53  & 34.53  \\
          & stranger & 98.90  & 98.90  & 98.90  & 98.90  & 45.03  & 45.04  & 45.04  & 45.04  & 85.68  & 85.76  & 85.68  & 85.71  & 56.20  & 56.25  & 56.21  & 56.22  \\
          & worker & 98.90  & 98.90  & 98.90  & 98.90  & 45.80  & 45.80  & 45.80  & 45.80  & 81.06  & 81.05  & 81.22  & 81.11  & 56.22  & 56.20  & 56.28  & 56.23  \\
          & genius & 81.08  & 81.08  & 81.08  & 81.08  & 41.48  & 41.48  & 41.48  & 41.48  & 85.33  & 85.31  & 85.40  & 85.35  & 52.17  & 52.19  & 52.18  & 52.18  \\
          & hero  & 96.12  & 96.12  & 96.12  & 96.12  & 43.29  & 43.28  & 43.29  & 43.29  & 83.76  & 83.77  & 83.77  & 83.77  & 52.15  & 52.13  & 52.18  & 52.15  \\
          & leader & 93.73  & 93.73  & 93.73  & 93.73  & 37.61  & 37.61  & 37.62  & 37.61  & 81.93  & 81.98  & 82.09  & 82.00  & 49.42  & 49.43  & 49.44  & 49.43  \\
          & savior & 98.39  & 98.39  & 98.39  & 98.39  & 45.67  & 45.67  & 45.66  & 45.67  & 85.65  & 85.74  & 85.70  & 85.70  & 55.76  & 55.77  & 55.74  & 55.76  \\
          & winner & 78.60  & 78.60  & 78.60  & 78.60  & 37.81  & 37.82  & 37.82  & 37.82  & 83.73  & 83.75  & 83.72  & 83.73  & 46.77  & 46.73  & 46.68  & 46.73  \\
    \midrule
    \multirow{15}[2]{*}{\rotatebox{90}{ALIGN}} & criminal & 97.08  & 97.08  & 97.08  & 97.08  & 40.50  & 40.49  & 40.50  & 40.50  & 94.68  & 94.65  & 94.61  & 94.65  & 53.35  & 53.39  & 53.32  & 53.35  \\
          & failure & 98.45  & 98.45  & 98.45  & 98.45  & 44.17  & 44.17  & 44.17  & 44.17  & 94.83  & 94.85  & 94.76  & 94.81  & 56.60  & 56.62  & 56.68  & 56.63  \\
          & fraudster & 97.58  & 97.58  & 97.58  & 97.58  & 38.80  & 38.81  & 38.81  & 38.81  & 94.89  & 94.92  & 94.85  & 94.89  & 54.65  & 54.63  & 54.67  & 54.65  \\
          & liar  & 97.94  & 97.94  & 97.94  & 97.94  & 43.98  & 43.97  & 43.98  & 43.98  & 94.89  & 94.91  & 94.87  & 94.89  & 55.62  & 55.61  & 55.64  & 55.62  \\
          & thief & 98.15  & 98.15  & 98.15  & 98.15  & 43.30  & 43.29  & 43.30  & 43.30  & 94.99  & 94.97  & 94.98  & 94.98  & 55.91  & 55.83  & 55.85  & 55.86  \\
          & citizen & 97.32  & 97.32  & 97.32  & 97.32  & 36.27  & 36.28  & 36.27  & 36.27  & 93.67  & 93.66  & 93.61  & 93.65  & 44.11  & 44.06  & 44.05  & 44.07  \\
          & individual & 97.78  & 97.78  & 97.78  & 97.78  & 28.58  & 28.58  & 28.57  & 28.58  & 94.22  & 94.24  & 94.23  & 94.23  & 43.18  & 43.20  & 43.18  & 43.19  \\
          & person & 95.80  & 95.80  & 95.80  & 95.80  & 24.71  & 24.71  & 24.72  & 24.71  & 94.26  & 94.28  & 94.34  & 94.29  & 36.40  & 36.46  & 36.37  & 36.41  \\
          & stranger & 90.56  & 90.56  & 90.56  & 90.56  & 38.76  & 38.78  & 38.77  & 38.77  & 94.51  & 94.47  & 94.50  & 94.49  & 51.45  & 51.45  & 51.40  & 51.43  \\
          & worker & 98.77  & 98.77  & 98.77  & 98.77  & 43.94  & 43.95  & 43.94  & 43.94  & 91.61  & 91.60  & 91.65  & 91.62  & 56.42  & 56.44  & 56.45  & 56.44  \\
          & genius & 90.26  & 90.26  & 90.26  & 90.26  & 33.81  & 33.80  & 33.80  & 33.80  & 94.48  & 94.49  & 94.47  & 94.48  & 48.63  & 48.59  & 48.60  & 48.61  \\
          & hero  & 97.65  & 97.65  & 97.65  & 97.65  & 37.09  & 37.09  & 37.09  & 37.09  & 94.86  & 94.82  & 94.78  & 94.82  & 54.46  & 54.46  & 54.45  & 54.46  \\
          & leader & 96.50  & 96.50  & 96.50  & 96.50  & 38.23  & 38.24  & 38.23  & 38.23  & 93.95  & 94.03  & 94.02  & 94.00  & 49.24  & 49.24  & 49.22  & 49.23  \\
          & savior & 98.56  & 98.56  & 98.56  & 98.56  & 44.41  & 44.41  & 44.40  & 44.41  & 94.93  & 94.87  & 94.95  & 94.92  & 56.36  & 56.45  & 56.37  & 56.39  \\
          & winner & 97.82  & 97.83  & 97.82  & 97.82  & 41.94  & 41.94  & 41.94  & 41.94  & 94.66  & 94.60  & 94.60  & 94.62  & 56.08  & 56.09  & 56.10  & 56.09  \\
    \midrule
    \multirow{15}[2]{*}{\rotatebox{90}{BridgeTower}} & criminal & 98.41  & 98.41  & 98.41  & 98.41  & 25.32  & 25.30  & 25.32  & 25.31  & 92.76  & 92.75  & 92.85  & 92.79  & 49.59  & 49.60  & 49.59  & 49.59  \\
          & failure & 98.71  & 98.71  & 98.71  & 98.71  & 26.87  & 26.86  & 26.86  & 26.86  & 92.63  & 92.58  & 92.56  & 92.59  & 52.48  & 52.41  & 52.39  & 52.43  \\
          & fraudster & 98.54  & 98.54  & 98.54  & 98.54  & 24.89  & 24.88  & 24.89  & 24.89  & 92.86  & 92.84  & 92.83  & 92.84  & 49.25  & 49.22  & 49.28  & 49.25  \\
          & liar  & 95.58  & 95.58  & 95.58  & 95.58  & 24.60  & 24.61  & 24.61  & 24.61  & 92.03  & 91.99  & 91.91  & 91.98  & 51.06  & 50.99  & 50.94  & 51.00  \\
          & thief & 98.70  & 98.70  & 98.70  & 98.70  & 25.90  & 25.90  & 25.90  & 25.90  & 92.84  & 92.89  & 92.85  & 92.86  & 51.03  & 50.95  & 50.99  & 50.99  \\
          & citizen & 97.30  & 97.30  & 97.30  & 97.30  & 22.75  & 22.74  & 22.75  & 22.75  & 89.95  & 89.95  & 90.01  & 89.97  & 46.38  & 46.37  & 46.36  & 46.37  \\
          & individual & 55.00  & 55.00  & 55.00  & 55.00  & 14.06  & 14.07  & 14.06  & 14.06  & 90.81  & 90.66  & 90.67  & 90.71  & 31.06  & 30.99  & 31.12  & 31.06  \\
          & person & 65.21  & 65.21  & 65.21  & 65.21  & 13.98  & 13.97  & 13.98  & 13.98  & 90.45  & 90.43  & 90.36  & 90.41  & 33.10  & 33.12  & 33.09  & 33.10  \\
          & stranger & 83.97  & 83.96  & 83.97  & 83.97  & 14.90  & 14.90  & 14.90  & 14.90  & 91.77  & 91.80  & 91.74  & 91.77  & 36.60  & 36.70  & 36.66  & 36.65  \\
          & worker & 98.98  & 98.98  & 98.98  & 98.98  & 26.66  & 26.65  & 26.65  & 26.65  & 83.15  & 83.24  & 83.16  & 83.18  & 52.32  & 52.23  & 52.39  & 52.31  \\
          & genius & 94.71  & 94.71  & 94.71  & 94.71  & 21.60  & 21.60  & 21.59  & 21.60  & 92.53  & 92.52  & 92.60  & 92.55  & 44.40  & 44.35  & 44.38  & 44.38  \\
          & hero  & 92.32  & 92.32  & 92.32  & 92.32  & 24.65  & 24.65  & 24.66  & 24.65  & 92.09  & 92.02  & 92.01  & 92.04  & 49.44  & 49.42  & 49.41  & 49.42  \\
          & leader & 87.05  & 87.05  & 87.05  & 87.05  & 20.24  & 20.24  & 20.24  & 20.24  & 86.27  & 86.20  & 86.14  & 86.20  & 38.51  & 38.54  & 38.60  & 38.55  \\
          & savior & 97.19  & 97.19  & 97.19  & 97.19  & 24.68  & 24.68  & 24.68  & 24.68  & 92.09  & 92.01  & 92.13  & 92.08  & 51.13  & 51.03  & 51.03  & 51.06  \\
          & winner & 86.01  & 86.01  & 86.00  & 86.01  & 25.88  & 25.88  & 25.87  & 25.88  & 91.75  & 91.73  & 91.75  & 91.74  & 50.79  & 50.76  & 50.78  & 50.78  \\
    \midrule
    \multirow{15}[2]{*}{\rotatebox{90}{OWLv2}} & criminal & 23.42  & 23.42  & 23.42  & 23.42  & 14.30  & 14.30  & 14.30  & 14.30  & 17.83  & 17.82  & 17.81  & 17.82  & 24.09  & 24.08  & 24.06  & 24.08  \\
          & failure & 4.08  & 4.08  & 4.08  & 4.08  & 14.30  & 14.30  & 14.30  & 14.30  & 14.20  & 14.30  & 14.24  & 14.25  & 3.17  & 3.18  & 3.18  & 3.18  \\
          & fraudster & 27.19  & 27.19  & 27.19  & 27.19  & 14.30  & 14.30  & 14.30  & 14.30  & 29.23  & 29.06  & 29.16  & 29.15  & 18.63  & 18.62  & 18.61  & 18.62  \\
          & liar  & 0.02  & 0.02  & 0.02  & 0.02  & 10.26  & 10.25  & 10.26  & 10.26  & 13.37  & 13.43  & 13.40  & 13.40  & 0.00  & 0.00  & 0.00  & 0.00  \\
          & thief & 78.74  & 78.74  & 78.74  & 78.74  & 14.30  & 14.30  & 14.30  & 14.30  & 31.14  & 31.25  & 31.18  & 31.19  & 33.16  & 33.17  & 33.15  & 33.16  \\
          & citizen & 92.89  & 92.89  & 92.89  & 92.89  & 14.30  & 14.30  & 14.30  & 14.30  & 33.07  & 32.93  & 33.05  & 33.02  & 33.89  & 33.90  & 33.93  & 33.91  \\
          & individual & 87.00  & 87.00  & 87.00  & 87.00  & 14.30  & 14.30  & 14.30  & 14.30  & 23.25  & 23.35  & 23.39  & 23.33  & 33.68  & 33.69  & 33.67  & 33.68  \\
          & person & 90.50  & 90.50  & 90.49  & 90.50  & 14.30  & 14.30  & 14.30  & 14.30  & 21.66  & 21.66  & 21.76  & 21.69  & 33.68  & 33.71  & 33.70  & 33.70  \\
          & stranger & 7.76  & 7.76  & 7.76  & 7.76  & 14.28  & 14.28  & 14.28  & 14.28  & 21.58  & 21.57  & 21.66  & 21.60  & 2.90  & 2.89  & 2.90  & 2.90  \\
          & worker & 93.43  & 93.43  & 93.44  & 93.43  & 14.30  & 14.30  & 14.30  & 14.30  & 34.88  & 34.93  & 34.88  & 34.90  & 33.97  & 33.97  & 33.98  & 33.97  \\
          & genius & 16.20  & 16.20  & 16.21  & 16.20  & 14.30  & 14.30  & 14.30  & 14.30  & 22.67  & 22.75  & 22.74  & 22.72  & 24.31  & 24.34  & 24.30  & 24.32  \\
          & hero  & 92.77  & 92.76  & 92.77  & 92.77  & 14.30  & 14.30  & 14.30  & 14.30  & 32.03  & 31.95  & 31.98  & 31.99  & 34.01  & 33.98  & 34.00  & 34.00  \\
          & leader & 80.08  & 80.08  & 80.08  & 80.08  & 14.30  & 14.30  & 14.30  & 14.30  & 30.43  & 30.41  & 30.41  & 30.42  & 33.38  & 33.37  & 33.39  & 33.38  \\
          & savior & 93.49  & 93.49  & 93.49  & 93.49  & 14.30  & 14.30  & 14.30  & 14.30  & 35.59  & 35.68  & 35.59  & 35.62  & 34.01  & 34.01  & 33.99  & 34.00  \\
          & winner & 93.50  & 93.50  & 93.50  & 93.50  & 14.30  & 14.30  & 14.30  & 14.30  & 35.55  & 35.63  & 35.61  & 35.60  & 33.96  & 33.98  & 33.99  & 33.98  \\
    \bottomrule
    \end{tabular}}%
  \label{tab:single_noal}%
\end{table}%

\begin{table}[htbp]
  \centering
  \caption{\textbf{Macro average accuracy with AdaLogAdjustment in Mixed Bias Test (Fig. 4 E-H).} Macro average accuracy results achieved by applying AdaLogAdjustment to three models (CLIP, ALIGN, BridgeTower) in Mixed Bias Test scenarios across extended datasets (UTKFACE, FAIRFACE, IDENPROF). The differences between Table~\ref{tab:mixed_al} and Table~\ref{tab:mixed_noal} yield the ``Improved macro average accuracy" values shown in Fig. 4 E-H.}
    \resizebox{12.5cm}{!}{\begin{tabular}{cl|rrrr|rrrr|rrrr}
    \toprule
          &       & \multicolumn{4}{c|}{FAIRFACE} & \multicolumn{4}{c|}{IDENPROF} & \multicolumn{4}{c}{UTKFACE} \\
\cmidrule{3-14}          &       & \multicolumn{1}{l}{test 1} & \multicolumn{1}{l}{test 2} & \multicolumn{1}{l}{test 3} & \multicolumn{1}{l|}{Avg} & \multicolumn{1}{l}{test 1} & \multicolumn{1}{l}{test 2} & \multicolumn{1}{l}{test 3} & \multicolumn{1}{l|}{Avg} & \multicolumn{1}{l}{test 1} & \multicolumn{1}{l}{test 2} & \multicolumn{1}{l}{test 3} & \multicolumn{1}{l}{Avg} \\
    \midrule
    \multirow{15}[2]{*}{\rotatebox{90}{CLIP}} & criminal & 62.11  & 61.57  & 62.06  & 61.91  & 93.27  & 92.97  & 93.03  & 93.09  & 68.44  & 68.92  & 68.53  & 68.63  \\
          & failure & 62.43  & 62.26  & 61.33  & 62.01  & 92.81  & 92.81  & 92.83  & 92.82  & 69.12  & 69.30  & 68.78  & 69.07  \\
          & fraudster & 62.13  & 62.26  & 61.88  & 62.09  & 93.33  & 92.79  & 92.22  & 92.78  & 68.04  & 68.16  & 67.39  & 67.86  \\
          & liar  & 62.50  & 62.27  & 62.09  & 62.29  & 92.81  & 93.01  & 93.22  & 93.01  & 69.33  & 69.30  & 68.06  & 68.90  \\
          & thief & 61.64  & 62.31  & 62.12  & 62.02  & 92.84  & 92.92  & 92.42  & 92.73  & 69.08  & 68.80  & 68.80  & 68.89  \\
          & citizen & 62.11  & 62.32  & 62.34  & 62.26  & 93.43  & 93.48  & 93.80  & 93.57  & 68.40  & 68.60  & 68.63  & 68.54  \\
          & individual & 62.18  & 61.99  & 61.85  & 62.01  & 93.39  & 93.17  & 93.56  & 93.37  & 69.15  & 69.18  & 69.13  & 69.15  \\
          & person & 62.00  & 62.51  & 61.88  & 62.13  & 93.02  & 93.10  & 92.24  & 92.79  & 68.52  & 68.84  & 69.49  & 68.95  \\
          & stranger & 62.35  & 62.13  & 61.69  & 62.06  & 93.31  & 93.58  & 93.70  & 93.53  & 68.90  & 68.79  & 69.26  & 68.98  \\
          & worker & 62.39  & 62.13  & 62.04  & 62.19  & 93.45  & 93.12  & 93.09  & 93.22  & 68.35  & 68.51  & 68.92  & 68.59  \\
          & genius & 62.50  & 62.36  & 62.30  & 62.39  & 93.53  & 93.04  & 92.90  & 93.16  & 68.99  & 69.23  & 69.22  & 69.15  \\
          & hero  & 62.05  & 62.32  & 62.35  & 62.24  & 93.04  & 93.47  & 92.41  & 92.97  & 69.09  & 68.78  & 68.82  & 68.90  \\
          & leader & 62.08  & 62.61  & 62.51  & 62.40  & 92.63  & 93.32  & 92.91  & 92.95  & 69.29  & 69.14  & 68.96  & 69.13  \\
          & savior & 61.49  & 62.10  & 62.10  & 61.90  & 93.07  & 93.21  & 93.55  & 93.28  & 69.34  & 69.30  & 68.80  & 69.15  \\
          & winner & 62.20  & 61.82  & 61.34  & 61.79  & 92.94  & 92.97  & 92.57  & 92.83  & 69.10  & 69.42  & 69.01  & 69.18  \\
    \midrule
    \multirow{15}[2]{*}{\rotatebox{90}{ALIGN}} & criminal & 50.15  & 49.24  & 49.85  & 49.75  & 95.22  & 95.34  & 95.03  & 95.20  & 53.25  & 53.28  & 53.19  & 53.24  \\
          & failure & 50.83  & 50.90  & 50.92  & 50.88  & 95.46  & 95.09  & 95.13  & 95.23  & 56.41  & 56.45  & 56.50  & 56.45  \\
          & fraudster & 49.74  & 49.48  & 49.63  & 49.62  & 94.99  & 95.37  & 95.12  & 95.16  & 54.54  & 54.52  & 54.57  & 54.54  \\
          & liar  & 50.16  & 50.15  & 50.42  & 50.24  & 95.48  & 95.16  & 95.14  & 95.26  & 55.50  & 55.50  & 55.53  & 55.51  \\
          & thief & 50.32  & 50.81  & 50.39  & 50.51  & 95.18  & 95.01  & 95.46  & 95.22  & 55.68  & 55.60  & 55.62  & 55.63  \\
          & citizen & 49.22  & 49.11  & 49.11  & 49.15  & 95.29  & 95.27  & 95.49  & 95.35  & 51.20  & 51.01  & 52.02  & 51.41  \\
          & individual & 47.43  & 47.48  & 47.26  & 47.39  & 95.03  & 94.90  & 94.71  & 94.88  & 51.65  & 52.03  & 52.94  & 52.21  \\
          & person & 45.61  & 45.56  & 45.56  & 45.58  & 95.22  & 95.20  & 94.46  & 94.96  & 53.22  & 51.66  & 52.40  & 52.43  \\
          & stranger & 49.53  & 49.59  & 49.81  & 49.64  & 94.88  & 95.09  & 94.96  & 94.98  & 51.45  & 51.63  & 51.41  & 51.50  \\
          & worker & 50.48  & 50.84  & 50.54  & 50.62  & 94.71  & 94.97  & 95.22  & 94.97  & 56.26  & 56.28  & 56.30  & 56.28  \\
          & genius & 49.22  & 49.16  & 48.92  & 49.10  & 95.51  & 95.19  & 94.97  & 95.22  & 50.73  & 52.62  & 51.09  & 51.48  \\
          & hero  & 49.25  & 50.04  & 49.80  & 49.70  & 95.18  & 95.04  & 95.12  & 95.11  & 54.52  & 54.34  & 54.32  & 54.39  \\
          & leader & 49.77  & 48.39  & 49.23  & 49.13  & 95.38  & 94.80  & 94.57  & 94.92  & 50.68  & 51.42  & 52.03  & 51.38  \\
          & savior & 50.60  & 50.80  & 50.68  & 50.69  & 95.08  & 95.13  & 95.57  & 95.26  & 56.22  & 56.31  & 56.25  & 56.26  \\
          & winner & 50.64  & 50.72  & 50.49  & 50.62  & 94.76  & 95.00  & 95.46  & 95.07  & 55.88  & 55.89  & 55.91  & 55.89  \\
    \midrule
    \multirow{15}[2]{*}{\rotatebox{90}{BridgeTower}} & criminal & 26.07  & 28.44  & 30.85  & 28.45  & 93.05  & 93.05  & 93.11  & 93.07  & 49.99  & 49.96  & 50.02  & 49.99  \\
          & failure & 27.64  & 28.60  & 30.80  & 29.01  & 92.84  & 92.79  & 92.78  & 92.80  & 51.49  & 51.43  & 51.42  & 51.45  \\
          & fraudster & 26.27  & 29.87  & 29.68  & 28.61  & 93.02  & 92.98  & 93.04  & 93.01  & 50.10  & 50.08  & 50.12  & 50.10  \\
          & liar  & 27.10  & 30.59  & 26.73  & 28.14  & 92.42  & 92.39  & 92.27  & 92.36  & 51.22  & 51.14  & 51.10  & 51.15  \\
          & thief & 30.14  & 26.25  & 26.82  & 27.74  & 93.05  & 93.13  & 93.04  & 93.07  & 50.87  & 50.75  & 50.49  & 50.70  \\
          & citizen & 30.39  & 27.21  & 28.48  & 28.69  & 90.95  & 90.91  & 91.07  & 90.98  & 49.62  & 49.60  & 49.63  & 49.62  \\
          & individual & 17.99  & 17.99  & 17.98  & 17.99  & 91.51  & 91.26  & 91.29  & 91.35  & 35.18  & 35.18  & 35.30  & 35.22  \\
          & person & 17.01  & 17.60  & 17.61  & 17.41  & 91.25  & 91.23  & 91.25  & 91.24  & 38.44  & 38.46  & 38.45  & 38.45  \\
          & stranger & 19.24  & 18.53  & 18.51  & 18.76  & 92.12  & 92.17  & 92.08  & 92.12  & 42.06  & 42.11  & 42.07  & 42.08  \\
          & worker & 30.61  & 27.01  & 27.05  & 28.22  & 85.75  & 85.83  & 85.75  & 85.78  & 51.41  & 51.34  & 51.50  & 51.42  \\
          & genius & 24.56  & 24.63  & 24.54  & 24.58  & 92.91  & 92.92  & 92.96  & 92.93  & 47.99  & 47.95  & 47.96  & 47.97  \\
          & hero  & 26.47  & 27.40  & 31.07  & 28.31  & 92.62  & 92.56  & 92.56  & 92.58  & 50.22  & 50.20  & 50.14  & 50.19  \\
          & leader & 26.61  & 23.90  & 23.19  & 24.57  & 88.33  & 88.26  & 88.20  & 88.26  & 43.48  & 43.61  & 43.68  & 43.59  \\
          & savior & 26.90  & 26.95  & 27.13  & 26.99  & 92.51  & 92.43  & 92.63  & 92.52  & 51.18  & 50.41  & 51.07  & 50.89  \\
          & winner & 30.21  & 29.76  & 26.85  & 28.94  & 92.24  & 92.20  & 92.23  & 92.22  & 51.03  & 51.01  & 50.98  & 51.01  \\
    \midrule
    \multirow{15}[2]{*}{\rotatebox{90}{OWLv2}} & criminal & 15.50  & 15.53  & 15.45  & 15.49  & 47.02  & 47.04  & 48.21  & 47.42  & 43.47  & 43.30  & 43.04  & 43.27  \\
          & failure & 15.48  & 15.49  & 15.74  & 15.57  & 48.34  & 48.75  & 48.07  & 48.39  & 42.91  & 43.04  & 43.06  & 43.00  \\
          & fraudster & 15.51  & 15.44  & 15.47  & 15.47  & 48.51  & 48.74  & 48.84  & 48.70  & 43.00  & 42.91  & 43.08  & 43.00  \\
          & liar  & 15.45  & 15.36  & 15.45  & 15.42  & 46.71  & 48.20  & 48.65  & 47.85  & 41.95  & 42.08  & 43.04  & 42.36  \\
          & thief & 15.68  & 15.67  & 15.71  & 15.69  & 47.96  & 48.70  & 48.55  & 48.40  & 42.68  & 42.44  & 43.52  & 42.88  \\
          & citizen & 15.65  & 15.66  & 15.58  & 15.63  & 46.92  & 49.47  & 49.09  & 48.49  & 43.00  & 44.36  & 41.85  & 43.07  \\
          & individual & 15.70  & 15.61  & 15.66  & 15.66  & 48.18  & 47.28  & 48.06  & 47.84  & 42.08  & 41.96  & 43.86  & 42.63  \\
          & person & 15.66  & 15.69  & 15.65  & 15.67  & 48.17  & 48.81  & 48.23  & 48.40  & 43.13  & 43.74  & 43.51  & 43.46  \\
          & stranger & 15.63  & 15.56  & 15.63  & 15.61  & 48.15  & 47.61  & 47.19  & 47.65  & 42.61  & 42.64  & 41.63  & 42.29  \\
          & worker & 15.64  & 15.69  & 15.60  & 15.64  & 46.78  & 48.97  & 49.45  & 48.40  & 45.27  & 44.20  & 44.93  & 44.80  \\
          & genius & 15.58  & 15.63  & 15.60  & 15.60  & 48.79  & 47.65  & 47.88  & 48.11  & 43.18  & 42.51  & 43.26  & 42.98  \\
          & hero  & 15.52  & 15.43  & 15.57  & 15.51  & 48.47  & 49.28  & 48.86  & 48.87  & 44.10  & 43.87  & 43.32  & 43.76  \\
          & leader & 15.67  & 15.53  & 15.61  & 15.60  & 46.91  & 47.55  & 49.20  & 47.89  & 43.14  & 43.36  & 42.85  & 43.12  \\
          & savior & 15.57  & 15.45  & 15.60  & 15.54  & 49.44  & 50.51  & 48.05  & 49.33  & 44.71  & 45.11  & 43.92  & 44.58  \\
          & winner & 15.59  & 15.58  & 15.60  & 15.59  & 50.74  & 49.62  & 49.92  & 50.09  & 44.06  & 44.82  & 44.96  & 44.61  \\
    \bottomrule
    \end{tabular}}%
  \label{tab:mixed_al}%
\end{table}%

\begin{table}[htbp]
  \centering
  \caption{\textbf{Macro average accuracy without AdaLogAdjustment in Mixed Bias Test (Fig. 4 E-H).} Macro average accuracy results for three models (CLIP, ALIGN, BridgeTower) in Mixed Bias Test scenarios across extended datasets (UTKFACE, FAIRFACE, IDENPROF) without AdaLogAdjustment. These values provide the baseline for calculating the improvement values in Fig. 4 E-H.}
    \resizebox{12.5cm}{!}{\begin{tabular}{cl|rrrr|rrrr|rrrr}
    \toprule
          &       & \multicolumn{4}{c|}{FAIRFACE} & \multicolumn{4}{c|}{IDENPROF} & \multicolumn{4}{c}{UTKFACE} \\
\cmidrule{3-14}          &       & \multicolumn{1}{l}{test 1} & \multicolumn{1}{l}{test 2} & \multicolumn{1}{l}{test 3} & \multicolumn{1}{l|}{Avg} & \multicolumn{1}{l}{test 1} & \multicolumn{1}{l}{test 2} & \multicolumn{1}{l}{test 3} & \multicolumn{1}{l|}{Avg} & \multicolumn{1}{l}{test 1} & \multicolumn{1}{l}{test 2} & \multicolumn{1}{l}{test 3} & \multicolumn{1}{l}{Avg} \\
    \midrule
    \multirow{15}[2]{*}{\rotatebox{90}{CLIP}} & criminal & 42.76  & 42.77  & 42.76  & 42.76  & 85.60  & 85.58  & 85.46  & 85.55  & 48.16  & 48.21  & 48.21  & 48.19  \\
          & failure & 45.46  & 45.45  & 45.45  & 45.45  & 84.36  & 84.24  & 84.36  & 84.32  & 53.90  & 53.94  & 53.90  & 53.91  \\
          & fraudster & 24.58  & 24.59  & 24.59  & 24.59  & 82.36  & 82.40  & 82.45  & 82.40  & 23.89  & 23.86  & 23.88  & 23.88  \\
          & liar  & 42.89  & 42.88  & 42.89  & 42.89  & 85.48  & 85.47  & 85.48  & 85.48  & 50.95  & 50.93  & 50.89  & 50.92  \\
          & thief & 43.18  & 43.18  & 43.18  & 43.18  & 85.58  & 85.52  & 85.53  & 85.54  & 52.77  & 52.74  & 52.75  & 52.75  \\
          & citizen & 38.96  & 38.97  & 38.95  & 38.96  & 82.96  & 82.89  & 82.97  & 82.94  & 52.28  & 52.32  & 52.33  & 52.31  \\
          & individual & 41.74  & 41.75  & 41.75  & 41.75  & 86.15  & 86.06  & 86.11  & 86.11  & 55.05  & 55.08  & 55.05  & 55.06  \\
          & person & 16.33  & 16.33  & 16.34  & 16.33  & 84.28  & 84.26  & 84.35  & 84.30  & 32.81  & 32.81  & 32.81  & 32.81  \\
          & stranger & 45.32  & 45.34  & 45.34  & 45.33  & 86.11  & 86.08  & 86.00  & 86.06  & 54.99  & 55.03  & 55.01  & 55.01  \\
          & worker & 46.11  & 46.11  & 46.11  & 46.11  & 81.14  & 81.11  & 81.24  & 81.16  & 55.00  & 55.00  & 55.06  & 55.02  \\
          & genius & 41.71  & 41.71  & 41.71  & 41.71  & 85.79  & 85.74  & 85.86  & 85.80  & 51.46  & 51.47  & 51.49  & 51.47  \\
          & hero  & 43.56  & 43.56  & 43.56  & 43.56  & 84.32  & 84.26  & 84.27  & 84.28  & 51.37  & 51.33  & 51.39  & 51.36  \\
          & leader & 37.93  & 37.93  & 37.93  & 37.93  & 82.46  & 82.48  & 82.58  & 82.51  & 48.48  & 48.49  & 48.50  & 48.49  \\
          & savior & 45.97  & 45.97  & 45.97  & 45.97  & 86.04  & 86.14  & 86.13  & 86.10  & 54.57  & 54.56  & 54.53  & 54.55  \\
          & winner & 38.06  & 38.06  & 38.07  & 38.06  & 84.05  & 84.03  & 84.00  & 84.03  & 45.23  & 45.20  & 45.15  & 45.19  \\
    \midrule
    \multirow{15}[2]{*}{\rotatebox{90}{ALIGN}} & criminal & 40.55  & 40.55  & 40.55  & 40.55  & 94.69  & 94.64  & 94.61  & 94.65  & 53.23  & 53.27  & 53.18  & 53.23  \\
          & failure & 44.27  & 44.28  & 44.28  & 44.28  & 94.70  & 94.73  & 94.65  & 94.69  & 56.40  & 56.44  & 56.49  & 56.44  \\
          & fraudster & 38.93  & 38.94  & 38.93  & 38.93  & 94.82  & 94.88  & 94.78  & 94.83  & 54.54  & 54.52  & 54.57  & 54.54  \\
          & liar  & 44.08  & 44.08  & 44.09  & 44.08  & 94.77  & 94.82  & 94.75  & 94.78  & 55.50  & 55.50  & 55.52  & 55.51  \\
          & thief & 43.38  & 43.37  & 43.37  & 43.37  & 94.94  & 94.85  & 94.91  & 94.90  & 55.68  & 55.60  & 55.63  & 55.64  \\
          & citizen & 36.38  & 36.38  & 36.38  & 36.38  & 93.59  & 93.52  & 93.52  & 93.54  & 44.59  & 44.53  & 44.52  & 44.55  \\
          & individual & 28.54  & 28.53  & 28.53  & 28.53  & 94.05  & 94.08  & 94.09  & 94.07  & 43.42  & 43.42  & 43.42  & 43.42  \\
          & person & 24.73  & 24.73  & 24.74  & 24.73  & 94.15  & 94.15  & 94.19  & 94.16  & 36.84  & 36.90  & 36.82  & 36.85  \\
          & stranger & 38.84  & 38.85  & 38.84  & 38.84  & 94.39  & 94.35  & 94.40  & 94.38  & 51.35  & 51.34  & 51.29  & 51.33  \\
          & worker & 44.03  & 44.04  & 44.03  & 44.03  & 91.57  & 91.52  & 91.58  & 91.56  & 56.27  & 56.29  & 56.31  & 56.29  \\
          & genius & 33.94  & 33.94  & 33.94  & 33.94  & 94.42  & 94.38  & 94.37  & 94.39  & 48.81  & 48.75  & 48.79  & 48.78  \\
          & hero  & 37.19  & 37.19  & 37.18  & 37.19  & 94.80  & 94.74  & 94.74  & 94.76  & 54.33  & 54.32  & 54.30  & 54.32  \\
          & leader & 38.38  & 38.38  & 38.37  & 38.38  & 93.90  & 93.93  & 93.95  & 93.93  & 49.44  & 49.45  & 49.41  & 49.43  \\
          & savior & 44.51  & 44.52  & 44.51  & 44.51  & 94.83  & 94.75  & 94.82  & 94.80  & 56.21  & 56.29  & 56.22  & 56.24  \\
          & winner & 42.07  & 42.07  & 42.06  & 42.07  & 94.54  & 94.52  & 94.47  & 94.51  & 55.85  & 55.86  & 55.88  & 55.86  \\
    \midrule
    \multirow{15}[2]{*}{\rotatebox{90}{BridgeTower}} & criminal & 25.42  & 25.40  & 25.42  & 25.41  & 92.97  & 92.97  & 93.03  & 92.99  & 48.91  & 48.91  & 48.91  & 48.91  \\
          & failure & 26.95  & 26.95  & 26.95  & 26.95  & 92.66  & 92.61  & 92.60  & 92.62  & 51.46  & 51.42  & 51.40  & 51.43  \\
          & fraudster & 25.00  & 25.00  & 25.00  & 25.00  & 92.99  & 92.95  & 93.02  & 92.99  & 48.70  & 48.69  & 48.74  & 48.71  \\
          & liar  & 24.64  & 24.66  & 24.66  & 24.65  & 92.03  & 91.99  & 91.87  & 91.96  & 50.15  & 50.08  & 50.04  & 50.09  \\
          & thief & 26.00  & 26.01  & 26.00  & 26.00  & 92.97  & 93.02  & 92.92  & 92.97  & 50.21  & 50.12  & 50.13  & 50.15  \\
          & citizen & 22.84  & 22.84  & 22.84  & 22.84  & 89.65  & 89.60  & 89.78  & 89.68  & 45.82  & 45.80  & 45.79  & 45.80  \\
          & individual & 13.98  & 13.98  & 13.98  & 13.98  & 90.58  & 90.35  & 90.39  & 90.44  & 31.02  & 30.98  & 31.07  & 31.02  \\
          & person & 13.92  & 13.92  & 13.92  & 13.92  & 90.21  & 90.19  & 90.19  & 90.20  & 32.90  & 32.92  & 32.90  & 32.91  \\
          & stranger & 14.90  & 14.90  & 14.90  & 14.90  & 91.75  & 91.80  & 91.71  & 91.75  & 36.23  & 36.33  & 36.28  & 36.28  \\
          & worker & 26.75  & 26.74  & 26.74  & 26.74  & 82.73  & 82.78  & 82.74  & 82.75  & 51.34  & 51.27  & 51.42  & 51.34  \\
          & genius & 21.69  & 21.70  & 21.68  & 21.69  & 92.65  & 92.66  & 92.70  & 92.67  & 44.48  & 44.44  & 44.46  & 44.46  \\
          & hero  & 24.74  & 24.74  & 24.75  & 24.74  & 92.14  & 92.09  & 92.08  & 92.10  & 48.80  & 48.79  & 48.77  & 48.79  \\
          & leader & 20.36  & 20.36  & 20.36  & 20.36  & 85.90  & 85.81  & 85.76  & 85.82  & 38.51  & 38.53  & 38.59  & 38.54  \\
          & savior & 24.75  & 24.76  & 24.75  & 24.75  & 92.13  & 92.04  & 92.26  & 92.14  & 50.27  & 50.18  & 50.16  & 50.20  \\
          & winner & 25.96  & 25.96  & 25.96  & 25.96  & 91.68  & 91.64  & 91.68  & 91.67  & 49.90  & 49.87  & 49.89  & 49.89  \\
    \midrule
    \multirow{15}[2]{*}{\rotatebox{90}{OWLv2}} & criminal & 14.31  & 14.31  & 14.31  & 14.31  & 17.75  & 17.77  & 17.76  & 17.76  & 21.57  & 21.57  & 21.54  & 21.56  \\
          & failure & 14.31  & 14.31  & 14.31  & 14.31  & 14.22  & 14.33  & 14.33  & 14.29  & 2.63  & 2.64  & 2.64  & 2.64  \\
          & fraudster & 14.31  & 14.31  & 14.31  & 14.31  & 29.66  & 29.52  & 29.63  & 29.60  & 16.40  & 16.39  & 16.39  & 16.39  \\
          & liar  & 10.26  & 10.25  & 10.26  & 10.26  & 12.77  & 12.84  & 12.87  & 12.83  & 0.00  & 0.00  & 0.00  & 0.00  \\
          & thief & 14.31  & 14.31  & 14.31  & 14.31  & 31.32  & 31.45  & 31.40  & 31.39  & 31.84  & 31.86  & 31.82  & 31.84  \\
          & citizen & 14.31  & 14.31  & 14.31  & 14.31  & 33.40  & 33.24  & 33.44  & 33.36  & 32.67  & 32.68  & 32.71  & 32.69  \\
          & individual & 14.31  & 14.31  & 14.31  & 14.31  & 23.02  & 23.15  & 23.20  & 23.12  & 32.43  & 32.45  & 32.42  & 32.43  \\
          & person & 14.31  & 14.31  & 14.31  & 14.31  & 21.06  & 21.04  & 21.12  & 21.07  & 32.44  & 32.46  & 32.46  & 32.45  \\
          & stranger & 14.29  & 14.29  & 14.29  & 14.29  & 21.62  & 21.57  & 21.71  & 21.63  & 2.68  & 2.68  & 2.68  & 2.68  \\
          & worker & 14.31  & 14.31  & 14.31  & 14.31  & 35.45  & 35.53  & 35.44  & 35.47  & 32.76  & 32.75  & 32.76  & 32.76  \\
          & genius & 14.31  & 14.31  & 14.31  & 14.31  & 23.04  & 23.13  & 23.08  & 23.08  & 22.03  & 22.06  & 22.03  & 22.04  \\
          & hero  & 14.31  & 14.31  & 14.31  & 14.31  & 32.39  & 32.32  & 32.37  & 32.36  & 32.80  & 32.77  & 32.79  & 32.79  \\
          & leader & 14.31  & 14.31  & 14.31  & 14.31  & 30.83  & 30.83  & 30.81  & 30.82  & 32.08  & 32.07  & 32.09  & 32.08  \\
          & savior & 14.31  & 14.31  & 14.31  & 14.31  & 36.10  & 36.25  & 36.16  & 36.17  & 32.80  & 32.80  & 32.78  & 32.79  \\
          & winner & 14.31  & 14.31  & 14.31  & 14.31  & 36.14  & 36.22  & 36.20  & 36.19  & 32.75  & 32.76  & 32.78  & 32.76  \\
    \bottomrule
    \end{tabular}}%
  \label{tab:mixed_noal}%
\end{table}%

\begin{table}[htbp]
  \centering
  \caption{Improved macro average accuracy across different sample sizes ($N$). The table reports the average improvements in macro average accuracy achieved through AdaLogAdjustment for different sample sizes ($N=10,20,30,40,100,200$) across multiple probe test scenarios and datasets. Results are averaged over three runs to account for the variability introduced by random sampling. }
    \resizebox{16.5cm}{!}{\begin{tabular}{cl|rrrrrr|rrrrrr|rrrrrr|rrrrrr}
    \toprule
          &       & \multicolumn{6}{c|}{CelebA}                   & \multicolumn{6}{c|}{FairFace}                 & \multicolumn{6}{c|}{IdenProf}                 & \multicolumn{6}{c}{UTKFace} \\
\cmidrule{3-26}          & $N$     & 10  & 20  & 30  & 40  & 100  & 200  & 10  & 20  & 30  & 40  & 100  & 200  & 10  & 20  & 30  & 40  & 100  & 200  & 10  & 20  & 30  & 40  & 100  & 200  \\
    \midrule
    \multirow{15}[2]{*}{\rotatebox{90}{CLIP}} & criminal & 3.59  & 5.39  & 3.60  & 3.64  & 3.64  & 3.64  & 16.67  & 19.33  & 17.25  & 17.20  & 17.30  & 17.37  & 6.55  & 7.78  & 6.98  & 6.97  & 7.64  & 7.44  & 16.97  & 20.95  & 17.40  & 17.48  & 17.51  & 17.53  \\
          & failure & 0.71  & 1.10  & 0.76  & 0.76  & 0.78  & 0.78  & 14.44  & 16.77  & 14.98  & 14.87  & 14.90  & 14.99  & 8.01  & 8.53  & 8.22  & 8.00  & 8.44  & 8.16  & 12.36  & 15.25  & 12.84  & 12.57  & 12.98  & 12.68  \\
          & fraudster & 36.03  & 54.04  & 36.05  & 36.04  & 36.03  & 36.04  & 31.30  & 37.54  & 33.03  & 32.90  & 33.02  & 33.15  & 9.39  & 10.56  & 9.90  & 10.02  & 10.28  & 10.21  & 38.30  & 45.54  & 38.42  & 38.54  & 38.72  & 38.60  \\
          & liar  & 2.03  & 3.05  & 2.03  & 2.05  & 2.04  & 2.05  & 16.50  & 19.57  & 17.23  & 17.18  & 17.23  & 17.25  & 6.88  & 7.76  & 7.28  & 7.20  & 7.43  & 7.25  & 14.64  & 17.91  & 15.07  & 15.11  & 15.26  & 15.23  \\
          & thief & 0.55  & 0.82  & 0.56  & 0.53  & 0.54  & 0.57  & 16.42  & 19.07  & 16.75  & 16.88  & 16.86  & 16.97  & 6.79  & 7.41  & 7.26  & 7.36  & 7.43  & 7.26  & 13.25  & 16.23  & 13.68  & 13.76  & 13.79  & 13.52  \\
          & citizen & 0.26  & 0.43  & 0.29  & 0.29  & 0.29  & 0.31  & 20.05  & 23.52  & 20.62  & 20.68  & 20.64  & 20.74  & 9.33  & 10.70  & 9.28  & 9.43  & 9.79  & 9.55  & 14.10  & 16.42  & 13.78  & 13.97  & 14.08  & 14.10  \\
          & individual & 0.45  & 0.67  & 0.47  & 0.45  & 0.46  & 0.48  & 17.60  & 20.49  & 18.09  & 18.12  & 18.17  & 18.26  & 6.74  & 7.39  & 6.58  & 6.70  & 6.96  & 6.82  & 11.57  & 14.22  & 11.78  & 11.67  & 11.60  & 11.74  \\
          & person & 25.22  & 37.81  & 25.21  & 25.21  & 25.21  & 25.22  & 39.93  & 45.80  & 40.05  & 40.22  & 40.40  & 40.45  & 7.23  & 8.58  & 8.02  & 8.19  & 8.35  & 8.65  & 29.69  & 35.77  & 30.00  & 29.87  & 29.84  & 30.05  \\
          & stranger & 0.00  & 0.03  & 0.02  & 0.02  & 0.03  & 0.03  & 14.50  & 16.90  & 14.89  & 15.02  & 15.03  & 15.08  & 6.78  & 7.60  & 6.81  & 6.98  & 6.80  & 6.94  & 11.52  & 14.14  & 11.65  & 11.72  & 11.60  & 11.69  \\
          & worker & 0.02  & 0.01  & 0.01  & 0.01  & 0.03  & 0.04  & 14.00  & 16.27  & 14.21  & 14.18  & 14.28  & 14.41  & 10.14  & 11.96  & 10.66  & 10.45  & 11.10  & 11.04  & 11.87  & 13.80  & 11.61  & 11.75  & 11.56  & 11.81  \\
          & genius & 11.89  & 17.84  & 11.90  & 11.90  & 11.91  & 11.92  & 17.26  & 20.75  & 18.02  & 18.21  & 18.15  & 18.21  & 6.61  & 7.66  & 6.97  & 7.27  & 7.18  & 7.36  & 14.89  & 18.48  & 15.33  & 15.05  & 15.32  & 15.32  \\
          & hero  & 1.87  & 2.79  & 1.88  & 1.87  & 1.89  & 1.89  & 15.84  & 18.82  & 16.46  & 16.38  & 16.73  & 16.66  & 8.04  & 9.05  & 8.15  & 8.25  & 8.67  & 8.78  & 15.15  & 18.30  & 15.08  & 15.39  & 15.23  & 15.33  \\
          & leader & 3.45  & 5.18  & 3.45  & 3.48  & 3.48  & 3.47  & 20.58  & 24.59  & 21.47  & 21.20  & 21.53  & 21.60  & 8.56  & 10.74  & 9.89  & 9.55  & 9.69  & 9.84  & 16.83  & 21.06  & 17.33  & 17.26  & 17.43  & 17.65  \\
          & savior & 0.32  & 0.53  & 0.37  & 0.37  & 0.36  & 0.37  & 14.18  & 16.14  & 14.41  & 14.43  & 14.49  & 14.47  & 6.56  & 7.41  & 6.78  & 6.79  & 6.84  & 6.75  & 11.90  & 14.72  & 11.90  & 12.04  & 12.23  & 12.15  \\
          & winner & 13.54  & 20.33  & 13.56  & 13.56  & 13.55  & 13.56  & 20.60  & 23.87  & 21.31  & 21.44  & 21.35  & 21.46  & 7.49  & 9.00  & 8.48  & 8.32  & 8.80  & 8.57  & 19.53  & 23.83  & 19.82  & 19.70  & 19.79  & 19.68  \\
    \midrule
    \multirow{15}[2]{*}{\rotatebox{90}{ALIGN}} & criminal & 1.13  & 1.75  & 1.13  & 1.14  & 1.15  & 1.14  & 8.30  & 9.29  & 8.42  & 8.43  & 8.50  & 8.45  & 0.27  & 0.58  & 0.38  & 0.70  & 0.74  & 0.73  & 0.18  & 0.01  & 0.01  & 0.01  & 0.01  & 0.01  \\
          & failure & 0.23  & 0.40  & 0.24  & 0.25  & 0.25  & 0.25  & 4.71  & 6.71  & 5.61  & 5.54  & 5.67  & 5.68  & 0.10  & 0.48  & 0.33  & 0.51  & 0.54  & 0.64  & 0.01  & 0.01  & 0.01  & 0.01  & 0.01  & 0.01  \\
          & fraudster & 0.76  & 1.22  & 0.82  & 0.81  & 0.81  & 0.81  & 9.41  & 10.83  & 9.63  & 9.68  & 9.70  & 9.71  & 0.24  & 0.34  & 0.48  & 0.54  & 0.50  & 0.63  & -0.03  & 0.00  & 0.00  & 0.00  & 0.00  & 0.00  \\
          & liar  & 0.56  & 0.85  & 0.57  & 0.56  & 0.57  & 0.56  & 5.76  & 6.31  & 5.82  & 5.90  & 5.83  & 5.84  & 0.33  & 0.40  & 0.49  & 0.40  & 0.55  & 0.56  & 0.00  & 0.00  & 0.00  & 0.00  & 0.00  & -0.01  \\
          & thief & 0.42  & 0.67  & 0.46  & 0.45  & 0.46  & 0.43  & 5.70  & 7.24  & 6.30  & 6.22  & 6.50  & 6.38  & 0.29  & 0.30  & 0.30  & 0.37  & 0.51  & 0.55  & -0.03  & 0.00  & 0.00  & 0.00  & 0.00  & -0.01  \\
          & citizen & 0.98  & 1.52  & 0.98  & 0.99  & 0.98  & 0.98  & 11.15  & 12.93  & 11.33  & 11.37  & 11.32  & 11.35  & 1.11  & 1.73  & 1.38  & 1.35  & 1.56  & 1.57  & 6.03  & 7.55  & 6.85  & 7.20  & 7.09  & 6.82  \\
          & individual & 0.66  & 1.01  & 0.68  & 0.68  & 0.68  & 0.68  & 16.35  & 18.81  & 16.45  & 16.49  & 16.53  & 16.53  & 0.85  & 0.85  & 1.03  & 1.00  & 1.05  & 1.04  & 7.48  & 9.24  & 8.58  & 8.27  & 8.10  & 8.32  \\
          & person & 2.00  & 2.99  & 1.99  & 1.99  & 1.99  & 1.99  & 18.02  & 20.82  & 18.18  & 18.33  & 18.31  & 18.29  & 0.49  & 0.78  & 0.79  & 1.01  & 1.00  & 1.01  & 12.85  & 16.16  & 13.82  & 13.54  & 13.51  & 13.88  \\
          & stranger & 5.48  & 8.22  & 5.48  & 5.48  & 5.48  & 5.48  & 9.67  & 10.91  & 9.62  & 9.67  & 9.82  & 9.73  & 0.51  & 0.60  & 0.77  & 0.80  & 0.80  & 0.85  & 0.25  & 0.44  & 0.80  & 0.83  & 1.03  & 1.21  \\
          & worker & 0.01  & 0.04  & 0.01  & 0.07  & 0.04  & 0.02  & 5.65  & 6.68  & 5.73  & 5.79  & 5.77  & 5.88  & 3.17  & 3.43  & 3.00  & 3.20  & 3.20  & 3.39  & 0.02  & -0.02  & -0.01  & -0.01  & -0.01  & -0.01  \\
          & genius & 5.62  & 8.48  & 5.65  & 5.64  & 5.64  & 5.64  & 13.30  & 15.31  & 13.44  & 13.42  & 13.48  & 13.46  & 0.63  & 0.82  & 0.82  & 0.86  & 0.93  & 0.85  & 3.20  & 3.04  & 1.77  & 2.15  & 2.24  & 2.17  \\
          & hero  & 0.77  & 1.20  & 0.78  & 0.78  & 0.77  & 0.78  & 10.85  & 12.64  & 11.21  & 11.19  & 11.20  & 11.12  & 0.34  & 0.37  & 0.54  & 0.51  & 0.56  & 0.70  & 0.43  & 0.08  & -0.01  & 0.01  & 0.01  & 0.03  \\
          & leader & 1.55  & 2.32  & 1.53  & 1.53  & 1.54  & 1.53  & 9.51  & 10.91  & 9.97  & 9.84  & 10.03  & 9.94  & 0.79  & 0.96  & 1.10  & 0.94  & 1.22  & 1.35  & 2.11  & 2.17  & 2.02  & 1.68  & 1.80  & 1.76  \\
          & savior & 0.13  & 0.24  & 0.16  & 0.18  & 0.17  & 0.17  & 5.22  & 6.32  & 5.43  & 5.53  & 5.67  & 5.63  & 0.26  & 0.47  & 0.39  & 0.44  & 0.53  & 0.54  & 0.02  & 0.02  & 0.02  & 0.02  & 0.02  & 0.01  \\
          & winner & 0.67  & 0.96  & 0.65  & 0.64  & 0.65  & 0.64  & 6.93  & 8.69  & 7.47  & 7.46  & 7.42  & 7.46  & 0.32  & 0.55  & 0.58  & 0.68  & 0.76  & 0.69  & 0.02  & 0.03  & 0.02  & 0.02  & 0.02  & 0.02  \\
    \midrule
    \multirow{15}[2]{*}{\rotatebox{90}{BridgeTower}} & criminal & 0.23  & 0.34  & 0.23  & 0.23  & 47.61  & 47.61  & 3.14  & 3.01  & 1.20  & 2.17  & 1.00  & 1.04  & 0.10  & 0.11  & 0.10  & 0.11  & 28.27  & 28.20  & 1.04  & 1.21  & 1.02  & 1.04  & 17.78  & 17.97  \\
          & failure & 0.08  & 0.11  & 0.08  & 0.08  & 60.40  & 60.47  & 2.92  & 2.02  & 2.14  & 0.77  & 0.97  & 0.97  & 0.14  & 0.16  & 0.15  & 0.15  & 31.54  & 31.23  & 0.03  & 0.03  & 0.02  & 0.03  & 35.23  & 35.31  \\
          & fraudster & 0.19  & 0.29  & 0.19  & 0.19  & 45.10  & 45.10  & 2.66  & 3.57  & 1.82  & 1.06  & 1.02  & 1.00  & 0.03  & 0.03  & 0.03  & 0.03  & 17.84  & 17.71  & 1.34  & 1.62  & 1.32  & 1.34  & 22.53  & 22.19  \\
          & liar  & 0.99  & 1.48  & 0.99  & 0.99  & 63.09  & 63.11  & 1.98  & 3.40  & 2.08  & 2.28  & 4.39  & 4.33  & 0.29  & 0.32  & 0.28  & 0.28  & 31.76  & 31.81  & 0.94  & 1.14  & 0.64  & 0.95  & 37.09  & 37.17  \\
          & thief & 0.13  & 0.19  & 0.13  & 0.13  & 10.75  & 10.75  & 2.37  & 1.70  & 1.90  & 2.11  & 1.09  & 1.08  & 0.08  & 0.09  & 0.08  & 0.08  & 16.05  & 15.63  & 0.64  & 0.61  & 0.61  & 0.63  & 9.93  & 9.63  \\
          & citizen & 0.56  & 0.83  & 0.56  & 0.56  & 1.31  & 1.33  & 3.93  & 5.85  & 3.28  & 3.06  & 1.08  & 1.10  & 1.04  & 1.16  & 1.07  & 1.04  & 14.18  & 14.36  & 3.41  & 4.08  & 3.41  & 3.39  & 9.75  & 9.49  \\
          & individual & 14.72  & 22.08  & 14.72  & 14.72  & 5.26  & 5.24  & 3.40  & 3.98  & 3.55  & 3.38  & 1.09  & 1.10  & 0.74  & 0.82  & 0.75  & 0.71  & 22.42  & 22.73  & 3.80  & 4.61  & 3.84  & 3.87  & 8.96  & 9.20  \\
          & person & 12.00  & 18.00  & 12.00  & 12.00  & 2.88  & 2.91  & 3.21  & 3.46  & 3.24  & 3.20  & 1.13  & 1.13  & 0.84  & 0.92  & 0.82  & 0.85  & 24.08  & 24.59  & 4.90  & 5.96  & 4.97  & 4.95  & 9.79  & 9.63  \\
          & stranger & 5.51  & 8.27  & 5.51  & 5.51  & 58.05  & 58.04  & 3.51  & 3.84  & 3.59  & 3.44  & 1.02  & 1.00  & 0.34  & 0.37  & 0.34  & 0.34  & 24.01  & 24.44  & 4.91  & 5.97  & 4.93  & 4.96  & 35.63  & 35.66  \\
          & worker & 0.04  & 0.05  & 0.04  & 0.04  & 0.98  & 0.95  & 2.26  & 1.46  & 1.27  & 2.28  & 1.11  & 1.13  & 2.86  & 3.13  & 2.86  & 2.86  & 12.78  & 12.48  & 0.00  & 0.09  & 0.05  & 0.07  & 10.58  & 10.69  \\
          & genius & 1.50  & 2.25  & 1.50  & 1.50  & 52.44  & 52.43  & 2.59  & 2.88  & 3.41  & 2.72  & 1.05  & 1.08  & 0.25  & 0.26  & 0.24  & 0.24  & 23.24  & 23.55  & 3.31  & 3.98  & 3.33  & 3.31  & 17.45  & 17.56  \\
          & hero  & 2.24  & 3.36  & 2.24  & 2.24  & 1.38  & 1.38  & 2.25  & 3.56  & 1.94  & 1.93  & 1.14  & 1.12  & 0.43  & 0.48  & 0.43  & 0.43  & 15.53  & 15.13  & 1.33  & 1.58  & 1.33  & 1.29  & 10.61  & 10.44  \\
          & leader & 4.54  & 6.81  & 4.54  & 4.54  & 9.82  & 9.85  & 5.88  & 4.16  & 5.06  & 3.00  & 1.13  & 1.10  & 2.15  & 2.35  & 2.14  & 2.16  & 16.34  & 16.59  & 4.29  & 5.09  & 4.26  & 4.27  & 9.33  & 9.59  \\
          & savior & 0.45  & 0.67  & 0.45  & 0.45  & 0.87  & 0.88  & 3.57  & 2.22  & 3.34  & 1.99  & 1.10  & 1.08  & 0.33  & 0.35  & 0.32  & 0.32  & 13.62  & 13.21  & 0.88  & 0.78  & 0.88  & 0.88  & 10.73  & 10.74  \\
          & winner & 4.05  & 6.07  & 4.05  & 4.05  & 0.92  & 0.91  & 0.77  & 2.90  & 3.47  & 1.86  & 1.08  & 1.07  & 0.44  & 0.49  & 0.44  & 0.42  & 13.41  & 13.24  & 1.01  & 1.20  & 1.01  & 1.01  & 10.70  & 10.80  \\
    \midrule
    \multirow{15}[2]{*}{\rotatebox{90}{OWLv2}} & criminal & 47.53  & 71.22  & 47.65  & 47.57  & 0.23  & 0.23  & 1.03  & 0.97  & 0.98  & 1.00  & 2.26  & 1.84  & 27.78  & 30.08  & 28.48  & 27.50  & 0.09  & 0.09  & 16.88  & 21.43  & 17.55  & 17.72  & 1.03  & 1.02  \\
          & failure & 60.27  & 90.54  & 60.34  & 60.51  & 0.08  & 0.08  & 0.94  & 1.12  & 1.01  & 0.98  & 1.78  & 2.37  & 30.64  & 34.56  & 30.75  & 31.30  & 0.14  & 0.14  & 34.96  & 42.05  & 35.15  & 35.01  & 0.03  & 0.02  \\
          & fraudster & 44.98  & 67.59  & 45.09  & 45.05  & 0.19  & 0.19  & 1.00  & 1.05  & 0.91  & 0.97  & 1.20  & 2.61  & 16.58  & 19.97  & 17.88  & 17.90  & 0.02  & 0.04  & 22.40  & 26.67  & 22.25  & 22.50  & 1.33  & 1.31  \\
          & liar  & 62.75  & 94.41  & 63.05  & 63.16  & 0.99  & 0.99  & 4.32  & 4.85  & 4.29  & 4.33  & 2.22  & 2.28  & 31.48  & 34.81  & 31.88  & 31.96  & 0.29  & 0.32  & 36.97  & 44.65  & 37.05  & 37.17  & 0.95  & 0.93  \\
          & thief & 10.75  & 16.03  & 10.77  & 10.76  & 0.13  & 0.13  & 1.07  & 1.35  & 1.04  & 1.10  & 1.66  & 2.81  & 14.97  & 17.63  & 15.46  & 15.86  & 0.08  & 0.07  & 10.17  & 11.94  & 9.70  & 9.97  & 0.64  & 0.62  \\
          & citizen & 1.33  & 1.96  & 1.31  & 1.30  & 0.56  & 0.56  & 1.10  & 1.28  & 1.13  & 1.12  & 3.31  & 2.89  & 13.92  & 15.76  & 14.15  & 14.14  & 1.03  & 0.97  & 9.60  & 11.27  & 9.28  & 9.24  & 3.39  & 3.40  \\
          & individual & 5.23  & 7.82  & 5.24  & 5.18  & 14.72  & 14.72  & 1.05  & 1.32  & 1.07  & 1.03  & 3.49  & 3.49  & 22.63  & 24.94  & 22.54  & 22.41  & 0.76  & 0.72  & 9.54  & 11.18  & 9.20  & 9.11  & 3.88  & 3.84  \\
          & person & 2.60  & 4.36  & 2.94  & 2.86  & 12.00  & 11.99  & 1.10  & 1.34  & 1.02  & 1.07  & 3.21  & 3.21  & 23.76  & 27.13  & 24.59  & 23.83  & 0.83  & 0.87  & 8.88  & 11.94  & 9.75  & 9.07  & 4.96  & 4.95  \\
          & stranger & 58.02  & 87.08  & 58.08  & 58.07  & 5.51  & 5.51  & 1.01  & 1.18  & 1.01  & 1.06  & 3.58  & 3.58  & 22.55  & 26.50  & 24.43  & 24.46  & 0.33  & 0.32  & 35.21  & 41.64  & 35.49  & 35.58  & 4.99  & 4.93  \\
          & worker & 0.84  & 1.36  & 0.93  & 0.87  & 0.04  & 0.04  & 1.10  & 1.32  & 1.11  & 1.12  & 3.28  & 2.52  & 11.73  & 13.74  & 12.82  & 12.47  & 2.84  & 2.80  & 9.85  & 13.00  & 10.38  & 10.57  & 0.07  & 0.06  \\
          & genius & 52.38  & 78.44  & 52.35  & 52.40  & 1.50  & 1.50  & 0.97  & 1.23  & 1.03  & 1.02  & 2.52  & 2.58  & 21.93  & 25.84  & 23.20  & 23.38  & 0.23  & 0.24  & 17.64  & 20.94  & 17.70  & 17.37  & 3.31  & 3.32  \\
          & hero  & 1.14  & 1.89  & 1.24  & 1.27  & 2.24  & 2.24  & 0.93  & 1.14  & 1.05  & 1.08  & 2.25  & 1.70  & 14.73  & 17.20  & 15.76  & 14.88  & 0.44  & 0.42  & 10.38  & 11.93  & 10.90  & 10.49  & 1.33  & 1.33  \\
          & leader & 9.88  & 14.83  & 9.87  & 9.79  & 4.54  & 4.54  & 1.15  & 1.20  & 1.06  & 1.02  & 4.16  & 4.33  & 16.59  & 18.11  & 16.51  & 16.59  & 2.17  & 2.08  & 9.15  & 12.04  & 9.52  & 9.47  & 4.28  & 4.27  \\
          & savior & 0.95  & 1.30  & 0.94  & 0.89  & 0.45  & 0.45  & 1.01  & 1.20  & 1.10  & 1.10  & 1.89  & 2.54  & 13.22  & 14.17  & 13.49  & 13.05  & 0.31  & 0.29  & 10.73  & 12.68  & 10.62  & 10.65  & 0.86  & 0.86  \\
          & winner & 0.82  & 0.86  & 0.94  & 0.83  & 4.05  & 4.05  & 1.06  & 1.21  & 1.11  & 1.07  & 1.80  & 2.27  & 13.09  & 14.90  & 12.40  & 13.15  & 0.44  & 0.44  & 10.86  & 12.87  & 10.87  & 10.91  & 1.01  & 1.01  \\
    \bottomrule
    \end{tabular}}%
  \label{tab:ablation_N}%
\end{table}%

\end{document}